\documentclass[twoside,11pt]{article}

\usepackage{jmlr2e}
\usepackage{times}
\usepackage{graphicx} 
\usepackage{subcaption}
\usepackage{natbib}
\usepackage{amsmath}
\usepackage{amssymb}
\usepackage{amsfonts}
\usepackage{algorithm}
\usepackage{algorithmic}

\newcommand{\fracpartial}[2]{\frac{\partial #1}{\partial  #2}}
\def\abovespace{\abovestrut{0.20in}}

\def\belowspace{\belowstrut{0.10in}}
\newtheorem{defn}{Definition}

\jmlrheading{17}{2017}{?-??}{1/17}{??/??}{Mehmet E. Basbug and Barbara E. Engelhardt}

\ShortHeadings{Adaptive Clustering for Heterogeneous Data}{Basbug and Engelhardt}
\firstpageno{1}

\begin{document}

\title{AdaCluster : Adaptive Clustering for Heterogeneous Data}

\author{\name Mehmet E. Basbug \email mbasbug@princeton.edu \\
       \addr Department of Electrical Engineering\\
       Princeton University\\
       Princeton, NJ 08540, USA
       \AND
       \name Barbara E. Engelhardt \email bee@princeton.edu \\
       \addr Department of Computer Science\\
       \addr Center for Statistics and Machine Learning\\
       Princeton University\\
       Princeton, NJ 08540, USA}

\editor{EDITOR}

\maketitle

\begin{abstract}
  Clustering algorithms start with a fixed divergence, which captures the possibly asymmetric distance between a sample and a centroid. In the mixture model setting, the sample distribution plays the same role. When all attributes have the same topology and dispersion, the data are said to be \emph{homogeneous}. If the prior knowledge of the distribution is inaccurate or the set of plausible distributions is large, an adaptive approach is essential. The motivation is more compelling for \emph{heterogeneous} data, where the dispersion or the topology differs among attributes. We propose an adaptive approach to clustering using classes of parametrized Bregman divergences. We first show that the density of a steep exponential dispersion model (EDM) can be represented with a Bregman divergence. We then propose \textit{AdaCluster}, an expectation-maximization (EM) algorithm to cluster heterogeneous data using classes of steep EDMs. We compare AdaCluster with EM for a Gaussian mixture model on synthetic data and nine UCI data sets. We also propose an adaptive hard clustering algorithm based on Generalized Method of Moments. We compare the hard clustering algorithm with k-means on the UCI data sets. We empirically verified that adaptively learning the underlying topology yields better clustering of heterogeneous data.
\end{abstract}


\section{Introduction}
\label{sec:ada_introduction}
Despite the general tendency towards more complex models and algorithms, k-means remains one of the most popular clustering algorithms~\citep{kulis2012revisiting}. The k-means algorithm makes three simplifying assumptions: i) clusters are disjoint, implying a hard assignment of samples to clusters; ii) the number of clusters is fixed and specified ahead of time; and iii) the distance between two samples is measured with Euclidean distance~\citep{bishop2006pattern}. Many extensions to k-means have been proposed to relax these assumptions. For example, soft k-means, an expectation maximization (EM) algorithm for a homogeneous Gaussian mixture model, allows for a soft assignment of samples to clusters. In fact, the k-means algorithm is the equivalent of the soft k-means under the asymptotically small variance assumption~\citep{bishop2006pattern}. This probabilistic model connection opened the door to non-parametric clustering using the DP-means algorithm, removing the second assumption of static numbers of clusters. DP-means was derived using the Gibbs sampler for a Dirichlet process (DP) Gaussian mixture model~\citep{kulis2012revisiting}. Finally, the Gaussian distribution implicit in the third assumption of Euclidean geometry has been generalized to regular exponential family distributions~\citep{banerjee2005clustering}. The generalization is established via the bijection between regular exponential family distributions and regular Bregman divergences. The resulting algorithm, \emph{Bregman hard clustering}, also has a soft counterpart, \emph{Bregman soft clustering}~\citep{banerjee2005clustering}. Furthermore, \citep{jiang2012small} showed that DP-means may be used with any regular Bregman divergence, resulting in a nonparametric hard-clustering algorithm for homogeneous mixture of regular exponential family distributions. After relaxing the three k-means assumptions, we are still bound to specify a fixed divergence metric to measure the distance from the centroid. In the case of mixture models, the corresponding choice is for the sample attribute distributions.

Describing data in its natural habitat can yield tremendous benefits, especially when the errors are so large that the Gaussian assumption does not hold~\citep{fisher1953dispersion}. The motivation to identify the true topological properties of the data has led to the study of dispersion models and most notably to the generalized linear models~\citep{nelder1972generalized}. In this paper, we attempt to use dispersion models in clustering setting and provide ways of identifying the topology of the data with mild assumptions on the data distribution.

With a good understanding of the topology of the data, we can create a custom mixture model by specifying the known distribution of each sample attribute. Under certain regularity and independence conditions, we can combine these attribute-specific mixture models into a single mixture model. This can be a tedious job when the number of attributes is large. A more challenging and common scenario is when we do not know the underlying distribution of each attribute a priori. The conventional approach is to cast this problem as model selection. We can select between alternative models using a criterion such as a Bayes factor, a likelihood ratio, or an information criterion. In the case of homogeneous data, i.e. when all the attributes have the same topology and dispersion, the model selection reduces to identifying the single true distribution within the set of plausible distributions. Model selection in the case of heterogeneous data tends to be more challenging due to combinatorial nature of the problem. If we have $M_{j}$ choices for the $j^{th}$ attribute, then we need to explore $\prod_{j}M_{j}$ models, which can be overwhelming. We address this problem in three steps. We first present a large family of distributions, namely steep exponential dispersion models (EDMs), where the analytic expressions for the mean and dispersion parameter estimators are available. We then propose parametrized sub-families of steep EDMs for different data types (positive, non-negative, real, discrete, continuous etc.). Finally, we derive an adaptive clustering algorithm for heterogeneous data that can learn the underlying distribution of each attribute given its type.

In Section \ref{sec:ada_background}, we introduce convex duality concepts, Bregman divergences, exponential family distributions, and exponential dispersion models. We show that the density of a steep exponential dispersion model can be represented in terms of a Bregman divergence. This finding allows us to parametrize class of steep EDMs by describing the divergence-generating functions in the dual domain. With the dual formulation, we are able to keep the support of the class consistent and analyze the variance-mean relationship more clearly.

In Section~\ref{sec:ada_parameterized}, we introduce parametrized families of steep EDMs for non-negative discrete, continuous, positive continuous, and non-negative continuous data. Section~\ref{subsec:ada_edm_nnd} describes a class for non-negative discrete data that includes Poisson and negative binomial distributions. This class is primarily suited for count data such as number of hits recorded with a Geiger counter, page views for a website, patient days spent in a clinic and game scores in a contest~\citep{hilbe2011negative}. For continuous data on the real line, we suggest another class including generalized hyperbolic secant (GHS) and Gaussian distributions in  Section~\ref{subsec:ada_edm_rc}. This class includes asymmetric distributions with applications to financial data~\citep{fischer2013generalized}. Furthermore, when data live on a unit interval, the proposed class can be utilized by transforming data with a logit function first. Most notably, beta distribution and logit-normal distribution map to GHS and Gaussian distributions under the logit transform, respectively. The families included in Section~\ref{subsec:ada_edm_nnd} and~\ref{subsec:ada_edm_rc} are sub-families of the Morris class characterized by the quadratic polynomial variance function~\citep{morris1982natural}. In Section~\ref{subsec:ada_edm_pc}, we discuss another prominent class of EDMs, the Tweedie class, with a defining property of a power variance function~\citep{tweedie1947functions}. We show that many members of the Tweedie class are steep EDMs, hence, can be represented with Bregman divergences. We show that the Tweedie class can be used to analyze positive continuous data as well as for non-negative continuous data depending on the choice of the hyper-parameter domain.

In Section~\ref{sec:ada_mixture}, we write the quasi-log likelihood of heterogeneous data under mixture of steep EDMs in terms of Bregman divergences using saddle-point approximation. We propose an EM algorithm with closed-form updates for the mean and dispersion parameters and numerical procedures for learning the hyper-parameters identifying the underlying distribution of each attribute. The resulting algorithm, \textit{AdaCluster}, is similar to the Bregman soft-clustering algorithm~\citep{banerjee2005clustering} with the added capability of handling heterogenous data. We also consider the Bayesian treatment of the model and show that the MAP estimates of the mean parameters also have closed form when we have conjugate priors. Furthermore, we show that inverse-Gamma distribution is the conjugate prior to the dispersion parameter of steep EDMs under the saddle-point approximation.

In Section~\ref{sec:ada_hard}, we turn our attention to the hard clustering problem for heterogeneous data. We first note that the AdaCluster algorithm does not have a hard clustering counterpart under the asymptotically small variance assumption. We further show that the likelihood approach falls short in learning the hyper-parameters defining the topology of the data. We then propose an algorithm based on generalized method of moments (GMoM) using the moment conditions derived from the parametrized variance functions. Since variance functions and divergence-generating functions characterize steep EDMs uniquely, the parametrized families introduced in Section~\ref{sec:ada_parameterized} can be used within the hard-clustering setting as well. We derive a k-means like algorithm, \textit{GMoM-HC}, that adaptively learns the topology and partitions data into hard clusters.

Finally, in Section~\ref{sec:ada_results}, we start by examining the performance of EM algorithm for GMM in homogeneous mixture models with non-Gaussian distributions. We consider the simple task of density estimation with mixture models and show that the GMM fit to non-Gaussian data fails to capture the true density as the variance-mean relationship deviates from the Gaussian assumption of independence. In another synthetic data analysis, we compare EM for GMM and AdaCluster in clustering heterogeneous data. We obtain substantial improvements with AdaCluster in terms of normalized mutual information and likelihood. Finally, we look at nine data sets from the UCI repository with distinct topology and dispersion characteristics. We compare AdaCluster and GMoM-HC with EM for GMM and k-means in terms of clustering performance and conclude that AdaCluster clusters data better than GMM while GMoM-HC and k-means perform comparably. 

\section{Background}
\label{sec:ada_background}
\subsection{Convex duality}
\label{subsec:ada_convex}
We begin with a discussion of two fundamental properties of convex functions and their relationship to one another. The first property is \emph{steepness}, which is important with respect to the existence and uniqueness of a maximum likelihood estimator for exponential family distributions (discussed further in Section~\ref{subsec:ada_steep}). The second property is \emph{strict convexity}, which enables us to generate Bregman divergences. Bregman divergences are an important class of distortion functions that have favorable properties such as non-negativity, convexity, and linearity, which we exploit later~\citep{banerjee2005clustering}. We then state a theorem establishing the connection between steep convex functions and a particular type of strict convex functions, \emph{essentially strict convex functions}. For the remainder of the paper we assume that the convex functions are \emph{proper}, i.e., they are finite, and their effective domain is non-empty. We start by formally defining the steepness property.
\begin{defn}\citep[Section 26]{rockafellar1970convex}
A proper convex function $f$ is \emph{essentially smooth} if it is differentiable throughout nonempty $A = int(dom\;f)$\footnotemark and $\lim_{i\rightarrow \infty} \left | \triangledown f(\boldsymbol{x}_{i}) \right | = \infty$ whenever $\boldsymbol{x}_{1}$, $\boldsymbol{x}_{2}$, $\dots$, is a sequence in $A$ converging to a boundary point $\boldsymbol{x}\in bd\;A$.
\end{defn}
As in \citep{nielsen1978information}, we use the terms \textit{essentially smooth} and \textit{steep} interchangeably. Although the term \textit{steep} may be used to describe non-convex functions, here we limit our focus to convex functions.
\footnotetext{Throughout the paper we use \textit{int}, \textit{ri}, \textit{bd} to denote the interior, relative interior and the boundary of a set, respectively. For a function $f$, we denote the domain, sub-differential mapping, gradient, and convex conjugate with $dom\;f$, $\partial f$, $\triangledown f$, and $f^*$. We denote the interval $[0,\infty)$ with $\mathbb{R}_{0}$ and $(0,\infty)$ with $\mathbb{R}_{+}$. Similarly, we denote the set $\{0,1,\dots\}$ with $\mathbb{Z}_{0}$ and $\{1,2,\dots\}$ with $\mathbb{Z}_{+}$.}

Strict convexity is another intrinsic property of convex functions. The effective domain of a convex function is used to determine if the function is strictly convex; however, more refined characterizations of strict convexity exist. One such characterization is \textit{essential strict convexity}, which we define as follows.
\begin{defn}\citep[Section 26]{rockafellar1970convex}
A proper convex function $f$ is called \emph{essentially strictly convex} if $f$ is strictly convex on every convex subset of $dom\;\partial f$.
\end{defn}
To better understand the difference between \emph{essentially strictly convex} and \emph{strictly convex} functions, we state the following theorem.
\begin{theorem}\citep[Theorem 5.22]{nielsen1978information}
\label{th:ada_domain_hierarchy}
Suppose $f$ is a convex function. Then
\begin{align*}
ri(dom\;f) \subset dom\;\partial f \subset dom\;f.
\end{align*}
\end{theorem}
From Theorem~\ref{th:ada_domain_hierarchy}, it follows that, if $f$ is essentially strictly convex, then $f$ is strictly convex on $ri(dom\;f)$ but not necessarily on $dom\;f$. Furthermore, every essentially strictly convex function with an open domain is also strictly convex.

\begin{example}\citep[Section 26]{rockafellar1970convex}
\label{ex:ada_essentially_strict_convexity}
An example of an essentially strictly convex but not strictly convex function is
\begin{align}
f(\boldsymbol{x}) &=\begin{cases}
x_{2}^{2}/(2 x_{1}) - 2 \sqrt{x_{2}} & x_{1} > 0, x_{2} \geq 0\\
0 & x_{1} = 0, x_{2} = 0.
\end{cases}
\end{align}
$f$ is strictly convex on $dom\;\partial f = \mathbb{R}^2_{+}$ but not on $dom\;f$ since $f$ is identically zero for $x_{2}=0$ and $x_{1} \neq 0$. Also note that $f$ is strictly convex on $ri(dom\;f) = \mathbb{R}^2_{+}$. Incidentally, $f$ is also a steep function.
\end{example}

In information geometry, strict convexity is a useful property because strictly convex functions enable the construction of Bregman divergences. Bregman divergences are defined as follows.
\begin{defn}\citep{bregman1967relaxation}
Let $f:A\rightarrow \mathbb{R}$ be a strictly convex function defined on a convex set $A\subseteq \mathbb{R}^{d}$ such that $f$ is differentiable on nonempty $ri(A)$. The \emph{Bregman divergence} $d_{f} : A \times ri(A) \rightarrow \mathbb{R}_{0}$ is defined as
\begin{align}
d_{f}(\boldsymbol{x},\boldsymbol{y}) = f(\boldsymbol{x}) - f(\boldsymbol{y}) - \langle \boldsymbol{x} - \boldsymbol{y},\triangledown f(\boldsymbol{y})\rangle. \label{eq:ada_bregman_defn}
\end{align}
\end{defn}
Bregman divergences include many useful distortion functions such as squared loss, Mahalanobis distance, KL-divergence, logistic loss, and Itakura-Saito distance. Bregman divergences are non-negative, convex in the first argument, linear, and, often, non-symmetric. From the statistical perspective, Bregman divergences are appealing because, when the density of an exponential family distribution can be represented in terms of a Bregman divergence, the maximum likelihood estimate of the natural parameter is unique and can be calculated easily from the sample mean~\citep{banerjee2005clustering}. This observation hints at a duality between steep convex functions and strictly convex functions. In general, there is no duality between the two families; however, the connection between essentially smooth and essentially strictly convex functions can be established with the following theorem.
\begin{theorem}\citep[Theorem 26.3]{rockafellar1970convex}
\label{th:ada_steep_convex_duality}
A closed proper convex function is essentially strictly convex if and only if its conjugate is essentially smooth.
\end{theorem}
An immediate corollary of Theorem~\ref{th:ada_steep_convex_duality} is that the conjugate of an essentially smooth and lower semi-continuous function is essentially strictly convex. This is due to the bi-conjugate theorem: If a convex function $f$ is lower semi-continuous, then $f^{**} = f$~\citep[Theorem 12.2]{rockafellar1970convex}. Theorem~\ref{th:ada_steep_convex_duality} is especially important in characterizing the dual of an exponential family distribution. We use Theorem~\ref{th:ada_steep_convex_duality} to show that the density of a certain subset of exponential family distributions can be described in terms of Bregman divergences. Before examining the connection between the exponential family and Bregman divergences, we formally introduce exponential family distributions.

\subsection{Exponential family distributions}
\label{subsec:ada_exponential}
In statistical theory, the discussion of exponential family distributions can be traced back to~\citep{fisher1934two}. A comprehensive account of the theoretical properties of the exponential family was given by~\citep{nielsen1978information,lehmann2006theory}. The relation to information geometry was first explored in~\citep{efron1975defining} and later detailed in~\citep{shun2012differential}. In this section, we start with the definition of the full, regular, steep and natural exponential families. We then present the notion of conjugate prior to an exponential family distribution and finally discuss the general convexity properties of exponential family distributions and their relation to the topology.
\newpage
\begin{defn}
A family of distributions $\mathcal{F}_{(\Psi,\boldsymbol{\Theta})} = \left \{ p_{\Psi}(\cdot\mid\boldsymbol{\theta}) : \boldsymbol{\theta} \in \boldsymbol{\Theta} \subset \mathbb{R}^n \right \}$ is called an \emph{exponential family} if
\begin{align}
p_{\Psi}(\boldsymbol{x}\mid \boldsymbol{\theta}) &= \exp\left(\left\langle \boldsymbol{\theta}, T(\boldsymbol{x}) \right \rangle - \Psi(\boldsymbol{\theta})\right)h(\boldsymbol{x})\label{eq:ada_exponential_family_density}
\end{align}
where $\boldsymbol{\theta}$ is the natural parameter, $\Psi(\boldsymbol{\theta})$ is the log-partition function, $h(\boldsymbol{x})$ is the base measure, and $T(\boldsymbol{x})$ is the minimal sufficient statistics. Let $S_{\boldsymbol{\theta}} = \{\boldsymbol{x} : p_{\Psi}(\boldsymbol{x}\mid \boldsymbol{\theta}) > 0 \}$ be the support of $p_{\Psi}(\boldsymbol{x}\mid \boldsymbol{\theta})$. The smallest interval containing the union of $\left \{ S_{\boldsymbol{\theta}} \mid \boldsymbol{\theta} \in \boldsymbol{\Theta} \right \}$ is called the \emph{convex support} and denoted by $C$. Furthermore, when $\boldsymbol{\Theta} = dom(\Psi)$, the family is said to be \emph{full}. If $\boldsymbol{\Theta} = int(\boldsymbol{\Theta}) = dom(\Psi)$, the family is called \emph{regular}. If $\Psi(\boldsymbol{\theta})$ is \emph{steep}, then the family is also called \emph{steep}. Finally, if $\boldsymbol{\Theta} \subset \mathbb{R}$ and $T(x) = x$, then the family is called \emph{natural}.
\end{defn}
Gaussian, gamma, binomial, and Poisson distributions are well-known members of the \textit{regular} exponential family. Steep exponential family is a larger set that includes inverse-Gaussian, stable and Bessel-type distributions~\citep{bar1986reproducibility} as well as all the regular exponential family distributions~\citep[Theorem 8.2]{nielsen1978information}. In mixture model setting, the regular exponential family is particularly easy to work with as the distributions in this family have densities that can be expressed in terms of Bregman divergences~\citep{banerjee2005clustering}. In Section~\ref{subsec:ada_steep}, we present a similar result for the steep exponential family.

Natural exponential family is another useful sub-family of the exponential family with notable members such as Poisson, exponential, geometric, and Bernoulli distributions. Although the natural exponential family is not as expressive as we might want for analysis of arbitrary attributes, they form the basis for a much wider range of distributions, exponential dispersion models, which we discuss in further detail in Section~\ref{subsec:ada_edm}.

Bayesian treatment of the exponential family often involves a conjugate prior on the natural parameter. Having a conjugate prior ensures the posterior distribution of the natural parameter, $p_{\Psi}(\boldsymbol{\theta} \mid \boldsymbol{x})$, belongs to the same family as the prior distribution, $p_{\Psi}(\boldsymbol{\theta} \mid \boldsymbol{a},b)$. The conjugate prior for an exponential family distribution $p_{\Psi}(\boldsymbol{x} \mid \boldsymbol{\theta})$ with minimal statistics $T(\boldsymbol{x}) = \boldsymbol{x}$ has the following form
\begin{align}
p_{\Psi}(\boldsymbol{\theta}\mid \boldsymbol{a},b) &= \exp\left(b(\left\langle \boldsymbol{\theta},\boldsymbol{a} \right \rangle -\Psi(\boldsymbol{\theta}))\right)m(\boldsymbol{a},b)\label{eq:ada_exponential_family_conjugate}
\end{align}
where $\boldsymbol{a} \in ri(C)$ and $b > 0$ are the parameters of the conjugate prior, and $m(\boldsymbol{a},b)$ is a base measure. The geometric interpretation of the conjugate prior for exponential family distributions is discussed in~\citep{agarwal2010geometric}. The parameter $\boldsymbol{a}$ reflects our prior knowledge about where the mean is located. The second parameter $b$ measures how confident we are about this belief in relation to the sample size. This interpretation becomes clearer when we discuss the duality in Section~\ref{subsec:ada_steep}.

To discuss the topological interpretation of an exponential family distribution, it is essential to spell out the convexity properties of the log-partition function and its convex conjugate. The following theorem summarizes the fundamental convexity properties of $\Psi$ and $\Psi^{*}$.
\begin{theorem}\citep[Theorem 9.1]{nielsen1978information}
\label{th:ada_psi_properties}
For an exponential family $\mathcal{F}_{(\Psi,\boldsymbol{\Theta})}$, the log-partition function $\Psi$ is a closed and strictly convex function and satisfies $\Psi = \Psi^{**}$. Furthermore, $\Psi^*$ is a closed and essentially smooth convex function with $int(C) \subset dom(\Psi^*) \subset C$.
\end{theorem}
The topological properties of an exponential family distribution is encompassed in the log-partition function. To see this, we need to analyze the density of an exponential family distribution in the dual domain. In addition to the convex conjugate of the log-partition function, the first and second derivatives of $\Psi$ are also needed for such analysis. We define two useful functions in this regard.
\begin{defn}
For a non-degenerate exponential family $\mathcal{F}_{(\Psi,\boldsymbol{\Theta})}$, we define the \emph{mean-value mapping} $\boldsymbol{\tau} : int(\boldsymbol{\Theta})\rightarrow \Omega$ as $\boldsymbol{\tau}(\boldsymbol{\theta}) = \triangledown\Psi(\boldsymbol{\theta})$. The range of the mean-value mapping $\Omega$ is called the \emph{mean domain}. In addition, if $\boldsymbol{\tau}^{-1}(\boldsymbol{x})$ is differentiable within $\Omega$, then we define the \emph{variance function} $\boldsymbol{\upsilon} : \Omega \rightarrow \mathbb{R}_{+}$ as $\boldsymbol{\upsilon}(\boldsymbol{x}) = \triangledown\boldsymbol{\tau}^{-1}(\boldsymbol{x})$.
\end{defn}

In Table~\ref{table:ada_nef}, we show the densities of six natural exponential family distributions with the mapping between the common parameters and the natural parameter, the corresponding log-partition function, mean-value mapping and variance functions. One striking observation is that the variance function has at most a quadratic term and can be written as a polynomial for these six distributions. This may seem like a coincidence; however, many of the common exponential family distributions can be characterized by a quadratic polynomial variance function~\citep{morris1982natural}. The parametrization of variance functions is often employed to characterize family of distributions. We'll discuss three different parametrization of variance functions in Section~\ref{sec:ada_parameterized}. The variance function also comes up in saddle-point approximation of the density which we'll discuss in the Section~\ref{subsec:ada_edm}. However, we first turn our attention to the mean-value mapping and its key role in finding the dual of a density function.

\begin{table}[t!]
\caption{{\bf Six common natural exponential family distributions.} Bernouilli, geometric, Poisson, exponential, Rayleigh, chi-squared distributions with the common density formulation, support ($S$) and the corresponding exponential family parametrization with natural parameter ($\theta$), log-partition function ($\Psi$), mean-value mapping ($\tau$) and the variance function ($\upsilon$).}\label{table:ada_nef}
\begin{center}
\begin{sc}
\def\arraystretch{1.5}
\begin{tabular}{llccccc}
\hline
distribution & density & S & $\theta$ & $\Psi(\theta)$ & $\tau(\theta)$ & $\upsilon(x)$ \\
\hline
Bernouilli & $p^{x}(1-p)^{1-x}$ & $\{0,1\}$ & $\log(\frac{p}{1-p})$ & $\log(1+e^{\theta})$&  $\frac{e^{\theta}}{1+e^{\theta}}$& $x(1-x)$ \\
Geometric & $(1-p)^{x-1}p$ & $\mathbb{Z}_{+}$ & $\log(\frac{1-p}{p})$ & $\log(\frac{e^{\theta}}{1+e^{\theta}})$&  $\frac{1}{1+e^{\theta}}$& $x(x-1)$ \\
Poisson & $e^{-\lambda}\frac{\lambda^{x}}{x!}$ & $\mathbb{N}$ &  $\log(\lambda)$ & $e^{\theta}$&  $e^{\theta}$& $x$ \\
Chi-Squared & $\frac{x^{(k-2)/2}e^{-x/2}}{2^{k/2}\Gamma(k/2)}$ & $\mathbb{R}_{0}$ &  $\log(\frac{k}{2})$ & $\frac{1}{2}e^{2\theta}$&  $e^{2\theta}$& $2x$ \\
Exponential & $\lambda e^{-\lambda x}$ & $\mathbb{R}_{0}$ &  $\frac{-1}{\lambda}$ & $-\log(\theta)$&  $\frac{-1}{\theta}$& $x^2$ \\
Rayleigh & $\frac{x}{\sigma^2} e^{-\frac{x^2}{2\sigma^2}}$ & $\mathbb{R}_{0}$ &  $\frac{-\sqrt{2\pi}}{(4-\pi)\sigma}$ & $\frac{-\pi\log(\theta(\pi-4))}{(4-\pi)}$&  $\frac{-\pi}{(4-\pi)\theta}$& $\frac{4-\pi}{\pi}x^2$ \\
\hline
\end{tabular}
\end{sc}
\end{center}
\end{table}

\subsection{Steep exponential family}
\label{subsec:ada_steep}
Steepness is an intrinsic property of an exponential family distribution and preserved under linear transformations. Steep exponential family distributions allow us to find analytic solutions for the maximum likelihood (ML) and maximum a posteriori (MAP) estimates of $\theta$. As we will discuss shortly, representing the density of a steep exponential family distribution in dual domain is immensely useful. Dual formulation allows us to parametrize family of distributions, find analytic estimates of the mean parameter and use saddle-point approximation. The following theorem outlines the key properties of steep exponential family.

\begin{theorem}\citep[Corollary 9.6]{nielsen1978information}
\label{th:ada_steep_ml}
Suppose $\Psi$ is steep, which is true in particular if $\Psi$ is regular. The maximum likelihood estimate exists if and only if $T(\boldsymbol{x}) \in int(C)$, and then it is unique. Furthermore, $\Omega = int(C)$ and $C \setminus \Omega = bd\;C$. Finally, the maximum likelihood estimator $\boldsymbol{\theta^{ML}}$ is the one-to-one mapping from $\Omega$ onto $int(\boldsymbol{\Theta})$ whose inverse is $\boldsymbol{\tau}$.
\end{theorem}

The relationship between the ML estimator and the mean-value mapping hints at a convex duality. More concretely, $(int(\boldsymbol{\Theta}),\Psi)$ and $(\Omega,\bar{\Psi}^*)$ are Legendre duals of each other, and duality is established through the mean-value mapping~\citep[Section~7]{nielsen1978information}. We note that $\bar{\Psi}^*$ is the restriction of $\Psi^*$ to $\Omega$ and $\tau$ is a one-to-one mapping and has an inverse for steep exponential family distributions. The Legendre duality can be summarized with the following equations
\begin{align*}
\bar{\Psi}^*(\boldsymbol{x}) &= \sup_{\boldsymbol{\theta}\in int(\boldsymbol{\Theta})}(\left\langle \boldsymbol{\theta}, \boldsymbol{x} \right \rangle -\Psi(\boldsymbol{\theta}))\\
\boldsymbol{\tau}^{-1}(\boldsymbol{x}) &=
\underset{\boldsymbol{\theta}\in int(\boldsymbol{\Theta})}{\text{argsup}}(\left\langle \boldsymbol{\theta}, \boldsymbol{x} \right \rangle -\Psi(\boldsymbol{\theta}))\\
\Psi(\boldsymbol{\theta}) &= \sup_{\boldsymbol{x}\in \Omega}(\left\langle \boldsymbol{x}, \boldsymbol{\theta} \right \rangle - \bar{\Psi}^*(\boldsymbol{x}))\\
\boldsymbol{\tau}(\boldsymbol{\theta}) &=
\underset{\boldsymbol{x}\in \Omega}{\text{argsup}}(\left\langle \boldsymbol{x}, \boldsymbol{\theta} \right \rangle - \bar{\Psi}^*(\boldsymbol{x})).
\end{align*}

The goal of the dual formulation is to express the density in terms of divergence from the mean. Legendre dual of the log-partition function is the first step; however, we need to make sure that the divergence from mean to each point in the support---more generally the convex support since the support may vary with the natural parameter---is well-defined. Notice that the Legendre dual is defined on $\Omega$ and not $C$; hence, we can only talk about strict convexity on $\Omega$. To achieve dual formulation of density as a divergence from mean, we need to investigate the convex conjugate of $(\boldsymbol{\Theta},\Psi)$ and not just the Legendre dual of $(int(\boldsymbol{\Theta}),\Psi)$.

If we have a regular exponential family, then $\boldsymbol{\Theta}$ is an open set ($bd\;\boldsymbol{\Theta} = \emptyset$) and therefore $(\boldsymbol{\Theta} = int(\boldsymbol{\Theta}),\Psi)$ is of Legendre type. The Legendre dual of $(\boldsymbol{\Theta},\Psi)$ is $(\Omega,\bar{\Psi}^*)$; however, this does not imply that the convex support is also an open set. The convex support and the convex conjugate should be calculated with the closure construction of $\bar{\Psi}^*$. As noted in~\citep[Theorem 7.5]{rockafellar1970convex}, taking limits of $\bar{\Psi}^*$ at the boundaries of $C$ can also be used to find the convex conjugate of $\Psi$.

We note that for a continuous steep exponential family distribution, the probability mass on $bd\;\Omega$ is zero; therefore, the closure construction may not be as important~\citep{jorgensen1997theory}. For steep discrete distributions, however, $C \setminus \Omega$ plays an important role. A curious example highlighting this point and showing the interplay of $\Theta, \Omega, S$, and $C$ is the Bernoulli distribution.
\begin{example}
\label{ex:ada_bernouilli}
For a Bernoulli distribution (see Table~\ref{table:ada_nef}), we have $\Psi(\theta) = \log(1+e^{\theta})$ with $\Theta = \mathbb{R}$. Since $\Theta$ is open, we have a regular---thus, also steep---exponential family distribution. The mean-value mapping is $\tau(\theta) = \frac{e^{\theta}}{1+e^{\theta}}$ with range $\Omega = (0,1)$. The Legendre dual of $(\Theta,\Psi)$ is $\bar{\Psi}^*(t) = t\log(t) + (1-t)\log({1-t})$ with $dom(\bar{\Psi}^*) = (0,1)$. The support is $S = \{0,1\}$; hence, the convex support is $C = [0,1]$. We have $\Omega = ri(C)$ and $C \setminus \Omega = bd\;\Omega$. Note that the support, $S$, is disjoint from the mean domain, $\Omega$. Also note that \emph{all} probability mass is on $C \setminus \Omega$ and the Legendre dual of $\Psi$ is not defined on $C \setminus \Omega$. We should find the convex conjugate of $\Psi$ and investigate its behavior on $C$. The convex conjugate can be calculated by taking limits: $\Psi^*(t) = \bar{\Psi}^*(t)\;\forall t\in \Omega$, $\Psi^*(0) = \lim_{t\rightarrow 0^{+}} \bar{\Psi}^*(t) = 0$ and $\Psi^*(1) = \lim_{t\rightarrow 1^{-}} \bar{\Psi}^*(t) = 0$. In this particular case, $\Psi^*$ turns out to be strictly convex on $C$.
\end{example}

Next, we describe the concluding theorem of this section, where we show that the density of a steep natural exponential family can be written in terms of a Bregman divergence. A similar connection was explored in previous work~\citep{banerjee2005clustering} for the regular exponential family, which is a subset of the steep exponential family. In general, not every steep exponential family distribution has a corresponding Bregman divergence.
\begin{theorem}
\label{th:ada_steep_bregman}
Suppose we have a steep natural exponential family distribution $p_{\Psi}(x\mid \theta)$. Then the convex conjugate of the log-partition function $\phi \doteq \Psi^*$ is a strictly convex function, and $dom\;\phi = C$. Furthermore, the density can be expressed as
\begin{align*}
p_{\Psi}(x\mid \theta) &= \exp\left(x\theta - \Psi(\theta)\right)h(x)\\
&= \exp\left(-d_{\phi}(x,\mu)\right)g_{\phi}(x)\\
&= p_{\phi}(x\mid \mu)
\end{align*}
where $d_{\phi}(x,y)$ is the Bregman divergence generated from $\phi$, $\mu = \tau^{-1}(\theta)$ is the mean parameter, and $g_{\phi}(x) = exp(\phi(x))h(x)$ is the base measure and $p_{\phi}(x\mid \mu)$ is the dual formulation of the density.
\end{theorem}
\begin{proof}
From Theorem~\ref{th:ada_psi_properties}, $\phi^*=\Psi$, and $\Psi$ is essentially smooth; hence, $\phi$ is essentially strictly convex by Theorem~\ref{th:ada_steep_convex_duality}. Moreover, $\phi$ is closed and therefore lower semi-continuous. By Theorem~\ref{th:ada_psi_properties} and Theorem~\ref{th:ada_steep_ml}, $dom\;\phi = C$. If $C$ is open, we conclude that $\phi$ is strictly convex due to Theorem~\ref{th:ada_domain_hierarchy}. When $C$ is a bounded or a half-bounded interval, the inclusion of the boundary points do not change strict convexity since $\phi$ is lower semi-continuous. Therefore $\phi$ is strictly convex on $C$. Furthermore, $\phi$ is an essentially smooth function by Theorem~\ref{th:ada_psi_properties}; therefore, $\phi$ is differentiable on $int(C)$ and $C$ is non-empty. It then follows that $\phi$ generates a Bregman divergence $d_{\phi} : C \times int(C) \rightarrow \mathbb{R}_{0}$. The dual density formulation follows from~\citep[p.49]{jorgensen1997theory}.
\end{proof}
We note that Theorem~\ref{th:ada_steep_bregman} does not generalize to the steep exponential family with multi-dimensional natural parameter i.e. $\boldsymbol{\Theta} \nsubseteq \mathbb{R}$. Recall the function $f$ in Example~\ref{ex:ada_essentially_strict_convexity} and suppose we have an exponential family distribution with the log-partition function $f^{*}$. Due to Theorem~\ref{th:ada_steep_convex_duality}, $f^{*}$ is steep since $f$ is essentially strictly convex. The exponential family distribution $f^{*}$ generates is steep but not natural. Thus, Theorem~\ref{th:ada_steep_bregman} does not guarantee that the corresponding divergence is of Bregman type. In fact, we know that $f$ is not strictly convex; therefore, $f$ does not generate a Bregman divergence.

Although steep natural exponential family includes many useful distributions such as Gaussian with known variance, exponential, Bernoulli, Poisson, geometric, chi-squared, Rayleigh, binomial with known count and negative binomial with known count, it is not as expressive as we desire to characterize the underlying distribution of an arbitrary sample attribute. To better model an arbitrary error distribution, we need to capture both the mean location and the dispersion. Therefore, we are interested in two-parameter univariate distributions such as Gaussian, gamma, inverse-Gaussian, negative binomial and binomial distributions in our analysis. Fortunately, there is a straightforward way to expand the natural exponential family and capture the topological characteristics of data more accurately. In the next section, we explore steep exponential dispersion models and describe how to expand the steep natural exponential family.

\begin{figure*}[p!]
  \begin{center}
      \begin{subfigure}[h]{0.375\textwidth}
        \includegraphics[width=\textwidth]{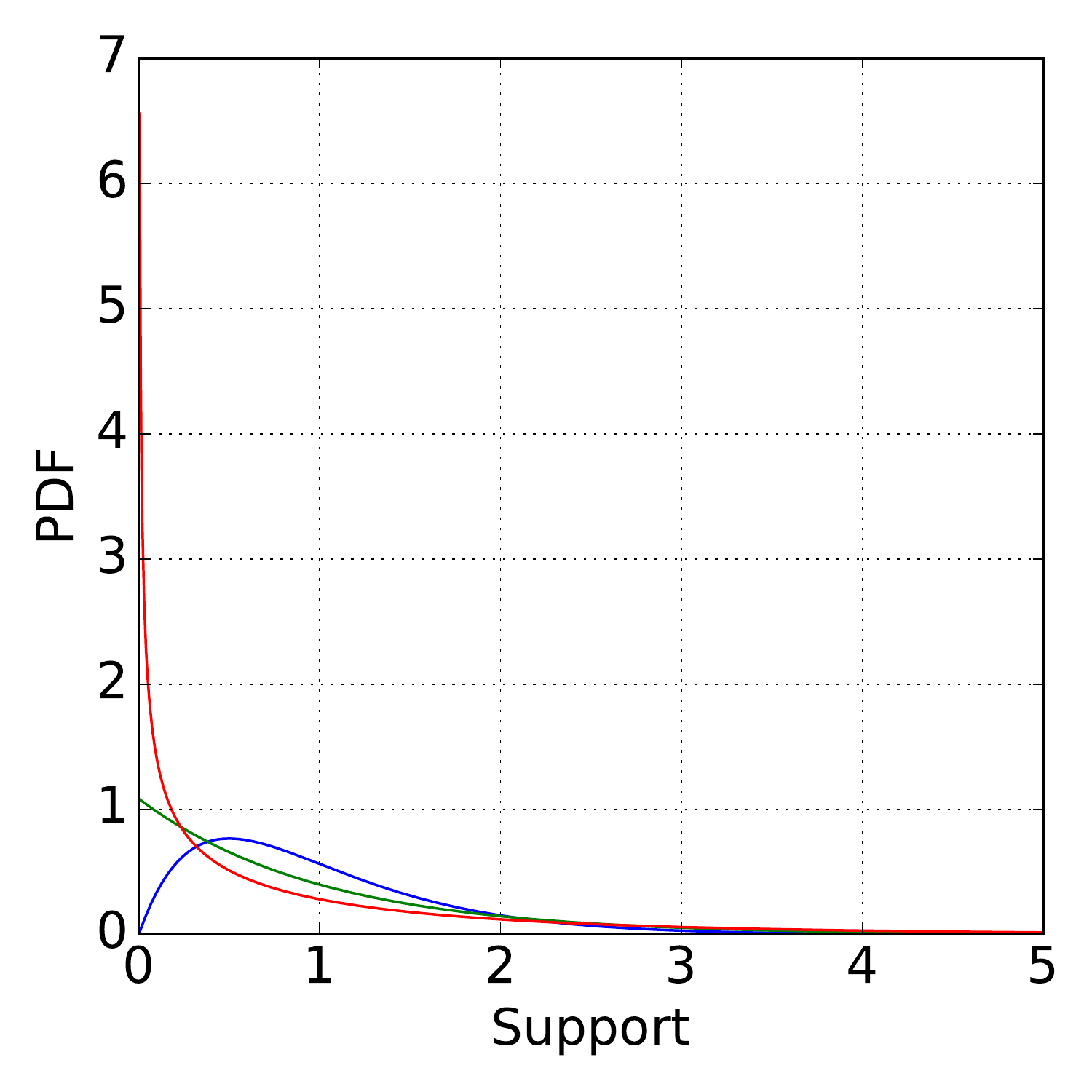}
        \caption{}
        \label{fig:ada_topology_000}
      \end{subfigure}
      ~
      \begin{subfigure}[h]{0.375\textwidth}
        \includegraphics[width=\textwidth]{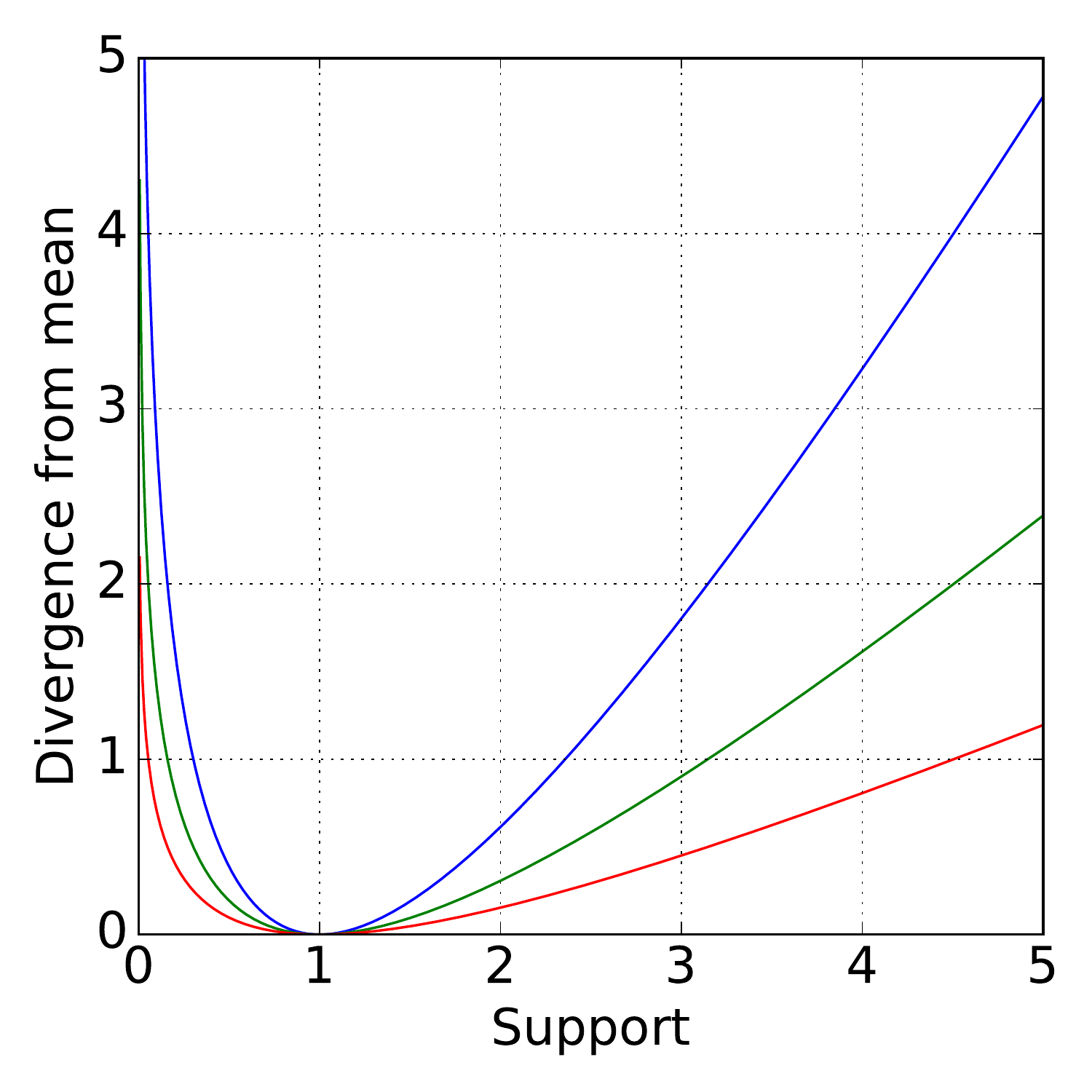}
        \caption{}
        \label{fig:ada_topology_div_000}
      \end{subfigure}

      \begin{subfigure}[h]{0.375\textwidth}
        \includegraphics[width=\textwidth]{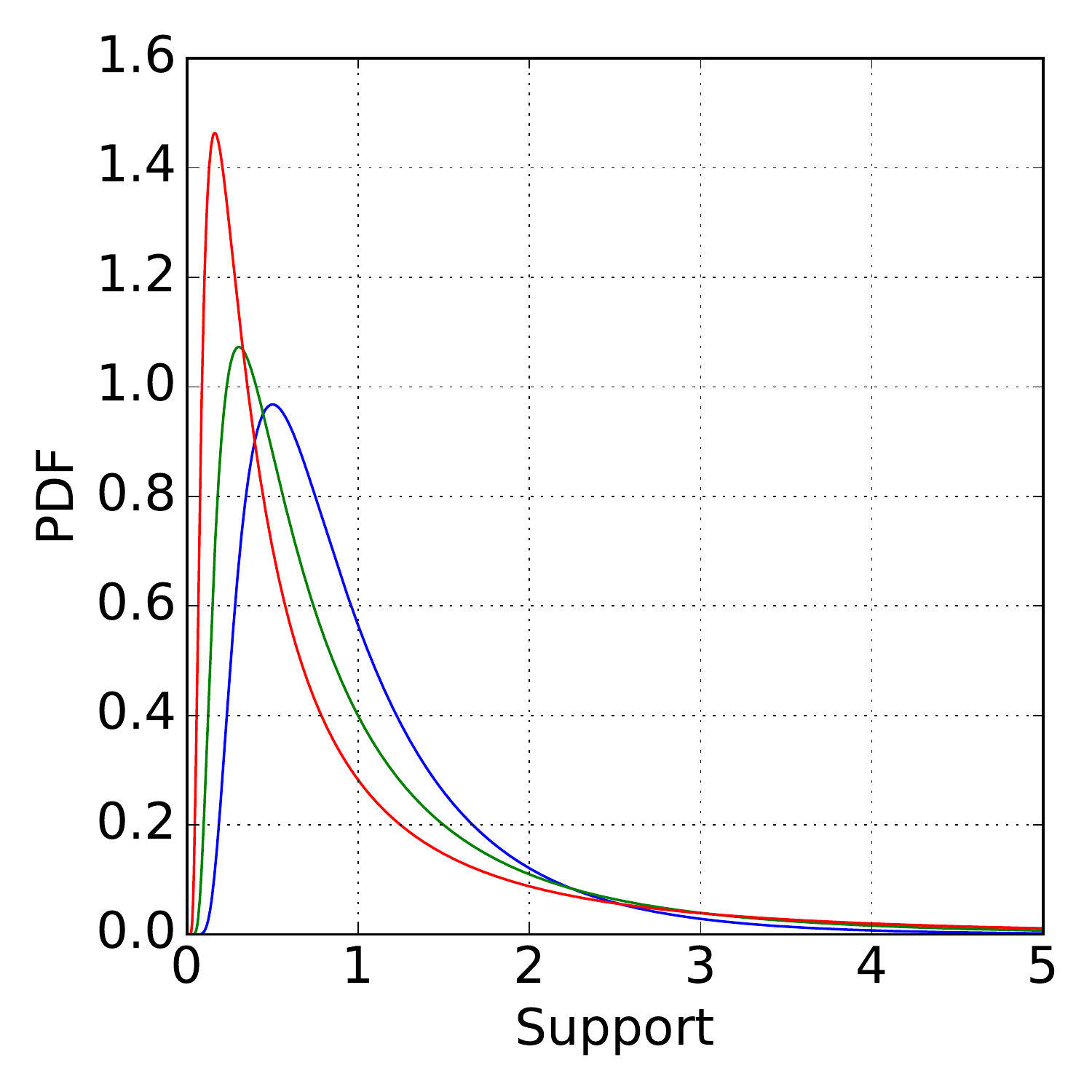}
        \caption{}
        \label{fig:ada_topology_-1000}
      \end{subfigure}
      ~
      \begin{subfigure}[h]{0.375\textwidth}
        \includegraphics[width=\textwidth]{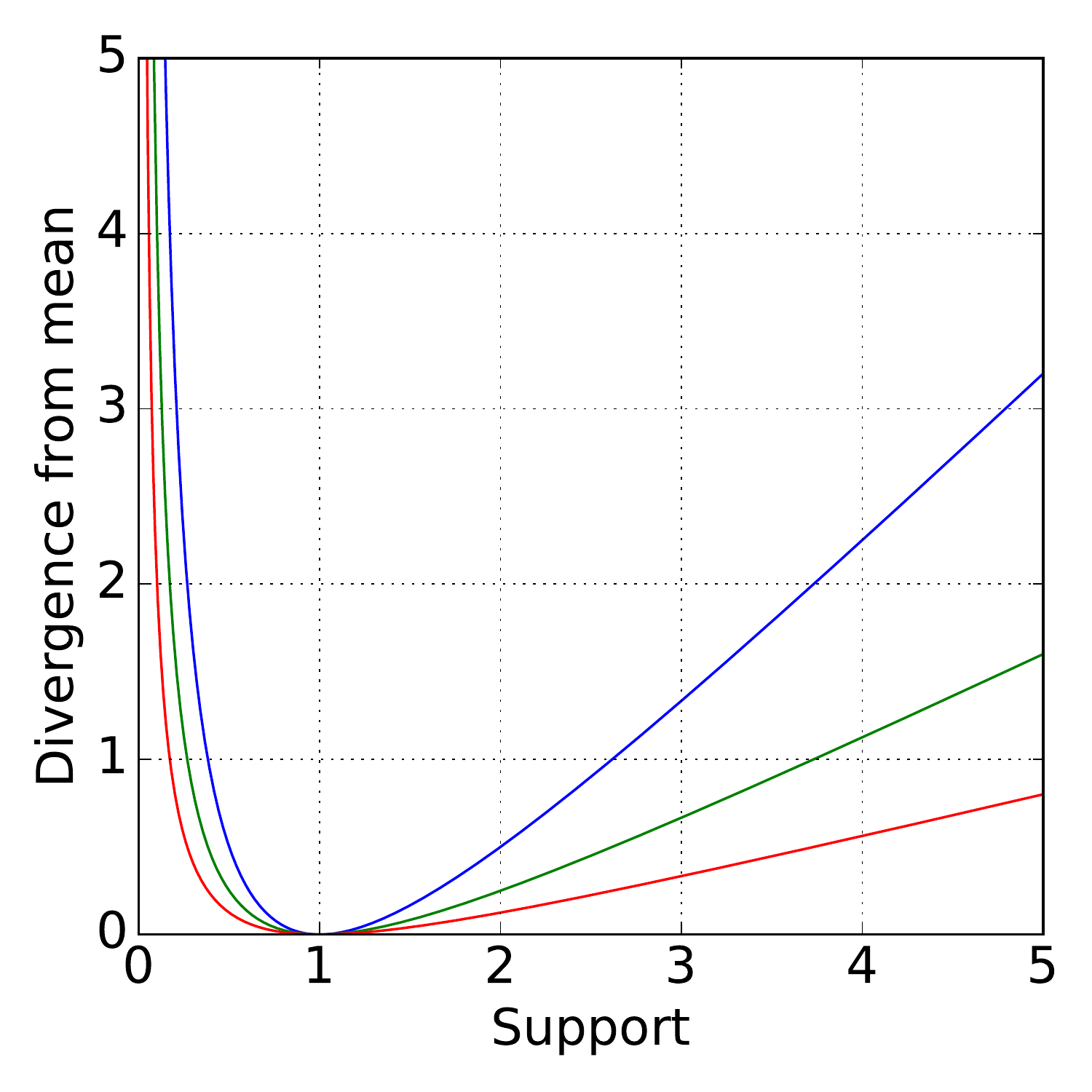}
        \caption{}
        \label{fig:ada_topology_div_-1000}
      \end{subfigure}

      \begin{subfigure}[h]{0.375\textwidth}
        \includegraphics[width=\textwidth]{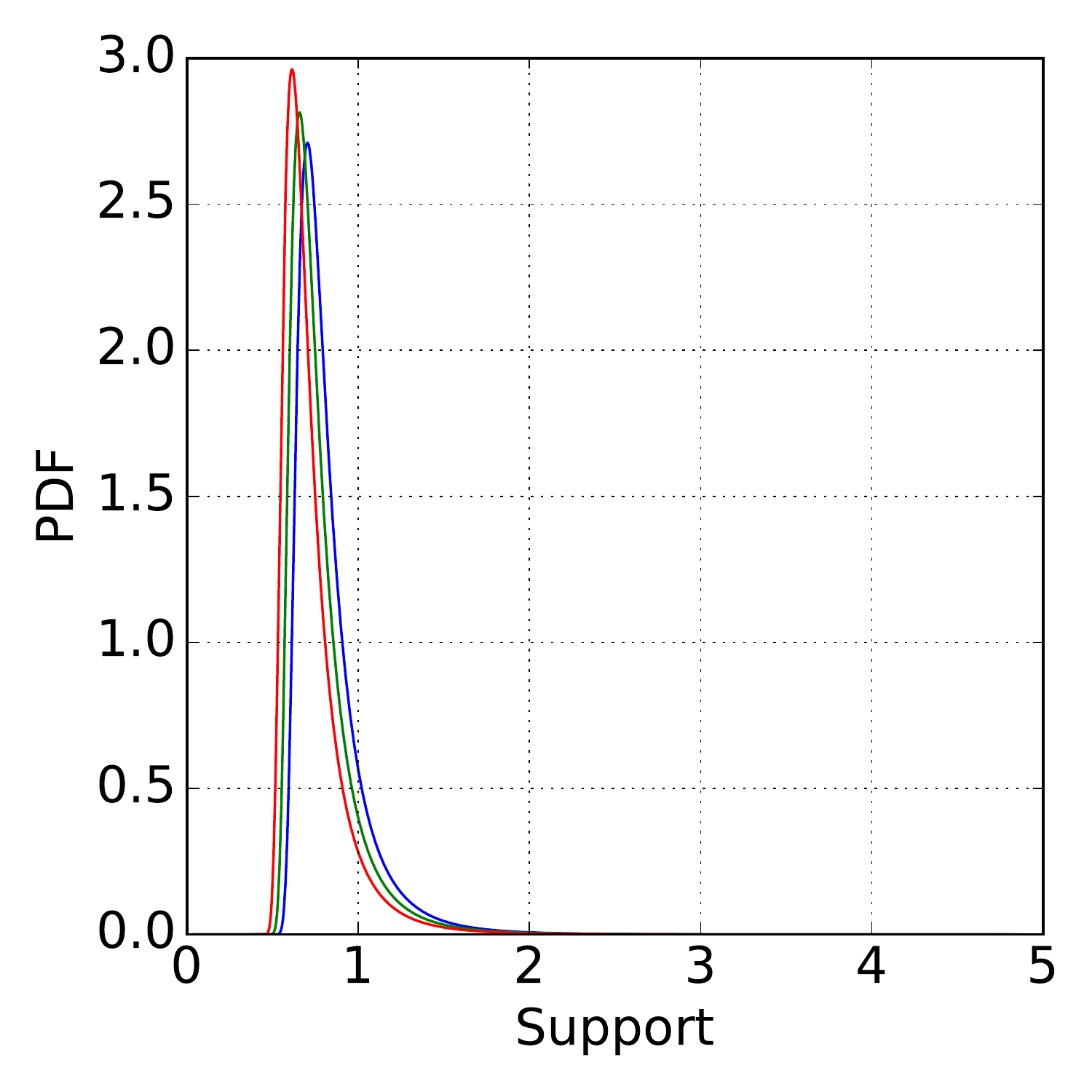}
        \caption{}
        \label{fig:ada_topology_-10000}
      \end{subfigure}
      ~
      \begin{subfigure}[h]{0.375\textwidth}
        \includegraphics[width=\textwidth]{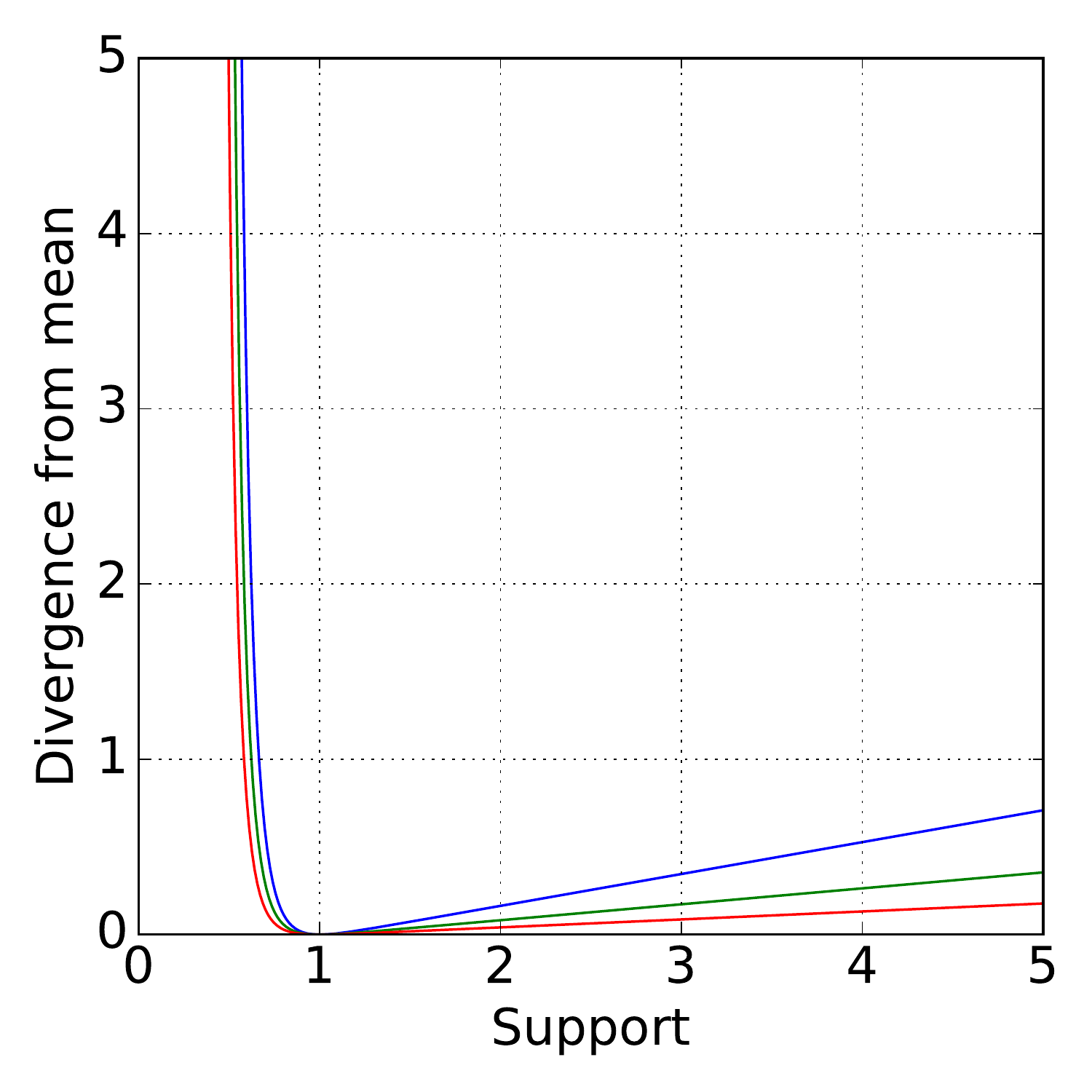}
        \caption{}
        \label{fig:ada_topology_div_-10000}
      \end{subfigure}
      \caption{{\bf Duality in steep EDMs with support on positive real numbers .} Probability density function and the corresponding Bregman divergence functions for three EDMs: a-b) gamma c-d) inverse-Gaussian and e-f) stable. In all cases, the mean parameter ($\mu$) is fixed to $1.0$ and the dispersion parameter ($\kappa$) is set to $0.5$ (blue) $1.0$ (green) $2.0$ (red).}
      \label{fig:ada_topology}
  \end{center}
  \vspace{-12pt}
\end{figure*}

\subsection{Exponential dispersion models}
\label{subsec:ada_edm}
In Theorem~\ref{th:ada_steep_bregman}, we showed that the probability density of a sample data point is related to its distance from the mean. In particular, when the density is a steep natural exponential family distribution, the divergence function is of Bregman type. Suppose we stretch out the underlying geometry around the mean i.e. we scale the divergence from the mean with a constant for every point in the convex support. The resulting geometry induces a new probability distribution---albeit closely related to the original one. For every steep natural exponential family distribution, we may scale the Bregman divergence with a positive constant, which preserves steepness of $\Psi$ and the strict convexity of $\phi$. The newly generated family of distributions is called steep \textit{exponential dispersion models}. In Fig.~\ref{fig:ada_topology}, we show how scaling plays out in both domains for three different distributions: gamma, inverse-Gaussian and a stable distribution. One might argue that the stretching does not fundamentally change the underlying topology but simply scales it. For the remainder of the paper, we refer to the geometry induced by the unit Bregman divergence as the \textit{topology} and the scaling of the unit divergence as the \textit{dispersion}.

There are two equivalent parametrization of EDMs: additive and reproductive~\citep[Chapter~3]{jorgensen1997theory}. We give the definition for the reproductive formulation of EDMs.
\begin{defn}
A family of distributions $\mathcal{F}_{(\Psi,\Theta)}= \left \{ p_{\Psi}(\cdot\mid \theta,\kappa) : \theta \in \Theta \subset \mathbb{R}, \kappa \in \mathbb{R}_{+} \right \}$ is called a \emph{reproductive exponential dispersion model} if
\begin{align*}
p_{\Psi}(x\mid \theta,\kappa) &= \exp\left( \frac{1}{\kappa} (x\theta - \Psi(\theta))\right)h(x,\kappa)
\end{align*}
where $\theta$ is the natural parameter, $\kappa$ is the dispersion parameter, $\Psi(\theta)$ is the unit log-partition function, and $h(x,\kappa)$ is the base measure.
\end{defn}
Suppose we have a steep natural exponential family distribution $p_{\Psi}(x\mid \theta)$, and let $p_{\phi}(x\mid \mu)$ be its dual formulation. With change of variables $\tilde{\theta} = \theta / \kappa$ and $\tilde{\Psi}(\tilde{\theta}) = \Psi(\kappa\tilde{\theta})/\kappa$, we obtain the following relationships using the properties of convex conjugacy:
\begin{align}
\tilde{\phi}(\tilde{t}) &= \phi(\tilde{t})/\kappa \\
d_{\tilde{\phi}}(x,y) &= d_{\phi}(x,y)/\kappa\\
\tilde{\tau}(\tilde{\theta}) &= \tau(\kappa\tilde{\theta})\\
\tilde{\upsilon}(x) &= \kappa\upsilon(x).\label{eq:ada_scaled_variance}
\end{align}
Notice that the mean-value mapping---hence the mean parameter corresponding to a fixed natural parameter---does not change after the scaling. The divergence from the mean is, however, is scaled by $1/\kappa$ as implied with stretching the geometry around mean. Finally, we write the density of a steep EDM as
\begin{align*}
p_{\Psi}(x \mid \theta,\kappa) = p_{\phi}(x\mid \mu,\kappa) &= \exp\left(-\frac{1}{\kappa}d_{\phi}(x,\mu)\right)g_{\phi}(x,\kappa)
\end{align*}
where $d_{\phi}(x,\mu)$ is the Bregman divergence generated from $\phi$. In the dual form, the density is parametrized with the mean parameter, $\mu$, and the dispersion parameter, $\kappa$. We note that the mean parameter only appears in the Bregman divergence term, $d_{\phi}(x,\mu)$. One important property of Bregman divergences is that the ML estimate of $\mu$ is independent of the divergence~\citep{banerjee2005clustering}. In fact, the ML estimate of the mean is simply the population mean. The following theorem summarizes the results for the ML and MAP estimators of the mean parameter and its dual, the natural parameter.
\begin{theorem}
\label{th:ada_steep_map}
Suppose we have $N$ i.i.d. samples $\{x_{i}\}_{i=1}^{N}$ from a steep EDM $p_{\Psi}(x\mid \theta,\kappa)$. If the sample mean $\bar{x}~\doteq~\frac{1}{N}\sum_{i=1}^{N}x_{i} \in int(C)$, then the unique ML estimate of the mean parameter is $\mu^{ML} = \bar{x}$ and the unique ML estimate of the natural parameter is $\theta^{ML} = \tau^{-1}(\mu^{ML}) \in int(\Theta)$. Suppose we have a conjugate prior $p_{\Psi}(\theta\mid a, b)$ with parameters $a \in int(C)$ and $b>0$ as in Eq~\ref{eq:ada_exponential_family_conjugate}. Then the unique MAP estimate of the mean parameter $\mu^{MAP} = \frac{a b \kappa + N \bar{x}}{b \kappa + N} \in int(C)$, and the unique MAP estimate of the natural parameter is $\theta^{MAP} = \tau^{-1}(\mu^{MAP}) \in int(\Theta)$.
\end{theorem}
\begin{proof}
Follows from Theorem~\ref{th:ada_steep_ml}.
\end{proof}

As opposed to the mean parameter, the dispersion parameter appears in the exponent and in $g_{\phi}(x,\kappa)$; in general, there is no closed-form expression for $g_{\phi}(x,\kappa)$. Fortunately, asymptotic theory suggests that, for $x \in \Omega$, we can approximate the density as
\begin{align}
p_{\phi}(x\mid \mu,\kappa) = \frac{1}{\sqrt{2\pi\kappa\upsilon(x)}}\exp\left(-\frac{1}{\kappa}d_{\phi}(x,\mu)\right)\;\;\;\text{as}\;\;\;\kappa \rightarrow 0.
\label{eq:ada_saddle_cont}
\end{align}
This approximation is known as the \textit{saddle-point approximation}~\citep{daniels1954saddlepoint}. One limitation of the saddle-point approximation is that the sample point must be in $\Omega$.
\begin{example}
\label{ex:ada_inverse_gaussian}
The inverse-Gaussian distribution has the unit log-partition function $\Psi(\theta) = -\sqrt{-2\theta}$ with $\Theta = (-\infty,0]$. Since $\Theta$ is closed and $\lim_{\theta \rightarrow 0}\Psi'(\theta) = \infty$, we have a \textit{steep} but not \textit{regular} exponential family distribution. The mean-value mapping is $\tau(\theta) = 1 / \sqrt{-2\theta}$ with range $\Omega = (0,\infty)$. The convex conjugate of $\Psi$ is $\phi(t) = 1/(2t)$, and the corresponding Bregman divergence is $d_{\phi}(x,y) = (x-y)^2 / (2xy^2)$. The saddle-point approximation is then given by
\begin{align}
p_{\phi}(x\mid \mu,\kappa) = \frac{1}{\sqrt{2\pi\kappa x^3}}\exp\left(-\frac{(x-\mu)^2}{2\kappa x\mu^2}\right).
\end{align}
Coincidentally, this is the exact density function for inverse Gaussian with mean $\mu$ and shape parameter $1/\kappa$. Also note that the support $S = \Omega = (0,\infty)$.
\end{example}
In the case of steep continuous EDMs, we know that $C\setminus \Omega = bd\;\Omega$ and the probability mass on $bd\;\Omega$ is zero~\citep{jorgensen1997theory}. Therefore, $x \in \Omega$ condition is not limiting for such distributions.

Recall the Bernouilli distribution in Example~\ref{ex:ada_bernouilli} has all the probability mass on $bd\;\Omega$. Although Bernouilli distribution itself is not an EDM; there exist discrete EDMs where the probability mass on $bd\;\Omega$ is not zero. In particular, when a steep discrete EDM with a support at zero, the saddle-point approximation must be modified to accommodate the point mass at $bd\;\Omega$ as
\begin{align}
p_{\phi}(x\mid \mu,\kappa) = \sqrt{\frac{\kappa}{2\pi\upsilon(\kappa x + \kappa c)}}\exp\left(-\frac{1}{\kappa}d_{\phi}(\kappa x,\kappa \mu)\right)\;\;\;\text{as}\;\;\;\kappa \rightarrow 0
\label{eq:ada_saddle_discrete}
\end{align}
where $c$ is a small positive constant (usually set to $1/3$)~\citep{mccullagh1989generalized,jorgensen1997theory}. When the support does not include $0$, we set $c = 0$.

\section{Parametrized classes of exponential dispersion models (EDMs)}
\label{sec:ada_parameterized}
We have shown that the density of a steep EDM may be expressed in terms of a Bregman divergence. Each steep EDM family is uniquely characterized by the log-partition function ($\Psi$), the divergence generating function ($\phi$), the mean-value mapping ($\tau$) or the variance function ($\upsilon$). Within a steep EDM family, each distribution is characterized by a dispersion parameter ($\kappa$) and a natural parameter ($\theta$) or a mean parameter ($\mu$). The inverse-Gaussian family in Example~\ref{ex:ada_inverse_gaussian} can be identified with $\Psi(\theta) = -\sqrt{-2\theta}$, $\tau(\theta) = 1 / \sqrt{-2\theta}$, $\phi(t) = 1/(2t)$ or $\upsilon(x) = 1/x^3$. Within the family, there is a unique inverse-Gaussian distribution for every pair $(\theta,\kappa) \in (-\infty,0] \times (0,\infty)$---equivalently, the same distribution can be specified by the pair $(\mu,\kappa) \in (0,\infty) \times (0,\infty)$ where $\mu = \tau(\theta)$.

When fitting a steep EDM distribution to samples, we need to start with a fixed family, such as gamma or inverse-Gaussian, and learn its pair of parameters $(\theta,\kappa)$ or $(\mu,\kappa)$. For a more versatile analysis, we may want to explore a set of steep EDMs instead of a pre-determined one. For instance, when dealing with positive continuous data, we may fit a gamma distribution, an inverse-Gaussian distribution or a stable distribution (Fig.~\ref{fig:ada_topology}). In this section, we investigate parametrized classes of steep EDMs for different data types so as to facilitate such versatile analysis.

We denote a class of steep EDM with $\Phi_{\mathcal{A}}= \left \{ \phi(\cdot \mid \alpha) : \alpha \in \mathcal{A} \right \}$, where $\alpha$ is a hyper-parameter such that $\phi(t \mid \alpha)$ is strictly convex in $t$, and $\fracpartial{\phi}{\alpha}$ and $\frac{\partial^2 \phi}{\partial t\partial \alpha}$ exist $\forall \alpha \in \mathcal{A}$. Each class $\Phi_{\mathcal{A}}$ can be fully specified by $\phi(t \mid \alpha)$ and $\mathcal{A}$. We refer to $\phi(t \mid \alpha)$ as the \emph{divergence-generating mother function} or simply the \emph{mother function}, and $\mathcal{A}$ as the hyper-parameter domain. We focus on classes where the support does not depend on the choice of $\alpha \in \mathcal{A}$ to ensure that the class can be specified by merely knowing the data support. 

We note that the variance function uniquely characterizes a natural exponential family~\citep[Theorem 2.11]{jorgensen1997theory}. It is common to describe a class of EDMs in terms of its \textit{unit variance function}---the variance function of the natural exponential family from which the EDM family is generated. Two well-studied EDM classes, the Morris class~\citep{morris1982natural} and the Tweedie class~\citep{bar1986reproducibility}, are characterized by the quadratic and power unit variance functions, respectively. Not all members of these classes are steep; therefore, the hyper-parameter domain needs to be specified carefully. We first look at a sub-class of the Morris class with support on natural numbers, and we show that the Poisson and negative binomial distributions are members of the proposed class. We then turn to continuous data on the real line and propose another sub-class of the Morris class that includes the Gaussian and generalized hyperbolic secant distributions as members. Next, we investigate the Tweedie class, and we show that, for positive continuous data, it is possible to identify the true distribution among Gaussian, gamma, and inverse-Gaussian distributions by estimating the hyper-parameter $\alpha$.

In each  sub-section, we start with a parametrized unit variance function ($\upsilon$) and derive the corresponding unit log-partition function ($\Psi$). We then analyze the domain of the unit log-partition function and the natural parameter domain ($\Theta$) to ensure the resulting class consists of steep distributions. Next, we derive the conjugate of $\Psi$ with the help of the mean-value mapping ($\tau$) and determine the convex support ($C$) from the range of the mean-value mapping ($\Omega$). Finally, we obtain the corresponding parametrized unit Bregman divergence ($d_{\phi}$) for each member of the proposed class. With the unit variance function and the unit Bregman divergence, we can express the density of a steep EDM within the class in the dual domain using the saddle-point approximation. The results are summarized in Table~\ref{tbl:ada_edm_family}, readers familiar with the Morris and Tweedie classes may skip to Section~\ref{sec:ada_mixture}.

\begin{table*}[t!]
\caption{{\bf Summary of the family of steep EDMs for different data types.} The hyper-parameter domain, $\mathcal{A}$, the variance function, $\upsilon(x\mid\alpha)$, and the Bregman divergence, $d_{\phi}(x,\mu\mid \alpha)$, used in \textit{AdaCluster} for different data types.}
\begin{center}
\begin{sc}
\begin{tabular}{llrrr}
\abovespace\belowspace
Data Type               &            Support &    $\mathcal{A}$ &  $\upsilon(x\mid\alpha)$ & $d_{\phi}(x,\mu\mid \alpha)$ \\
\hline
\abovespace\belowspace
  Non-negative discrete &   $\mathbb{Z}_{0}$ & $\mathbb{R}_{0}$ & $x(1+\alpha x)$ & Eq.~\ref{eq:ada_divergence_nnd} \\
      Positive discrete &   $\mathbb{Z}_{+}$ & $\mathbb{R}_{0}$ & $x(1+\alpha x)$ & Eq.~\ref{eq:ada_divergence_nnd} \\
        Real continuous &       $\mathbb{R}$ & $\mathbb{R}_{0}$ &  $1+\alpha x^2$ &  Eq.~\ref{eq:ada_divergence_rc} \\
Non-negative continuous &   $\mathbb{R}_{0}$ &          $(0,1]$ &  $x^{2-\alpha}$ &  Eq.~\ref{eq:ada_divergence_pc} \\
    Positive continuous &   $\mathbb{R}_{+}$ &    $(-\infty,2]$ &  $x^{2-\alpha}$ &  Eq.~\ref{eq:ada_divergence_pc} \\
\hline
\end{tabular}
\label{tbl:ada_edm_family}
\end{sc}
\end{center}
\vskip -12pt
\end{table*}

\begin{figure*}[t!]
  \begin{center}
      \begin{subfigure}[h]{0.43\textwidth}
        \includegraphics[width=\textwidth]{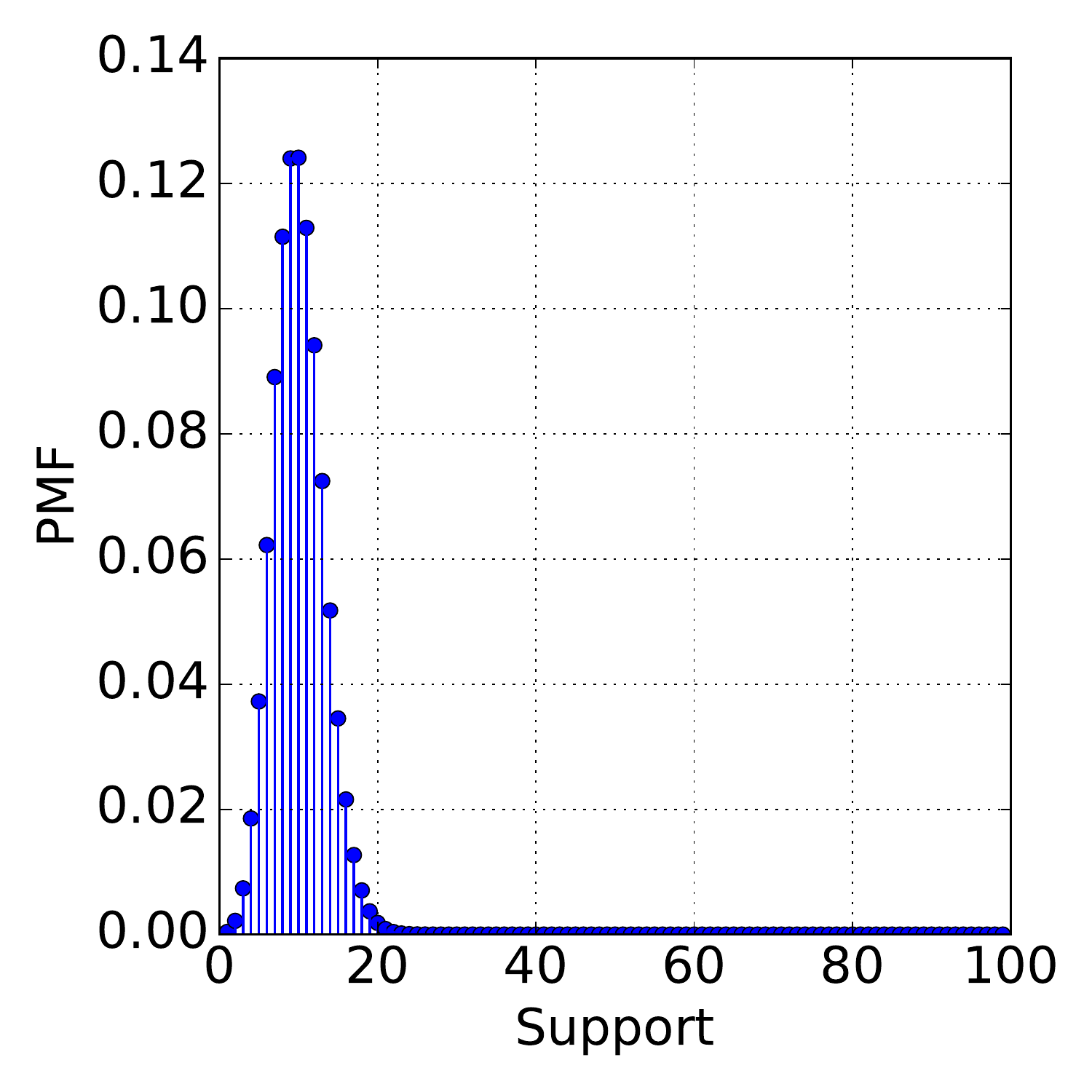}
        \caption{}
        \label{fig:ada_nnd_000}
      \end{subfigure}
      ~
      \begin{subfigure}[h]{0.43\textwidth}
        \includegraphics[width=\textwidth]{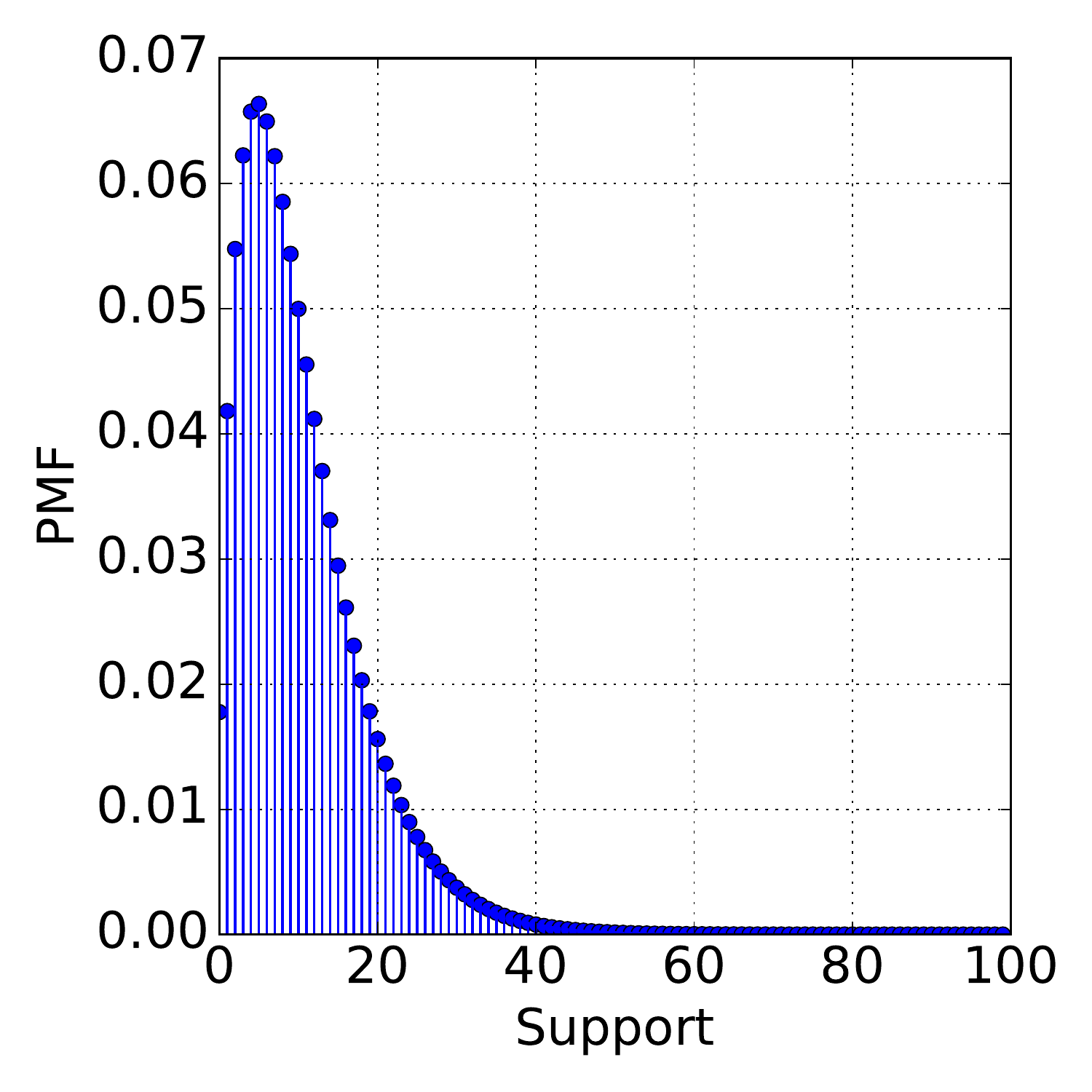}
        \caption{}
        \label{fig:ada_nnd_500}
      \end{subfigure}

      \begin{subfigure}[h]{0.43\textwidth}
        \includegraphics[width=\textwidth]{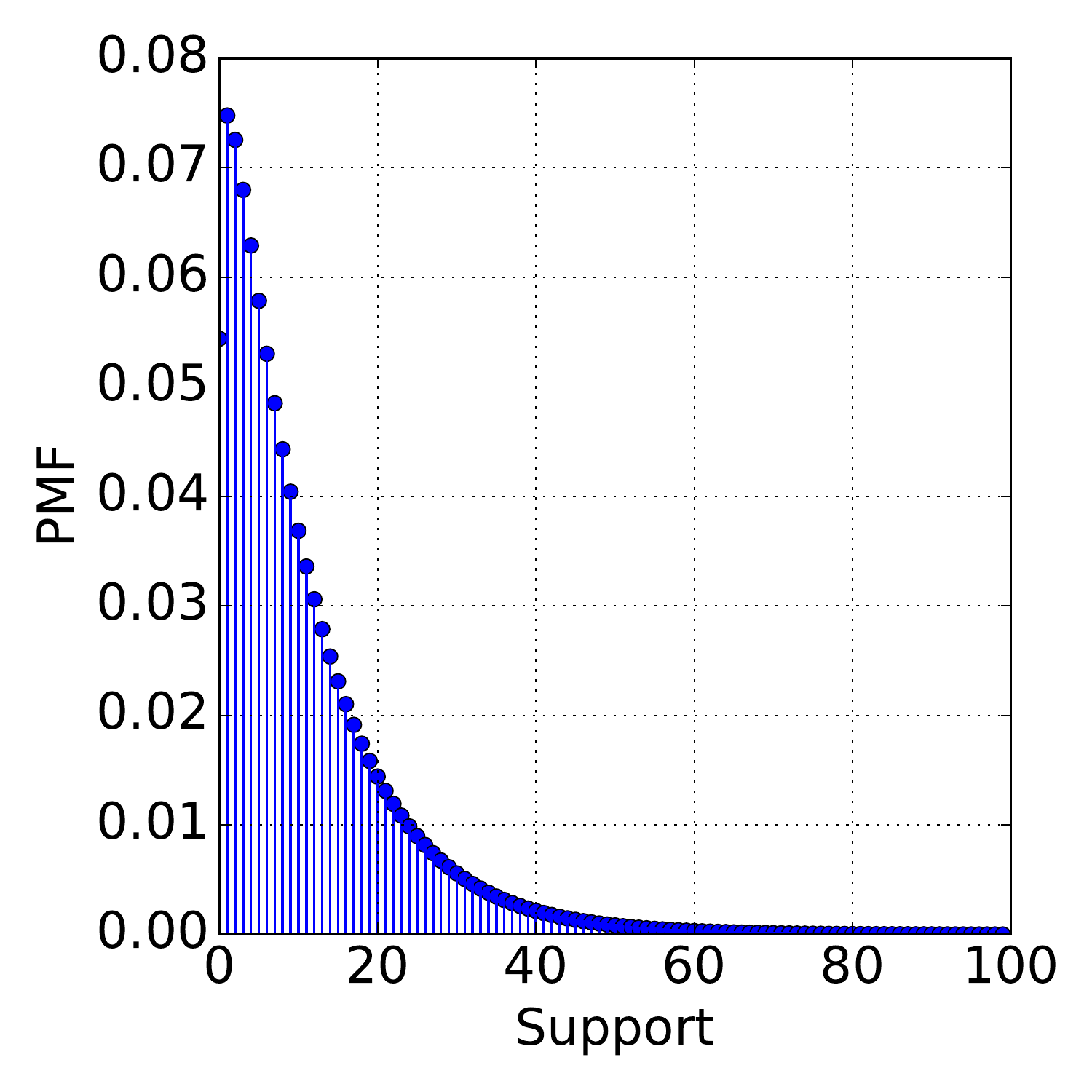}
        \caption{}
        \label{fig:ada_nnd_1000}
      \end{subfigure}
      ~
      \begin{subfigure}[h]{0.43\textwidth}
        \includegraphics[width=\textwidth]{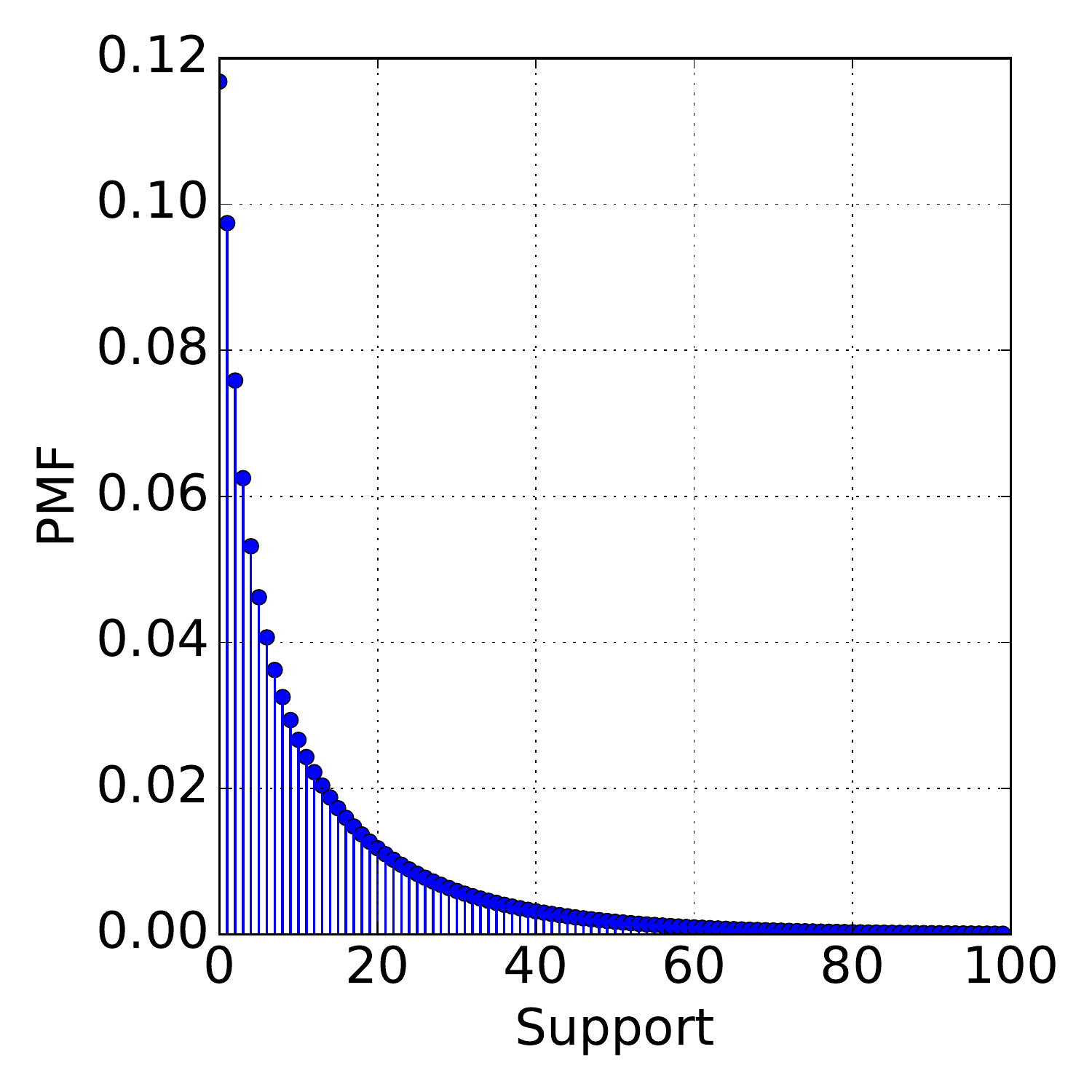}
        \caption{}
        \label{fig:ada_nnd_2000}
      \end{subfigure}
      \caption{{\bf Density of EDMs with support on the natural numbers.} Probability mass function of four members of the proposed EDM class for non-negative discrete data with $\mu=10$ and $\kappa=1$ when the hyper-parameter is set to a) $\alpha=0$ (Poisson), b) $\alpha=0.5$, c) $\alpha=1$ (negative binomial), d) $\alpha=2$}
      \label{fig:ada_nnd_pmf}
  \end{center}
\end{figure*}
\subsection{EDM class for non-negative discrete data}
\label{subsec:ada_edm_nnd}
The count data is often modeled with a Poisson or negative binomial distribution~\citep{bliss1953fitting,hilbe2011negative}. The support for the count data is unbounded, unlike binomial data. We start with the following quadratic unit variance function $\upsilon(x\mid \alpha) = x(1+ \alpha x)$ with $\alpha \in \mathbb{R}_{0}$. Solving the implied second order differential equation, $\Psi''(\theta) = \Psi'(\theta)(1+ \alpha \Psi'(\theta))$, we get
\begin{align*}
\Psi(\theta \mid \alpha)=\begin{cases}
-\frac{1}{\alpha}\log(1 - \alpha e^{\theta}) & \alpha > 0\\
e^{\theta} & \alpha = 0
\end{cases}
\end{align*}
with $\Theta = dom\;\Psi$ given by
\begin{align*}
\Theta = \begin{cases}
(-\infty,-\log \alpha )& \alpha >0\\
\mathbb{R}& \alpha = 0.
\end{cases}
\end{align*}
Since $\Theta$ is open, we get a regular exponential family. The mean-value mapping is given by
\begin{align*}
\tau(\theta \mid \alpha) &=\frac{e^{\theta}}{1-\alpha e^{\theta}}.
\end{align*}
The range of the mean-value mapping is then $\Omega = (0,\infty)$. The convex conjugate of $\Psi$ is given by
\begin{align*}
\phi(t\mid \alpha) &=\begin{cases}
t\log{t} -\frac{\alpha t +1}{\alpha}\log(\alpha t + 1) & \alpha >0\\
t\log{t} - t & \alpha = 0
\end{cases}
\end{align*}
with support $S = \mathbb{Z}_{0}$ and convex support $C = \mathbb{R}_{0}$. From Theorem~\ref{th:ada_steep_bregman}, we know that $\phi$ generates a Bregman divergence. The parametrized Bregman divergence, $d_{\phi}(x,y \mid \alpha) : \mathbb{R}_{0} \times \mathbb{R}_{+} \rightarrow \mathbb{R}_{0}$, is given by
\begin{align}
d_{\phi}(x,y \mid \alpha) &=\begin{cases}
\frac{1}{\alpha}\log(\alpha y +1) & \alpha >0, x = 0\\
(\frac{1}{\alpha} + x )\log(\frac{\alpha y +1}{\alpha x +1}) + x\log(\frac{x}{y})& \alpha >0, x \neq 0\\
y& \alpha = 0, x = 0\\
y - x + x\log(\frac{x}{y})& \alpha = 0, x \neq 0.
\end{cases}
\label{eq:ada_divergence_nnd}
\end{align}
Two members of this class are the Poisson ($\alpha = 0$) and the negative binomial ($\alpha = 1$) distributions. Having specified the variance function and the corresponding Bregman divergence, we can use the saddle-point approximation (Eq.~\ref{eq:ada_saddle_discrete}) to express the density in the dual domain. The probability mass functions for four members of this class, including Poisson and negative binomial, are depicted in Fig.~\ref{fig:ada_nnd_pmf}. Recall that the variance of a given steep EDM, $p_{\phi}(x\mid \mu,\kappa, \alpha)$, is $\kappa \mu + \alpha\kappa \mu^2$ (see Eq.~\ref{eq:ada_scaled_variance}). We see that, as $\alpha$ increases, the variance increases monotonically. The level of over-dispersion is therefore captured by the hyper-parameter $\alpha$. Distributions with higher $\alpha$ are useful for the analysis of over-dispersed count data.

\begin{figure*}[t!]
  \begin{center}
      \begin{subfigure}[h]{0.43\textwidth}
        \includegraphics[width=\textwidth]{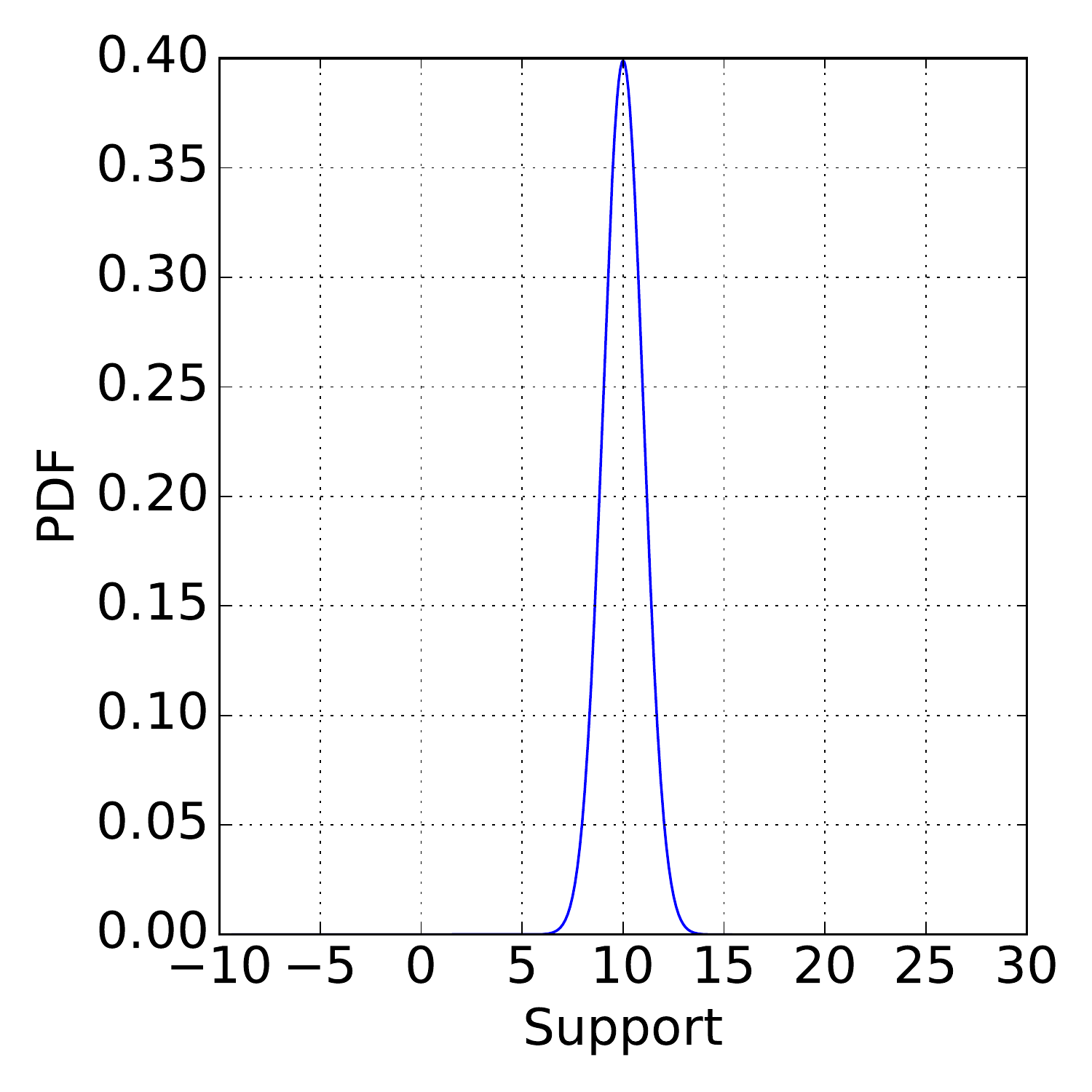}
        \caption{}
        \label{fig:ada_rc_000}
      \end{subfigure}
      ~
      \begin{subfigure}[h]{0.43\textwidth}
        \includegraphics[width=\textwidth]{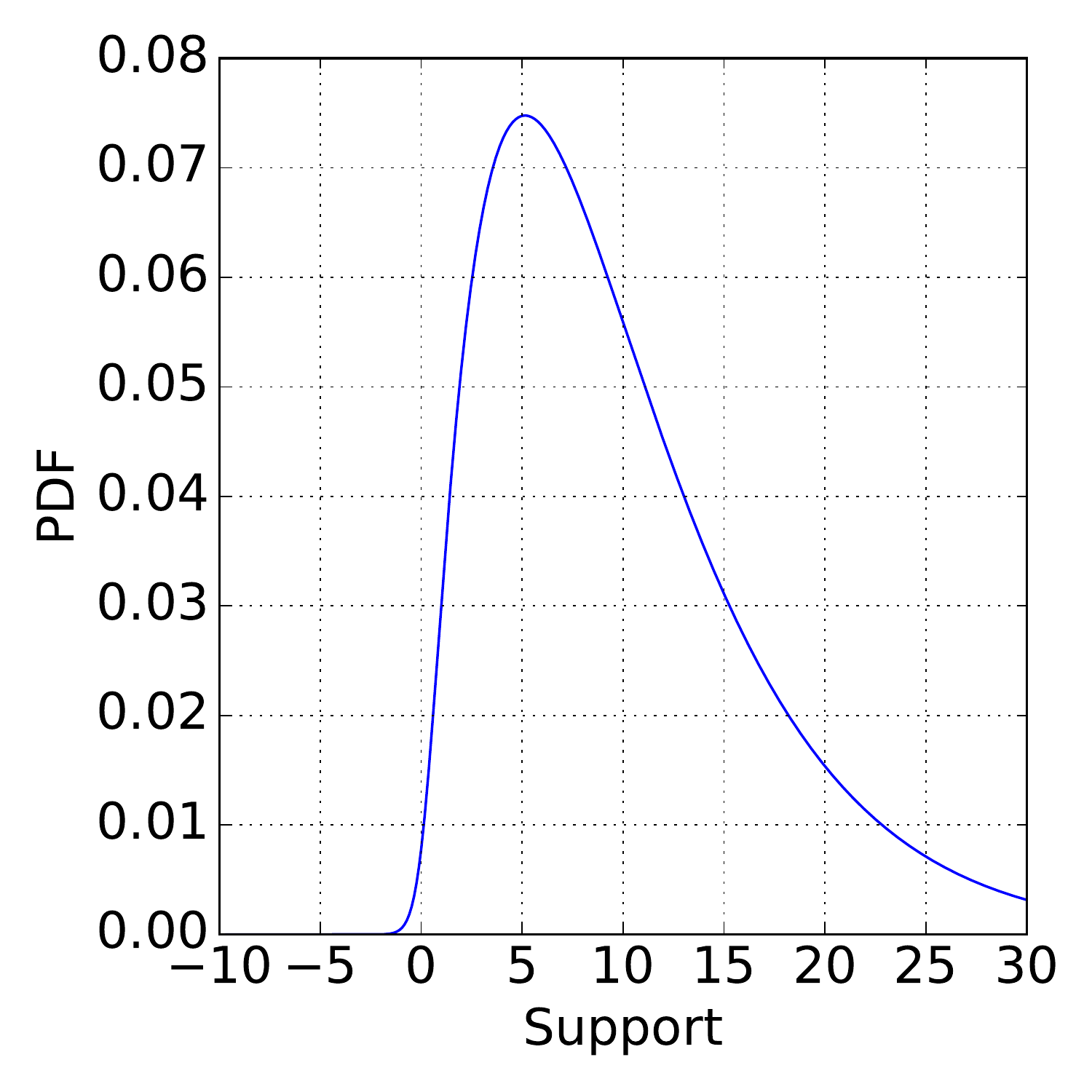}
        \caption{}
        \label{fig:ada_rc_500}
      \end{subfigure}

      \begin{subfigure}[h]{0.43\textwidth}
        \includegraphics[width=\textwidth]{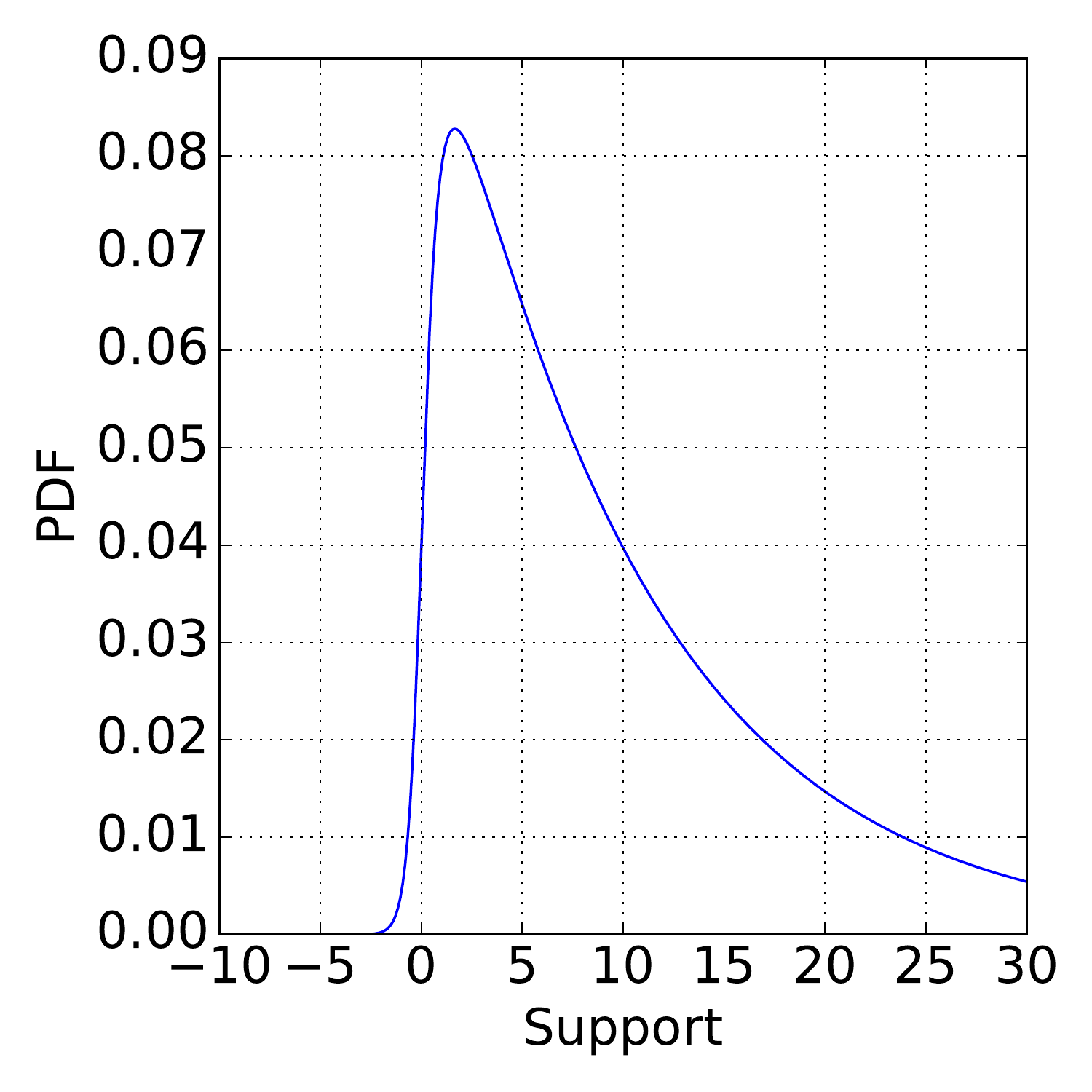}
        \caption{}
        \label{fig:ada_rc_1000}
      \end{subfigure}
      ~
      \begin{subfigure}[h]{0.43\textwidth}
        \includegraphics[width=\textwidth]{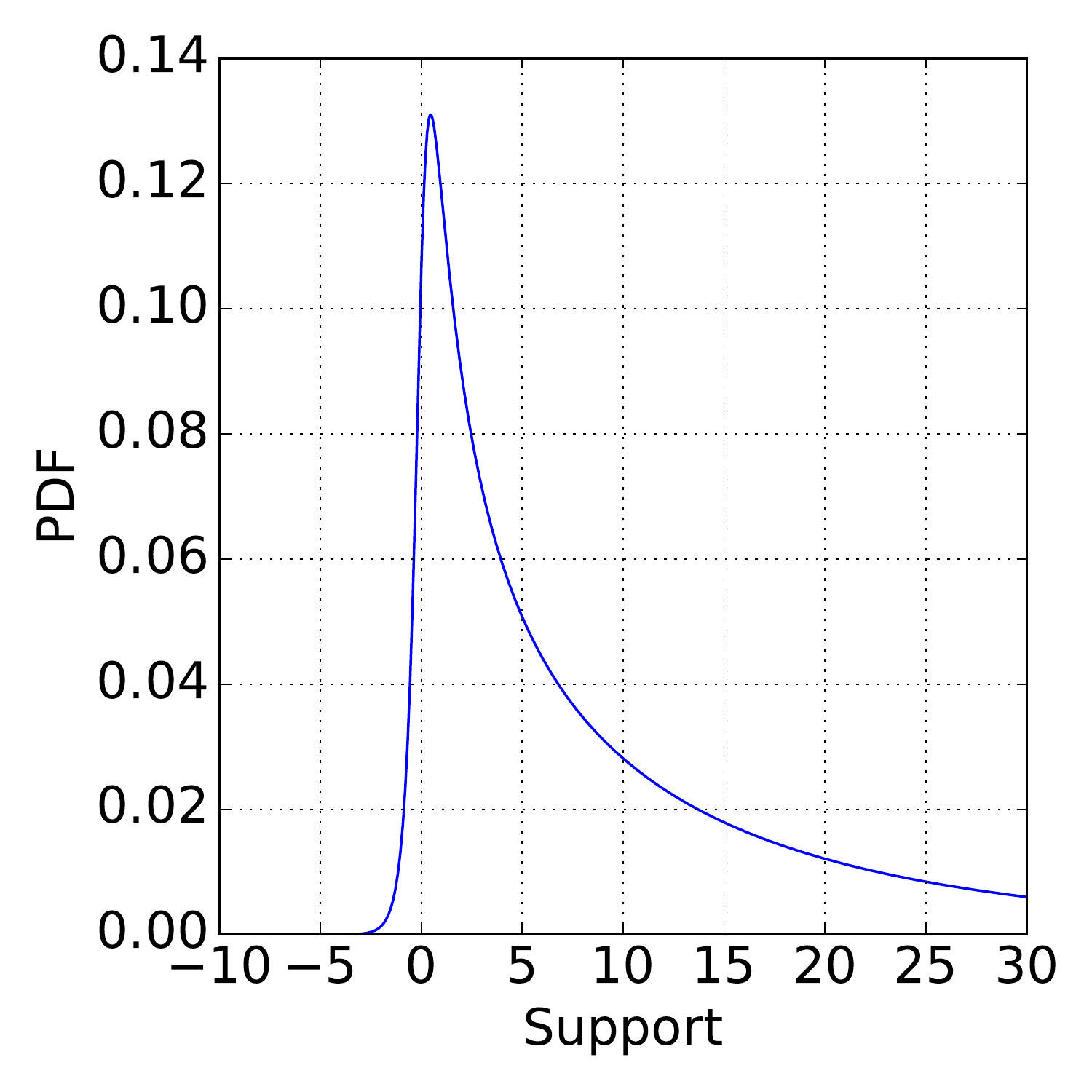}
        \caption{}
        \label{fig:ada_rc_2000}
      \end{subfigure}
      \caption{{\bf Density of EDMs with support on the real line.} Probability distribution function for four members of the proposed EDM class for continuous data on $\mathbb{R}$ with $\mu=10$ and $\kappa=1$ when the hyper-parameter is set to a) $\alpha=0$ (Gaussian), b) $\alpha=0.5$, c) $\alpha=1$ (GHS), d) $\alpha=2$}
      \label{fig:ada_rc_pdf}
  \end{center}
  \vspace{-12pt}
\end{figure*}
\subsection{EDM class for continuous data on the real line}
\label{subsec:ada_edm_rc}
The natural choice to model continuous data on $\mathbb{R}$ is the Gaussian distribution. The Gaussian distribution is symmetric around the mean and has a variance term independent of the mean term. There are cases---especially in financial applications---where the underlying distribution of the data is asymmetric around the mean~\citep{fischer2013generalized}. To better model such data, the generalized hyperbolic secant (GHS) distribution may be used~\citep{harkness1968generalized}.

Another use of GHS is to model data on the unit interval, $S=(0,1)$. The logit transform, $\log(x/(1-x))$, of a beta-distributed random variable has a GHS distribution~\citep[p.101]{johnson1949systems,jorgensen1997theory}. The beta distribution itself is not an EDM, and in fact there is no EDM with support on the unit interval~\citep[Chapter~3]{jorgensen1997theory}. GHS allows us to model data on the unit interval through the logit transform. Similarly, the logit transform of a logit-normal distributed random variable has a Gaussian distribution.

In this section, we propose a class of EDMs that includes Gaussian and hyperbolic secant distributions as members. We start with the quadratic variance function $\upsilon(x\mid \alpha) = 1 + \alpha x^2$ with $\alpha \in \mathbb{R}_{0}$. Solving the implied second order differential equation, $\Psi''(\theta) = 1+ \alpha \Psi'(\theta)^2$, we get the following log-partition function
\begin{align*}
\Psi(\theta\mid \alpha)=\begin{cases}
-\frac{1}{\alpha}\log(\cos(\sqrt{\alpha}\theta))& \alpha > 0\\
\frac{1}{2}\theta^2 & \alpha = 0
\end{cases}
\end{align*}
with $\Theta = dom\;\Psi$ given by
\begin{align*}
\Theta = \begin{cases}
(-\frac{\pi}{2\sqrt{\alpha}},\frac{\pi}{2\sqrt{\alpha}})& \alpha >0\\
\mathbb{R}& \alpha = 0.
\end{cases}
\end{align*}
Since $\Theta$ is open, we get a regular (and thus steep) exponential family distribution for every $\alpha \in \mathbb{R}_{0}$. The mean-value mapping is given by
\begin{align*}
\tau(\theta\mid \alpha) &=\begin{cases}
\frac{1}{\sqrt{\alpha}}\tan(\sqrt{\alpha}\theta)& \alpha >0\\
\theta& \alpha = 0.
\end{cases}
\end{align*}
The range of the mean-value mapping is then $\Omega = \mathbb{R}$. The convex conjugate of $\Psi$ is given by
\begin{align*}
\phi(t\mid \alpha) &=\begin{cases}
2\sqrt{\alpha}t \arctan(\sqrt{\alpha}t) - \frac{1}{2\alpha}\log(1+\alpha t^2) & \alpha >0\\
\frac{1}{2}t^2& \alpha = 0
\end{cases}
\end{align*}
with convex support $C = S = \mathbb{R}$. From Theorem~\ref{th:ada_steep_bregman}, we conclude that $\phi$ generates a Bregman divergence, $d_{\phi}(x,y \mid \alpha) : \mathbb{R} \times \mathbb{R} \rightarrow \mathbb{R}_{0}$, which is given by
\begin{align}
d_{\phi}(x,y \mid \alpha) &=\begin{cases}
\frac{1}{2\alpha}\left(2\sqrt{\alpha}x\left(\arctan{(\sqrt{\alpha}x)}-\arctan{(\sqrt{\alpha}y)}\right) + \log{\frac{1+\alpha y^2}{1+\alpha x^2}}\right) & \alpha >0\\
\frac{1}{2}(x-y)^2& \alpha = 0.
\end{cases}
\label{eq:ada_divergence_rc}
\end{align}
Two noteworthy members of this class are the Gaussian ($\alpha = 0$) and generalized hyperbolic secant ($\alpha = 1$) distributions. We can use the saddle-point approximation (Eq.~\ref{eq:ada_saddle_cont}) with the variance function and the Bregman divergence above to express the density of each member of the class. In Fig.~\ref{fig:ada_rc_pdf}, the probability distribution functions for several members of this class, including Gaussian and GHS, are shown. The variance of a given steep EDM, $p_{\phi}(x\mid \mu,\kappa,\alpha)$, in this class is $\kappa + \alpha \kappa \mu^2$. As $\alpha$ increases, we see that the asymmetric shape becomes more apparent. Similar to the case for the over-dispersed count data, increasing $\alpha$ implies a higher variance for a fixed mean parameter.

\subsection{The EDM class for positive continuous data}
\label{subsec:ada_edm_pc}
Tweedie is another prominent class of EDMs that includes gamma, inverse-Gaussian, Poisson, and Gaussian distributions~\citep{tweedie1947functions}. The seminal work of~\citep{bar1986reproducibility} gives a comprehensive account of the Tweedie class. In this section, we reiterate previous results in the context of our framework and establish the connection between the Tweedie class and Bregman divergences.

Suppose we have a power variance function $\upsilon(x\mid\alpha) = x^{2-\alpha}$ with $\alpha \in \mathbb{R}$. Later we'll see that not every choice of $\alpha$ yields a steep EDM; therefore, we will choose a smaller set for $\mathcal{A}$. After solving $\Psi''(\theta) = (\Psi'(\theta))^{2-\alpha}$, we obtain the following log-partition function
\begin{align}
\Psi(\theta \mid \alpha)=\begin{cases}
\frac{1}{\alpha}(((\alpha-1)\theta+1)^{\frac{\alpha}{\alpha-1}}-1)& \alpha \in \mathbb{R}\setminus\{0,1\}\\
e^{\theta}-1& \alpha = 1\\
-\log{(1-\theta)} & \alpha = 0
\end{cases}
\end{align}
with $\Theta$ given by
\begin{align}
\Theta = \begin{cases}
(-\infty,\frac{1}{1-\alpha}]& \alpha \in (-\infty,0)\\
(-\infty,\frac{1}{1-\alpha})& \alpha \in [0,1)\\
[\frac{1}{1-\alpha},\infty)& \alpha \in (1,\infty)\setminus\{2\}\\
\mathbb{R}& \alpha \in \{1,2\}.
\end{cases}
\end{align}
Note that $\Psi$ is continuous in $\alpha$, including the limiting cases for $\alpha=0$ and $\alpha=1$. When $\alpha \in [0,1] \cup \{2\}$, the corresponding distributions belong to the regular exponential family. Similarly, for $\alpha \in (-\infty,1] \cup \{2\}$, the corresponding distributions are steep. The mean-value mapping is given by
\begin{align*}
\tau(\theta \mid \alpha) &=\begin{cases}
((\alpha-1)\theta+1)^{\frac{1}{\alpha-1}}& \alpha \neq 1\\
e^{\theta}& \alpha = 1
\end{cases}
\end{align*}
with range $\Omega = (0,\infty)$, except when $\alpha=2$ where $\Omega = \mathbb{R}$. The convex conjugate is given by
\begin{align}
\phi(t\mid \alpha) = \begin{cases}
\frac{1}{\alpha(\alpha-1)}(t^{\alpha}-\alpha t + \alpha -1)& \alpha \in \mathbb{R}\setminus\{0,1\}\\
t\log{t} -t +1& \alpha = 1\\
t -\log{t} - 1 & \alpha = 0
\end{cases}
\end{align}
with convex support $C = [0,\infty)$ when $\alpha \in (0,\infty)\setminus \{2\}$, $C = (-\infty,\infty)$ when $\alpha=2$, and $C = (0,\infty)$ when $\alpha \in (-\infty,0]$. When $1<\alpha<2$, then $\Psi''(\frac{1}{1-\alpha}) = 0$, leading to a degenerate distribution. At the expense of having a non-full exponential family, we limit the convex support to be $C = (0,\infty)$ when $1<\alpha<2$. With this choice, the domain of $\Psi$ is limited to $\Theta = (\frac{1}{1-\alpha},\infty)$ when $1<\alpha<2$. The corresponding divergence is given by
\begin{align}
d_{\phi}(x,y \mid \alpha)=\begin{cases}
\frac{x^{\alpha}+(\alpha-1)y^{\alpha}-\alpha x y^{\alpha-1}}{\alpha(\alpha-1)} & \alpha \in \mathbb{R}\setminus{\{0,1\}}\\
x(\log{x} - \log{y}) + (y-x)& \alpha = 1\\
\frac{x}{y} - \log{(\frac{x}{y})} - 1 & \alpha = 0.
\end{cases}
\label{eq:ada_divergence_pc}
\end{align}
As a corollary of Theorem~\ref{th:ada_steep_bregman}, we conclude that $d_{\phi}(x,y \mid \alpha)$ is a Bregman divergence when $\alpha \in (-\infty,1] \cup \{2\}$. For the first time in the literature, we explicitly establish the connection between Tweedie class and Bregman divergences. Previously, an indirect relationship is explored through a related class of divergences, $\beta$-divergences~\citep{basu1998robust,cichocki2006csiszar}. In $\beta$-divergences, the parameter $\alpha$ corresponds to parameter $\beta$ and the rest of the formulation is the same except that the domain is restricted to $\mathbb{R}_{+}$. The relationship between Bregman divergences and $\beta$-divergences was established in ~\citep{hennequin2011beta}. The connection between $\beta$-divergences and the Tweedie class has been explored in~\citep{yilmaz2012alpha}. We note that the definition of $\beta$-divergences are slightly different in these two works.

When the support is $S = (0,\infty)$, we choose $\mathcal{A} = (-\infty,0]$, which includes the gamma ($\alpha~=~0$) and inverse-Gaussian ($\alpha=-1$) distributions. In fact, it is possible to extend $\mathcal{A}$ to $(-\infty,2]$, assuming the probability mass at $0$ is small for $0<\alpha<2$ and the probability of the interval $(-\infty,0]$ is small for $\alpha=2$. We note that $\mathcal{A} = (-\infty,2]$ includes Poisson ($\alpha = 1$) and Gaussian ($\alpha = 2$) distributions as well. At first, using the Poisson distribution to model positive continuous data looks unusual; however, the saddle-point approximation in Eq.~\ref{eq:ada_saddle_cont} requires only that the set-of-interest is a subset of the convex support and the variance function is positive. Both of these conditions are satisfied since $\mathbb{R}_{+} \subset C = \mathbb{R}_{0}$ and $\upsilon(x\mid\alpha=1) = x > 0~\forall x \in \mathbb{R}_{+}$.

\paragraph{Non-negative continuous data.}\label{subsec:ada_edm_nnc} Zero-inflated distributions are often modeled with compound-Poisson distributions~\citep{basbug2016hierarchical}. Within the Tweedie class, compound-Poisson-Gamma distributions correspond to the hyper-parameter range $0<\alpha<1$. Similar to positive continuous data, the divergence corresponding to the Poisson distribution can be used for non-negative continuous data, which makes the feasible hyper-parameter domain $\mathcal{A} = (0,1]$. One important caveat is that, when using the saddle-point approximation for the point mass at zero, we must use Eq.~\ref{eq:ada_saddle_discrete} instead of Eq.~\ref{eq:ada_saddle_cont}. 

\begin{figure*}[t!]
  \begin{center}
      \begin{subfigure}[h]{0.32\textwidth}
        \includegraphics[width=\textwidth]{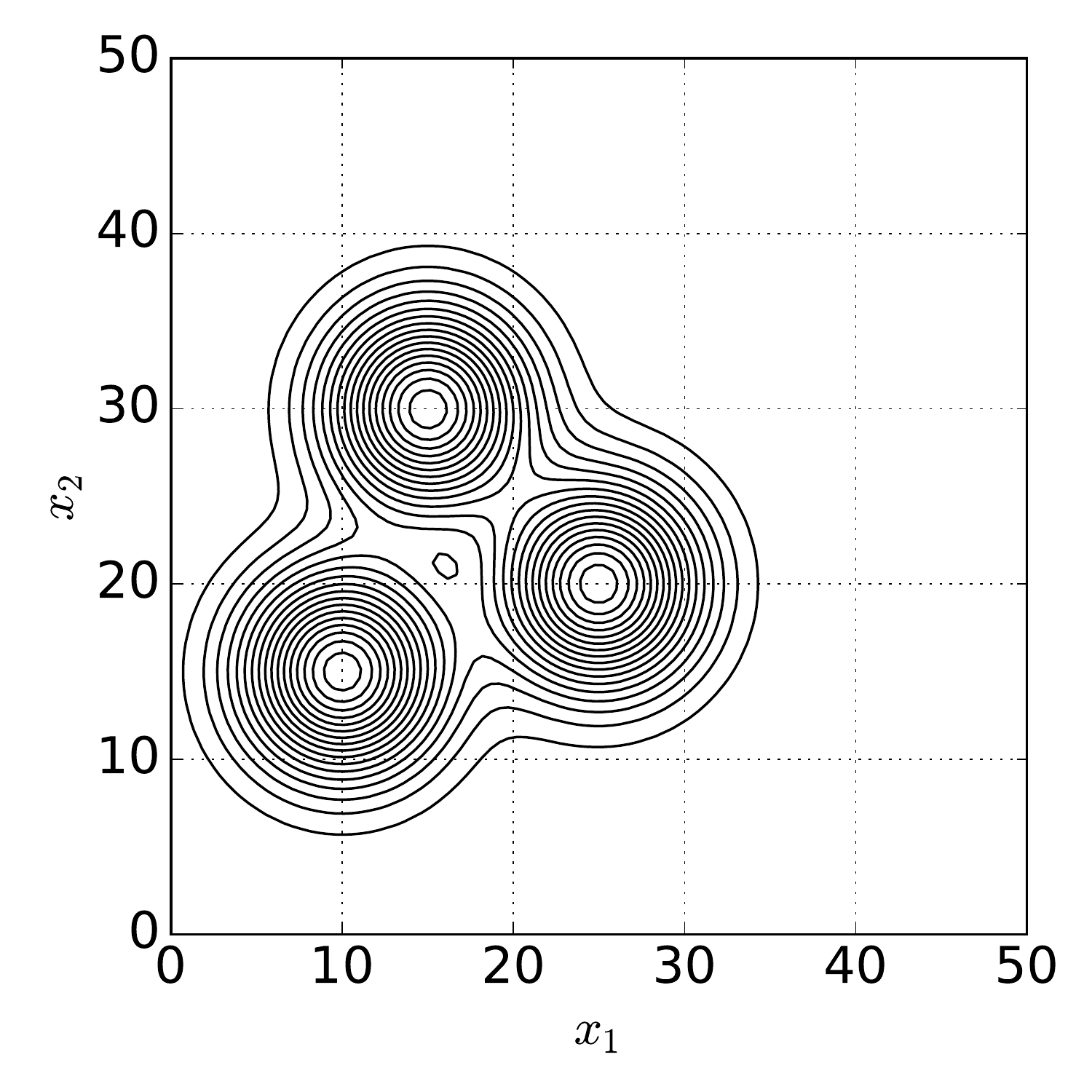}
        \caption{}
        \label{fig:ada_contour_homo_normal}
      \end{subfigure}
      ~
      \begin{subfigure}[h]{0.32\textwidth}
        \includegraphics[width=\textwidth]{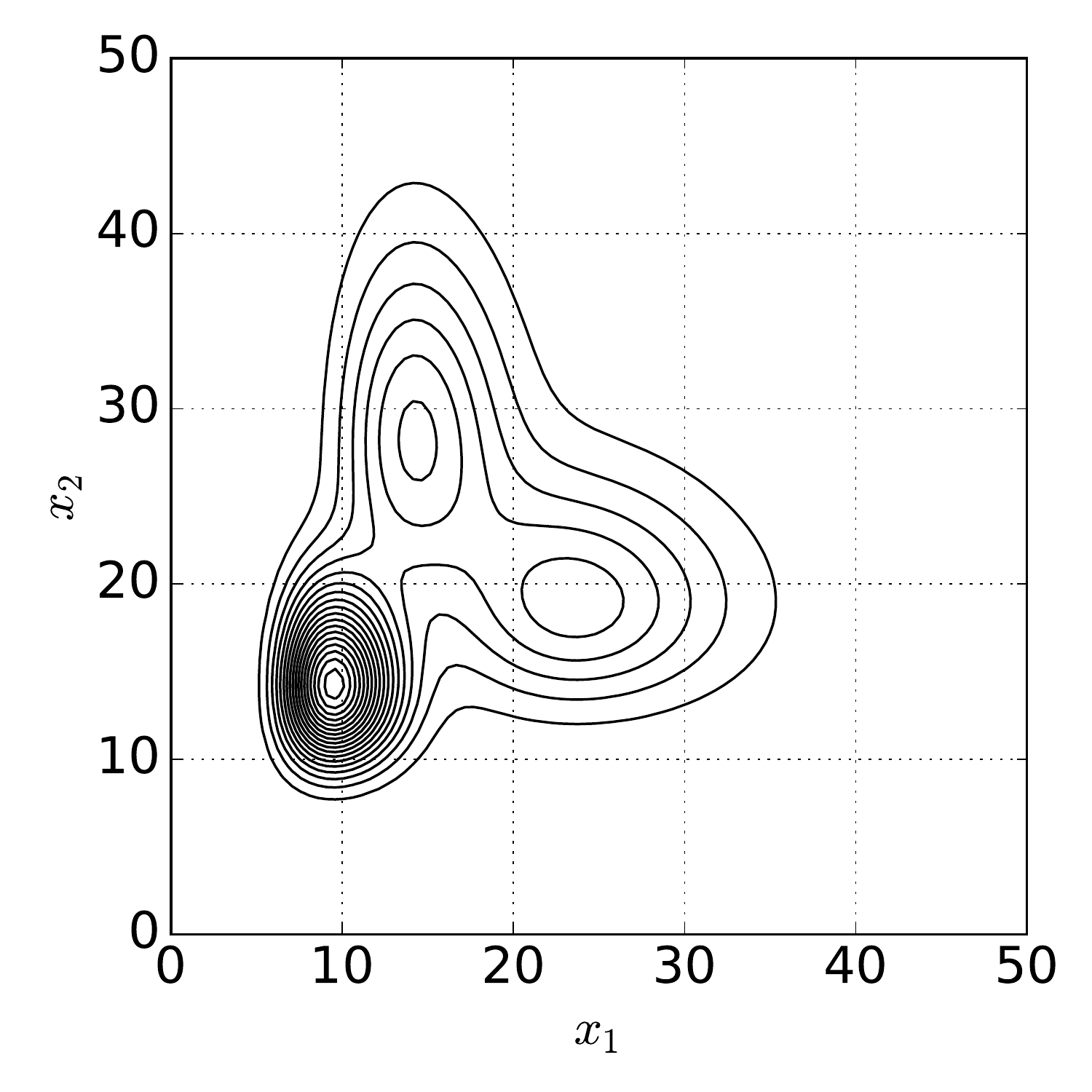}
        \caption{}
        \label{fig:ada_contour_homo_gamma}
      \end{subfigure}
      ~
      \begin{subfigure}[h]{0.32\textwidth}
        \includegraphics[width=\textwidth]{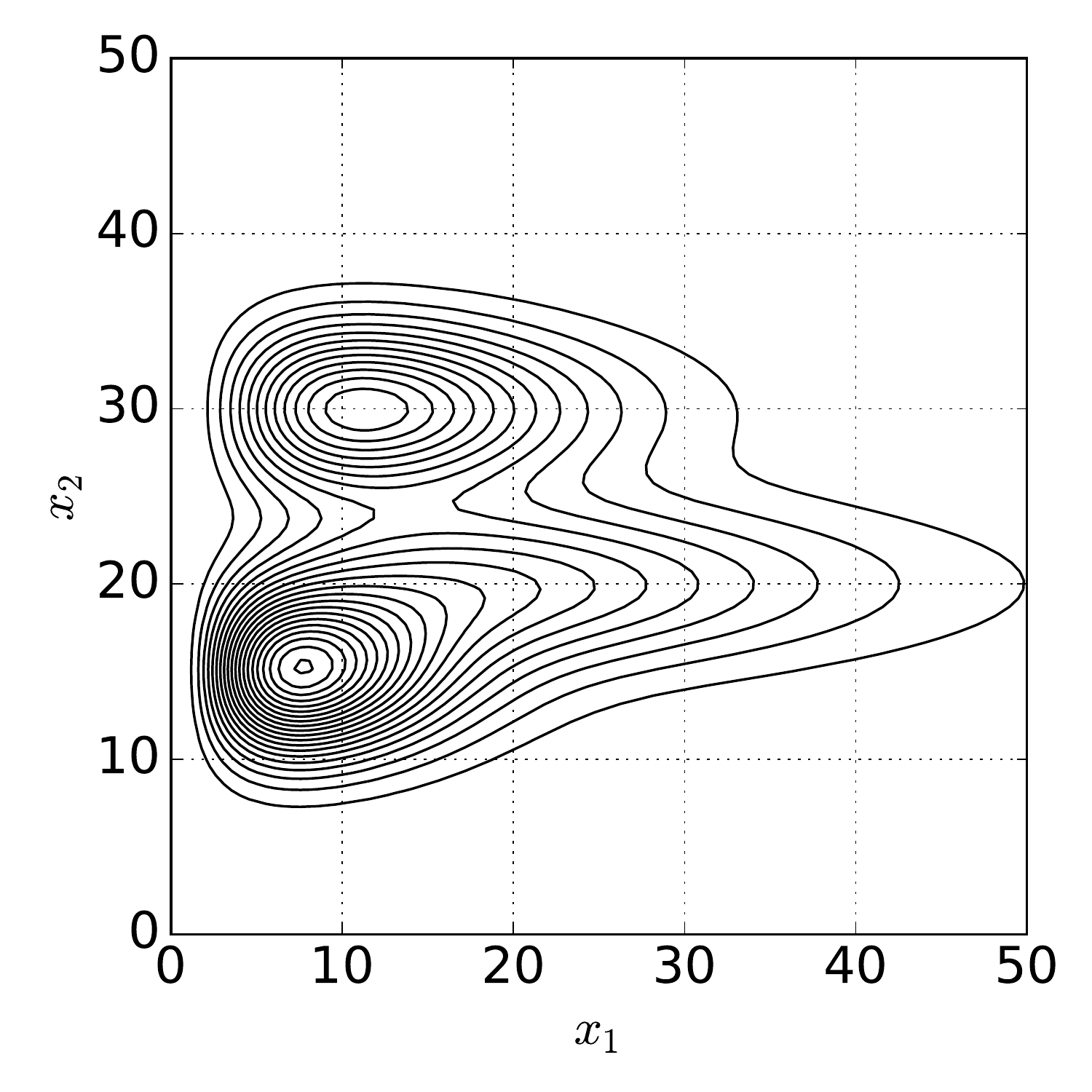}
        \caption{}
        \label{fig:ada_contour_heto_mixed}
      \end{subfigure}

      \begin{subfigure}[h]{0.32\textwidth}
        \includegraphics[width=\textwidth]{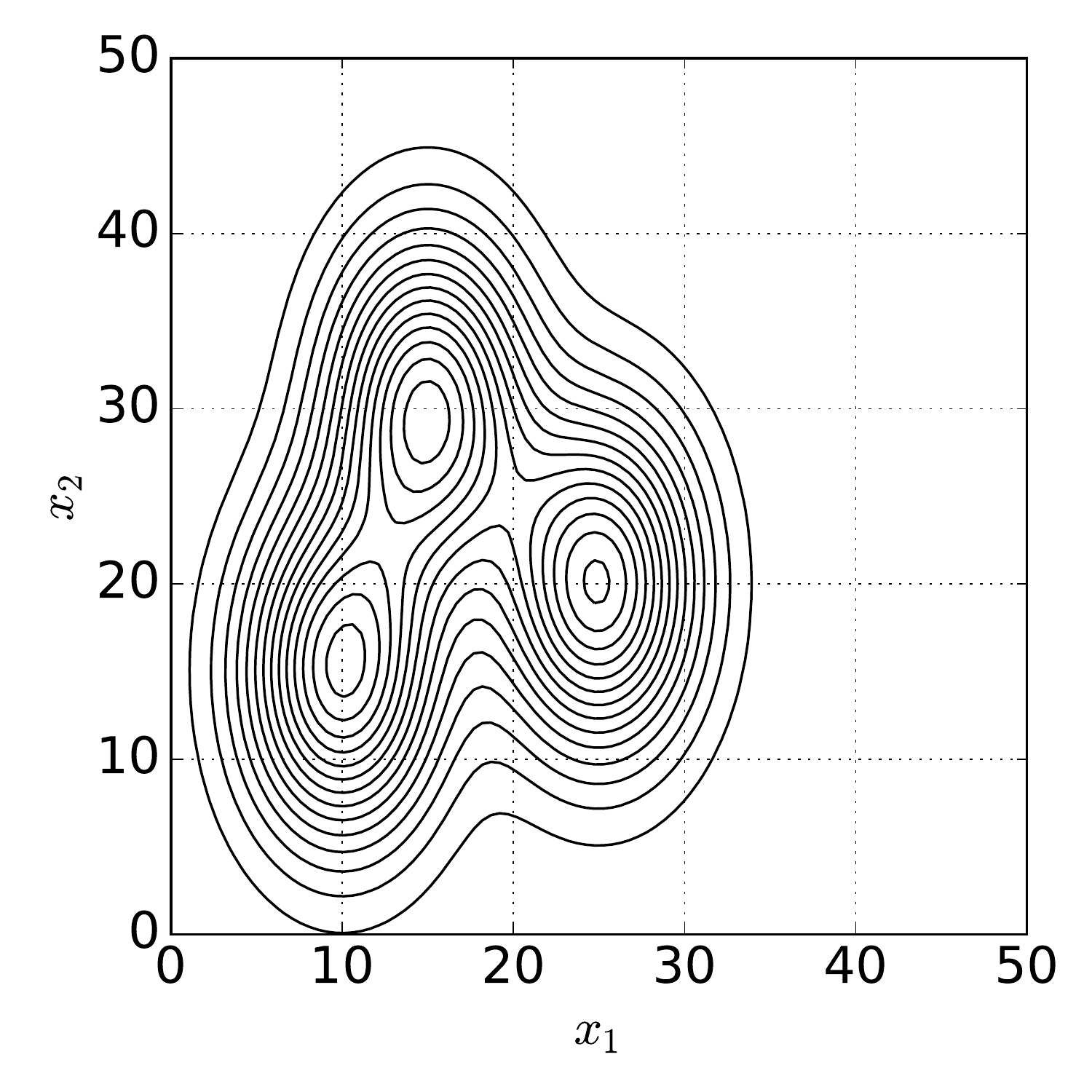}
        \caption{}
        \label{fig:ada_contour_heto_kappa_normal}
      \end{subfigure}
      ~
      \begin{subfigure}[h]{0.32\textwidth}
        \includegraphics[width=\textwidth]{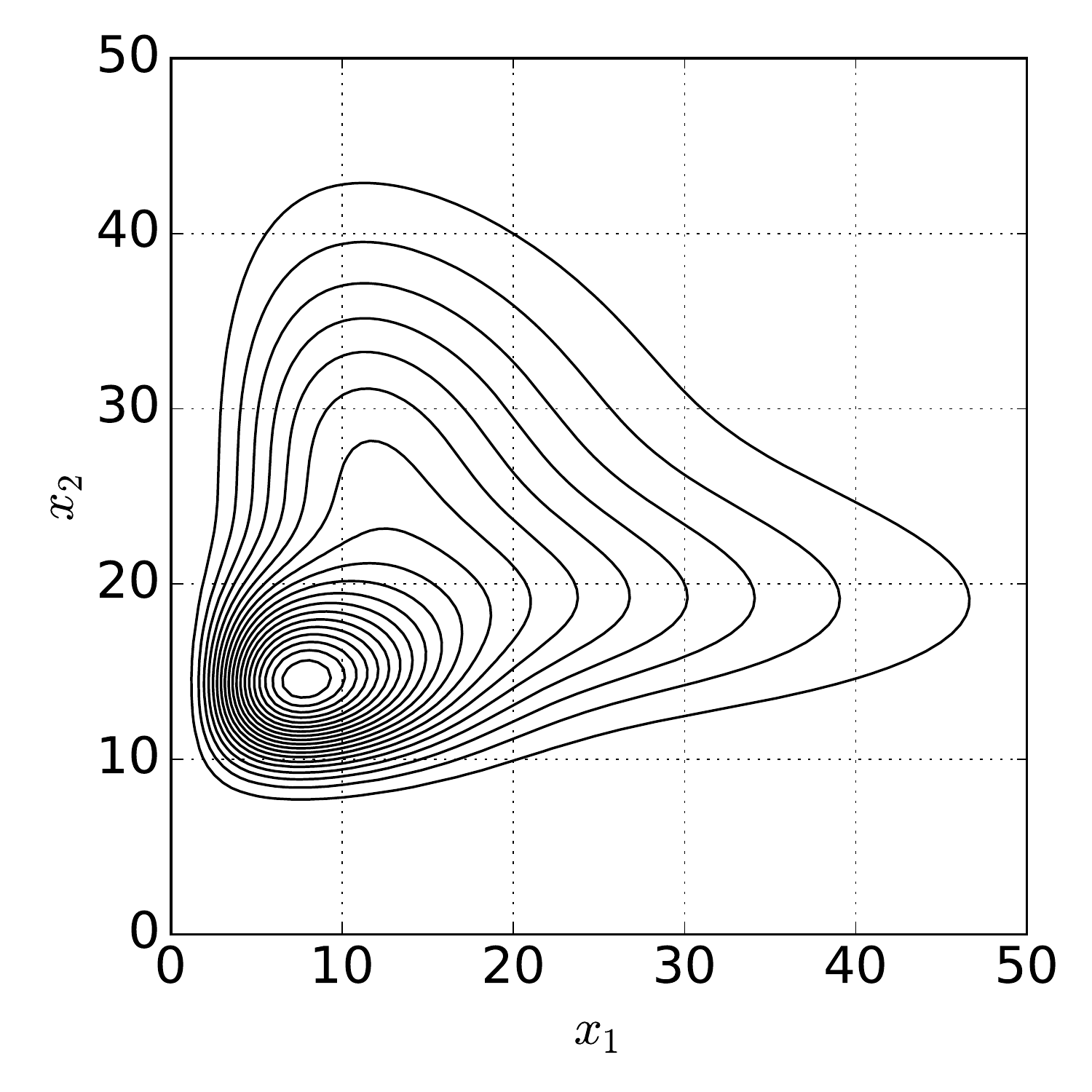}
        \caption{}
        \label{fig:ada_contour_heto_kappa_gamma}
      \end{subfigure}
      ~
      \begin{subfigure}[h]{0.32\textwidth}
        \includegraphics[width=\textwidth]{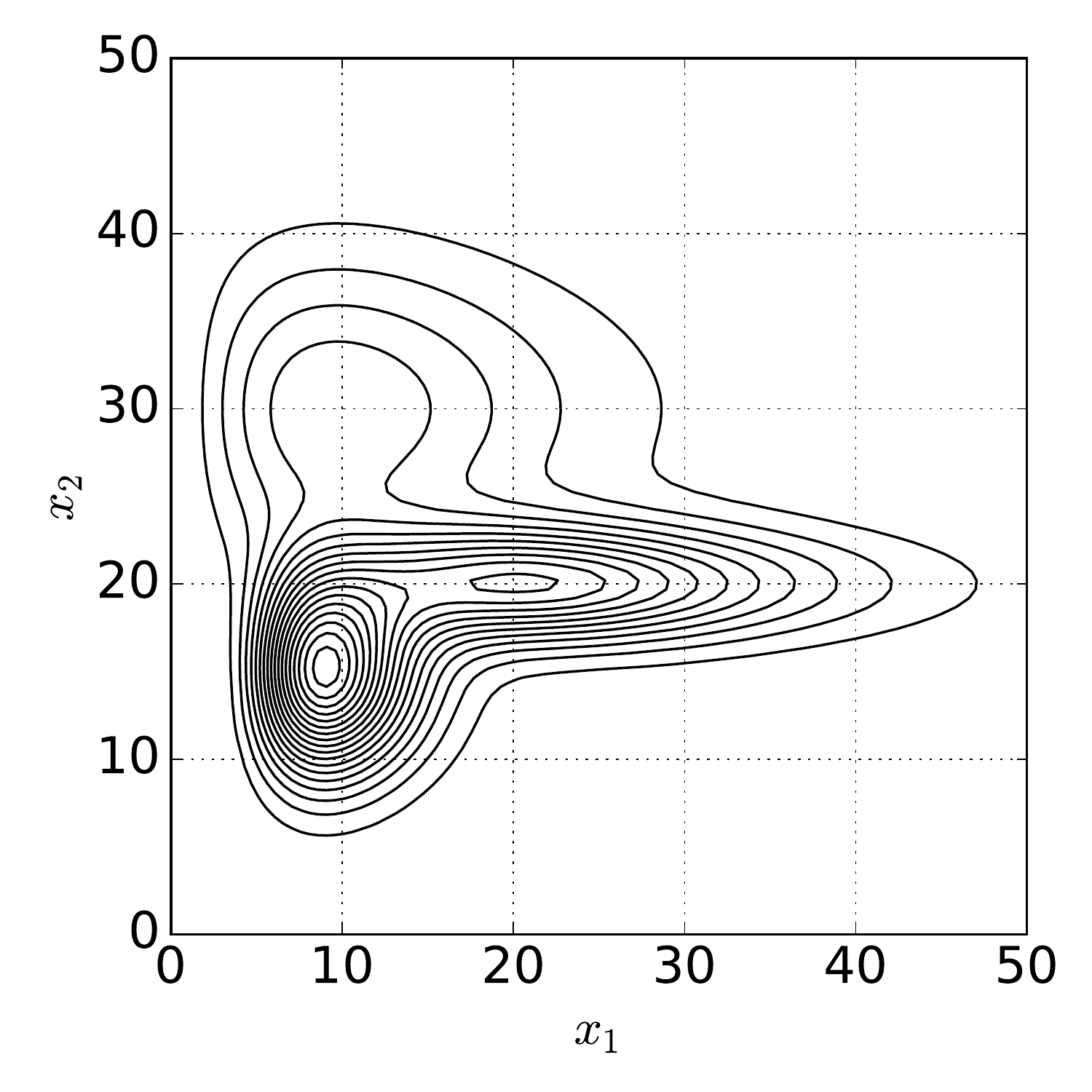}
        \caption{}
        \label{fig:ada_contour_heto_all}
      \end{subfigure}
      \caption{{\bf Homogeneous and heterogeneous mixture models.} Contour plots for various mixture models with fixed centroids located at (10,15), (25,20) and (15,30). Homogeneous mixture model with fixed dispersion across dimensions and mixture components and fixed Gaussian topology in both dimensions (a) and gamma topology in both dimensions (b); heterogeneous mixture model with dispersion varying in the first and second dimensions but fixed across mixture components and with gamma topology in the first dimension and Gaussian topology in the second dimension (c), Gaussian topology in both dimensions (d) and gamma topology in both dimensions (e); heterogeneous mixture model with dispersion varying in the first and second dimension and across the three mixture components and gamma topology in the first dimension and Gaussian topology in the second dimension (f). Bregman soft clustering can be used for (a) and (b), AdaCluster can handle (a-e). The most arbitrary form of heterogeneous mixture models (f) cannot be captured by either of the algorithms.}
      \label{fig:ada_contour}
  \end{center}
  \vspace{-12pt}
\end{figure*}

\section{Mixture of exponential dispersion models}
\label{sec:ada_mixture}
In previous sections, we have explored parametrized classes of steep EDMs for data with different support. If we are given a set of samples drawn from a unimodal steep EDM, then we can identify the underlying topology and dispersion characteristics by estimating the hyper-parameter $\alpha$ as well as ($\mu$,$\kappa$) pair. Recall that the topology of a steep EDM is uniquely characterized with the unit base log-partition function ($\Psi$), its dual ($\phi$), the unit mean-value mapping ($\tau$) or the unit variance function ($\upsilon$) and the dispersion characteristic ($\kappa$) corresponds to the scaling of the Bregman divergence ($d_{\phi}$). In a more complicated case, we might be dealing with a set of samples from a multimodal steep EDM where $\alpha$ and $\kappa$ are shared across modes. The density estimation for such data can be done through the expectation-maximization (EM) algorithm. In a more general case, we might get samples from a multi-dimensional multimodal steep EDM. If the underlying steep EDM for each attribute is known a priori, then deriving the EM algorithm is straightforward. When we do not know the topology a priori, we can use the classes of EDMs proposed in the previous section to fit a class of mixture of steep EDMs instead. Recall that the classes in Section~\ref{sec:ada_parameterized} can be specified by the data support. In this section, we propose an EM-like adaptive algorithm that can fit a class of mixture of steep EDMs and identify the mixture components, the underlying topology and dispersion characteristics.

We call a mixture model \textit{homogeneous} if the topology and the dispersion characteristics are the same across mixture components and across attributes. In a homogeneous mixture model, the underlying distribution of all of the mixture components across attributes share the same parameters except the mean parameter (or, equivalently, the natural parameter in dual form). A simple example is the Gaussian mixture model with identity covariance matrix (Fig.~\ref{fig:ada_contour_homo_normal}). For a non-symmetric distribution, homogeneity is harder to notice visually (see Fig.~\ref{fig:ada_contour_homo_gamma} for a homogenous Gamma mixture model). In \citep{banerjee2005clustering}, the authors show that the likelihood of data under a homogeneous mixture of regular exponential family distributions can be represented in terms of the corresponding Bregman divergence. Furthermore, the authors propose an expectation-maximization algorithm---\emph{Bregman soft clustering}---consisting of closed-form updates to fit any homogeneous mixture model that can be represented in terms of a Bregman divergence. Based on the discussion in Section~\ref{subsec:ada_edm}, we know that the density of steep EDMs can be represented in terms of Bregman divergences; hence, we  conclude that the homogeneous mixture of steep EDMs can be fit to data with the Bregman soft-clustering algorithm.
\newpage
We start by describing the closed-form updates of the Bregman soft-clustering algorithm when used to fit a mixture of steep EDMs. We write the log-likelihood of a set of observations $\mathcal{X}=\{\boldsymbol{x_{i}}\}_{i=1}^{N}$ with $J$ attributes under the homogeneous mixture of steep EDM, $p_{\phi}(x\mid \mu, \kappa)$, as
\begin{align}
\log p_{\phi}(\mathcal{\boldsymbol{X}} \mid \boldsymbol{\pi},\boldsymbol{\mu},\kappa) = \sum\limits_{i=1}^{N}\log\sum\limits_{h=1}^{K}\pi_{h}\exp\left(\sum\limits_{j=1}^{J}\left(-\frac{1}{\kappa}d_{\phi}(x_{ij},\mu_{hj}) + \log g_{\phi}(x_{ij},\kappa)\right)\right)
\label{eq:ada_homogeneous_loglik}
\end{align}
where $\boldsymbol{\pi} = \{\pi_{h}\}_{h=1}^{K}$ is the mixture proportions, and $d_{\phi}(x,y)$ is the Bregman divergence generated from a fixed strictly convex function $\phi$. The underlying distribution $p_{\phi}(x\mid \mu, \kappa)$ is fixed and specified a priori.

In the E-step of the Bregman soft-clustering algorithm, the weight of each mixture component is updated using:
\begin{align}
p_{\phi}(h \mid \boldsymbol{x_{i}},\boldsymbol{\pi},\boldsymbol{\mu},\kappa) \leftarrow \frac{\pi_{h}\exp\left(-\frac{1}{\kappa}\sum_{j=1}^{J}d_{\phi}(x_{ij},\mu_{hj})\right)}{\sum_{h=1}^{K}\pi_{h}\exp\left(-\frac{1}{\kappa}\sum_{j=1}^{J}d_{\phi}(x_{ij},\mu_{hj})\right)}
\label{eq:ada_bsc_e_step}
\end{align}
for $i=1,\dots,N$ and $h=1,\dots,K$. Note that the base measure terms $\log g_{\phi}(x_{ij},\kappa)$ cancel. It is often the case that the base measure does not have a closed form. The homogeneity assumption simplifies the EM algorithm, allowing us to specify the log-partition function $\Psi$ or its conjugate $\phi$, and not the full density.

In the M-step of Bregman soft clustering, the maximum likelihood estimates of $\boldsymbol{\pi}$ and $\boldsymbol{\mu}$ are given by
\begin{align}
\pi_{h} &\leftarrow \frac{1}{N}\sum\limits_{i=1}^{N}p_{\phi}(h \mid \boldsymbol{x_{i}},\boldsymbol{\pi},\boldsymbol{\mu},\kappa)\label{eq:ada_bsc_m_step_pi} \\
\mu_{hj} &\leftarrow \frac{\sum_{i=1}^{N}p_{\phi}(h \mid \boldsymbol{x_{i}},\boldsymbol{\pi},\boldsymbol{\mu},\kappa)x_{ij}}{\sum_{i=1}^{N}p_{\phi}(h \mid \boldsymbol{x_{i}},\boldsymbol{\pi},\boldsymbol{\mu},\kappa)}
\end{align}
for $h=1,\dots,K$ and $j=1,\dots,J$. The base measure terms do not appear in the closed-form updates.

So far, we only described the Bregman soft-clustering algorithm updates for a homogenous mixture of steep EDMs and did not modify the existing algorithm of~\citep{banerjee2005clustering}. To be able to employ Bregman soft clustering, we need to specify the topology using the divergence-generating function $\phi$, the variance function $\upsilon$ or the log-partition function $\Psi$. If we have multiple candidates for the underlying distribution, we can optimize the objective function in Eq.~\ref{eq:ada_homogeneous_loglik} although it requires specification of the full density including the base-measure terms ($g_{\phi}(\cdot,\kappa)$) which often do not have closed-form expressions.

The homogeneous mixture of steep EDMs, however, can fall short in capturing the true geometric properties of the data. In data sets collected from multiple complex sources, the dispersion parameters tend to be different across attributes even when the topology is the same (Fig.~\ref{fig:ada_contour_heto_kappa_normal} and Fig.~\ref{fig:ada_contour_heto_kappa_gamma}). For an arbitrary EDM, normalizing data can be burdensome since we would have to fit an individual mixture model to each attribute. Inference gets even more complicated when the topology is different across attributes (Fig.~\ref{fig:ada_contour_heto_mixed}). Census data is a good example, where we have non-negative discrete attributes such as age, positive continuous attributes such as height, possibly negative continuous attributes such as income, and binary attributes such as gender. When the number of attributes is small, a custom mixture model can be built; however, as the dimensionality of the data increases, creating a custom model for each attribute is tedious. To address this problem, we develop \textit{heterogeneous} mixture models, where the topology and the dispersion characteristics are attribute-specific and shared across mixture components.

Suppose we have a family of steep EDMs $\{\Phi_{\mathcal{A}}\}_{j}$ to model the underlying distribution of the $j^{th}$ attribute. More precisely, we assume that each distribution in that family has the same support and can be specified with the mother function $\phi_{j}(\cdot\mid \alpha)$ and hyper-parameter domain $\mathcal{A}_{j}$. We can write the quasi-log-likelihood of data under the heterogeneous mixture model using the saddle-point approximation as
\begin{align}
\hat{\mathcal{L}} = \log\hat{p}_{\boldsymbol{\phi}}(\mathcal{\boldsymbol{X}} \mid \boldsymbol{\pi}, \boldsymbol{\mu},\boldsymbol{\kappa},\boldsymbol{\alpha}) = \sum\limits_{i=1}^{N}\log\sum\limits_{h=1}^{K}\pi_{h}\exp\left(\Upsilon(\boldsymbol{x_{i}},\boldsymbol{\mu}_{h},\boldsymbol{\kappa},\boldsymbol{\alpha})\right),
\end{align}
where $\Upsilon(\boldsymbol{x},\boldsymbol{\mu},\boldsymbol{\kappa},\boldsymbol{\alpha}) \doteq -\sum_{j=1}^{J}\left(\frac{1}{\kappa_{j}}d_{\phi_{j}}(x_{j},\mu_{j}\mid \alpha_{j})+\frac{1}{2}\log(2\pi\kappa_{j}\upsilon_{j}(x_{j}\mid\alpha_{j}))\right)$ is an auxiliary function. In practice, we algorithmically detect the type of each attribute, and we set $\{\Phi_{\mathcal{A}}\}_{j}$ accordingly. Table~\ref{tbl:ada_edm_family} summarizes the properties of the EDM families for different data types. We present the \textit{AdaCluster} algorithm, which maximizes the quasi-log-likelihood $\hat{\mathcal{L}}$ using expectation-maximization.

In the E-step of AdaCluster, the weight of each mixture component is given by
\begin{align}
p_{\boldsymbol{\phi}}(h \mid \boldsymbol{x_{i}},\boldsymbol{\mu},\boldsymbol{\kappa},\boldsymbol{\alpha},\boldsymbol{\pi}) \leftarrow \frac{\pi_{h}\exp\left(\Upsilon(\boldsymbol{x_{i}},\boldsymbol{\mu}_{h},\boldsymbol{\kappa},\boldsymbol{\alpha})\right) }{\sum_{h=1}^{K}\pi_{h}\exp\left(\Upsilon(\boldsymbol{x_{i}},\boldsymbol{\mu}_{h},\boldsymbol{\kappa},\boldsymbol{\alpha})\right)}
\label{eq:ada_adacluster_e_step}
\end{align}
for $i=1,\dots,N$ and $h=1,\dots,K$. Compared to the E-step of Bregman soft clustering (Eq.~\ref{eq:ada_bsc_e_step} ), we observe that the base measure terms do not cancel in AdaCluster as a consequence of the heterogeneity assumption. The inclusion of the base measure terms require us to specify the full density and not just the divergence-generating functions $\boldsymbol{\phi}$. We use the saddle-point approximation to express the base measure in terms of the variance function (Eq.~\ref{eq:ada_saddle_cont},\ref{eq:ada_saddle_discrete}).

In the M-step of AdaCluster, the maximum likelihood estimate of $\pi_{h}$ is equivalent to Eq.~\ref{eq:ada_bsc_m_step_pi}. The derivative of the quasi log likelihood with respect to a canonical parameter $\xi$ is given by
\begin{align}
\frac{\partial \hat{\mathcal{L}}}{\partial \xi} &\leftarrow \sum\limits_{i=1}^{N} \sum\limits_{h=1}^{K} p_{\boldsymbol{\phi}}(h \mid \boldsymbol{x_{i}},\boldsymbol{\mu},\boldsymbol{\kappa},\boldsymbol{\alpha},\boldsymbol{\pi}) \frac{\partial }{\partial \xi}\Upsilon(\boldsymbol{x_{i}},\boldsymbol{\mu}_{h},\boldsymbol{\kappa},\boldsymbol{\alpha}).
\label{eq:ada_adacluster_m_step_canonical}
\end{align}
Using Eq.~\ref{eq:ada_adacluster_m_step_canonical}, we calculate the updates for $\boldsymbol{\mu},\boldsymbol{\kappa}$ and $\boldsymbol{\alpha}$ (Algorithm~\ref{alg:ada_adacluster}). We first note that the MLE of $\boldsymbol{\kappa}$ has a closed form. Since we assume the underlying distribution is a steep EDM, the maximum likelihood estimate of $\boldsymbol{\mu}$ also has a closed form (Theorem~\ref{th:ada_steep_map}). For the hyper-parameter vector $\boldsymbol{\alpha}$, numerical optimization techniques may be used to find the value of $\boldsymbol{\alpha}$ that maximizes the quasi-log-likelihood $\hat{\mathcal{L}}$.
\begin{algorithm}[t]
  \caption{AdaCluster : EM for a heterogeneous mixture of EDMs}
  \label{alg:ada_adacluster}
  \begin{algorithmic}
     \STATE {\bfseries Input:} Data $\mathcal{X} = \left \{ \boldsymbol{x_{i}} \right \}_{i=1}^{N}$, number of clusters $K$\\
     Parametrized family of steep EDMs $\{\Phi_{\mathcal{A}}^{j}\}_{j=1}^{J} = \left \{ \phi_{j}(\cdot \mid \alpha) : \alpha \in \mathcal{A}_{j} \right \}_{j=1}^{J}$
     \STATE {\bfseries Output:} Soft partition $\left\{\{p_{\boldsymbol{\phi}}(h \mid \boldsymbol{x_{i}},\boldsymbol{\mu},\boldsymbol{\kappa},\boldsymbol{\alpha},\boldsymbol{\pi})\}_{h=1}^{K}\right\}_{i=1}^{N}$
     \STATE {\bfseries Initialize} $\left \{ \boldsymbol{\mu_{h}} \right \}_{h=1}^{K}, \boldsymbol{\kappa}$ and $\boldsymbol{\alpha}$.
     \REPEAT
     \FOR{$i=1$ {\bfseries to} $N$}
     \FOR{$h=1$ {\bfseries to} $K$}
     \STATE Calculate $p_{\boldsymbol{\phi}}(h \mid \boldsymbol{x_{i}},\boldsymbol{\mu},\boldsymbol{\kappa},\boldsymbol{\alpha},\boldsymbol{\pi})$ as in Eq.~\ref{eq:ada_adacluster_e_step}
     \ENDFOR
     \ENDFOR
     \FOR{$h=1$ {\bfseries to} $K$}
     \STATE $\pi_{h} \leftarrow \frac{1}{N}\sum_{i=1}^{N}p_{\boldsymbol{\phi}}(h \mid \boldsymbol{x_{i}},\boldsymbol{\mu},\boldsymbol{\kappa},\boldsymbol{\alpha},\boldsymbol{\pi})$
     \ENDFOR
     \FOR{$j=1$ {\bfseries to} $J$}
     \FOR{$h=1$ {\bfseries to} $K$}
     \STATE $\mu_{hj} \leftarrow \frac{\sum_{i=1}^{N}p_{\boldsymbol{\phi}}(h \mid \boldsymbol{x_{i}},\boldsymbol{\mu},\boldsymbol{\kappa},\boldsymbol{\alpha},\boldsymbol{\pi})x_{ij}}{\sum_{i=1}^{N}p_{\boldsymbol{\phi}}(h \mid \boldsymbol{x_{i}},\boldsymbol{\mu},\boldsymbol{\kappa},\boldsymbol{\alpha},\boldsymbol{\pi})}$
     \ENDFOR
     \STATE $\kappa_{j} \leftarrow 2 \frac{\sum_{i=1}^{N} \sum_{h=1}^{K} p_{\boldsymbol{\phi}}(h \mid \boldsymbol{x_{i}},\boldsymbol{\mu},\boldsymbol{\kappa},\boldsymbol{\alpha},\boldsymbol{\pi})d_{\phi_{j}}(x_{ij},\mu_{hj}\mid \alpha_{j})}{\sum_{i=1}^{N} \sum_{h=1}^{K} p_{\boldsymbol{\phi}}(h \mid \boldsymbol{x_{i}},\boldsymbol{\mu},\boldsymbol{\kappa},\boldsymbol{\alpha},\boldsymbol{\pi})}$
     \STATE Optimize for $\alpha_{j}$ using the gradient
     \STATE $\frac{\partial \hat{\mathcal{L}}}{\partial \alpha_{j}} \leftarrow -\sum_{i=1}^{N} \sum_{h=1}^{K} p_{\boldsymbol{\phi}}(h \mid \boldsymbol{x_{i}},\boldsymbol{\mu},\boldsymbol{\kappa},\boldsymbol{\alpha},\boldsymbol{\pi})\left ( \frac{1}{\kappa_{j}}\frac{\partial d_{\phi_{j}}(x_{ij},\mu_{hj}\mid\alpha_{j})}{\partial \alpha_{j}} + \frac{1}{2\upsilon_{j}(x_{ij}\mid\alpha_{j})}\frac{\partial \upsilon_{j}(x_{ij}\mid\alpha_{j})}{\partial \alpha_{j}} \right )$
     \ENDFOR
     \UNTIL convergence of quasi-log-likelihood
  \end{algorithmic}
\end{algorithm}

To incorporate a prior knowledge about the data, we discuss the Bayesian treatment of the heterogeneous mixture of steep EDMs where we have conjugate priors on $\boldsymbol{\mu}$ and $\boldsymbol{\kappa}$. For instance, in census data, we may fit a mixture model to the census data of all U.S. states and use the estimated parameters as priors on the mixture model for the state-level census data. Apart from such hierarchical applications, the Bayesian approach also offers a way to regularize the dispersion parameters. In Gaussian mixture model, prior on $\kappa$ is equivalent to regularizing the covariance matrix estimation with a diagonal matrix.

Suppose we have a conjugate prior on $\theta_{hj} = \tau^{-1}(\mu_{hj})$ with parameters $a^{\mu}_{hj}$ and $b^{\mu}_{hj}$, as in Eq.~\ref{eq:ada_exponential_family_conjugate}. Then, the MAP estimate of $\mu_{hj}$ is given by
\begin{align}
\mu_{hj}^{MAP} \leftarrow \frac{a^{\mu}_{hj}b^{\mu}_{hj}\kappa_{j} + \sum_{i=1}^{N}p_{\boldsymbol{\phi}}(h \mid \boldsymbol{x_{i}},\boldsymbol{\mu},\boldsymbol{\kappa},\boldsymbol{\alpha},\boldsymbol{\pi})x_{ij}}{b^{\mu}_{hj}\kappa_{j} + \sum_{i=1}^{N}p_{\boldsymbol{\phi}}(h \mid \boldsymbol{x_{i}},\boldsymbol{\mu},\boldsymbol{\kappa},\boldsymbol{\alpha},\boldsymbol{\pi})}
\end{align}
for $h=1,\dots,K$ and $j=1,\dots,J$. In the mixture model setting, $a^{\mu}_{hj}$ reflects our prior knowledge about the location of the $h^{th}$ centroid along the $j^{th}$ dimension, and $b^{\mu}_{hj}$ represents the effective sample size of the prior. The MAP estimate reduces to the ML estimate when $b^{\mu}_{hj}$ is set to $0$. It is also possible to set $b^{\mu}_{hj} = \eta/\kappa_{j}$, where $\eta$ is the adjusted effective sample size of the prior; $\eta$ represents how much weight is given to the prior relative to the effective cluster size $\sum_{i=1}^{N}p_{\boldsymbol{\phi}}(h \mid \boldsymbol{x_{i}},\boldsymbol{\mu},\boldsymbol{\kappa},\boldsymbol{\alpha},\boldsymbol{\pi})$.

In describing the MAP estimate for $\boldsymbol{\kappa}$, we first note that the inverse-gamma (IG) distribution is the conjugate prior for the dispersion parameter of a steep EDM under the saddle-point approximation. Suppose we have an independent $IG(a^{\kappa}_{j},b^{\kappa}_{j})$ prior on each $\{\kappa_{j}\}_{j=1}^{J}$. Then, the MAP estimate for $\kappa_{j}$ is given by
\begin{align}
\kappa_{j}^{MAP} \leftarrow \frac{b^{\kappa}_{j} + \sum_{i=1}^{N} \sum_{h=1}^{K} p_{\boldsymbol{\phi}}(h \mid \boldsymbol{x_{i}},\boldsymbol{\mu},\boldsymbol{\kappa},\boldsymbol{\alpha},\boldsymbol{\pi})d_{\phi_{j}}(x_{ij},\mu_{hj}\mid\alpha_{j})}{a^{\kappa}_{j} + \frac{1}{2}\sum_{i=1}^{N} \sum_{h=1}^{K} p_{\boldsymbol{\phi}}(h \mid \boldsymbol{x_{i}},\boldsymbol{\mu},\boldsymbol{\kappa},\boldsymbol{\alpha},\boldsymbol{\pi})}
\end{align}
for $j=1,\dots,J$. For the Bayesian version of AdaCluster, we replace the $\mu_{hj}$ and $\kappa_{j}$ updates with MAP estimates, and we adjust the gradient $\frac{\partial \hat{\mathcal{L}}}{\partial \alpha_{j}}$ accordingly.

\section{Hard clustering for heterogeneous data}
\label{sec:ada_hard}
For homogeneous mixture of steep EDMs, the Bregman soft-clustering algorithm reduces to a k-means-like Bregman hard-clustering algorithm under the small variance assumption (i.e., as $\kappa \rightarrow 0$)~\citep{banerjee2005clustering}. The objective of the Bregman hard-clustering algorithm is given by
\begin{align}
\mathcal{L}_{BHC}(\mathcal{C}(\mathcal{X})\mid \boldsymbol{\mu}) = \sum\limits_{h=1}^{K}\sum\limits_{\boldsymbol{x_{i}}\in \boldsymbol{\mathcal{X}_{h}}}\sum\limits_{j=1}^{J}d_{\phi}(x_{ij},\mu_{hj})
\label{eq:ada_bhc_objective}
\end{align}
where $\mathcal{C}(\mathcal{X}) = \{\mathcal{X}_{h}\}_{h=1}^{K}$ is the partition of the data set and $d_{\phi}(x,y)$ is the corresponding Bregman divergence to the underlying steep EDM.

\begin{algorithm}[t!]
  \caption{GMoM Hard Clustering (GMoM-HC)}
  \label{alg:ada_gmomhc}
  \begin{algorithmic}
     \STATE {\bfseries Input:} Data $\mathcal{X} = \left \{ \boldsymbol{x_{i}} \right \}_{i=1}^{N}$, number of clusters $K$\\
     Parametrized family of steep EDMs $\{\Phi_{\mathcal{A}}^{j}\}_{j=1}^{J} = \left \{ \phi_{j}(\cdot \mid \alpha) : \alpha \in \mathcal{A}_{j} \right \}_{j=1}^{J}$
     \STATE {\bfseries Output:} Hard partition $\mathcal{C}(\mathcal{X}) = \{\mathcal{X}_{h}\}_{h=1}^{K}$
     \STATE {\bfseries Initialize} $\mathcal{C}(\mathcal{X})$ with k-means++
     \REPEAT
     \STATE $\boldsymbol{\lambda} =  {\operatorname{argmin}}_{\boldsymbol{\lambda'} \in \boldsymbol{\Lambda}}\;\;\mathcal{L}_{GMoM}(\mathcal{C}(\mathcal{X});\boldsymbol{\lambda'})$
     \STATE $\boldsymbol{W}_{hj} \leftarrow \boldsymbol{W}_{hj}(\mathcal{C}(\mathcal{X});\boldsymbol{\lambda})$
     \STATE $\boldsymbol{\mathcal{X}_{h}} \leftarrow \varnothing$ for $h = 1,\dots,K$
     \FOR{$i=1$ {\bfseries to} $N$}
     \STATE $h = \underset{h'}{\operatorname{argmin}}\sum\limits_{j=1}^{J}\boldsymbol{m}_{h'j}(\boldsymbol{x}_{i})^{T}\boldsymbol{W}_{h'j}\boldsymbol{m}_{h'j}(\boldsymbol{x}_{i})$
     \STATE $\boldsymbol{\mathcal{X}_{h}} \leftarrow \boldsymbol{\mathcal{X}_{h}} \cup \left \{ \boldsymbol{x_{i}} \right \}$
     \ENDFOR
     \UNTIL convergence
  \end{algorithmic}
\end{algorithm}

In the case of a heterogenous mixture of steep EDMs, Bregman soft-clustering does not reduce to Bregman hard-clustering in general. The only exception is when the dispersion parameter is shared across attributes, which may be possible if the data are normalized within each attribute. In this case, we can combine the Bregman divergences specific to each attribute into a single multi-dimensional Bregman divergence using the linearity property. We can show that Bregman soft-clustering reduces to Bregman hard-clustering under the small-variance assumption, with the objective
\begin{align}
\mathcal{L}_{BHC}(\mathcal{C}(\mathcal{X})\mid \boldsymbol{\mu}, \boldsymbol{\alpha}) = \sum\limits_{h=1}^{K}\sum\limits_{\boldsymbol{x_{i}}\in \boldsymbol{\mathcal{X}_{h}}}\sum\limits_{j=1}^{J}d_{\phi_{j}}(x_{ij},\mu_{hj}\mid\alpha_{j}).
\label{eq:ada_bhc_objective_naive}
\end{align}
We note that the topological properties of the attributes need to be specified ahead of time and the Bregman hard-clustering algorithm cannot adaptively infer the topology. One might attempt to learn the parameters $\boldsymbol{\alpha}$ by minimizing Eq.~\ref{eq:ada_bhc_objective} directly. This naive approach fails in practice; similar accounts have been described for the Tweedie class~\citep{dunn2005series,dunn2008evaluation}.

We explore alternatives to the likelihood approach to adaptively learn the parameters of a general heterogeneous mixture of steep EDMs in the setting of hard clustering. Suppose we have $N$ i.i.d. samples $\{\boldsymbol{x_{i}}\}_{i=1}^{N}$ from a steep EDM $p_{\phi}(x\mid \mu,\kappa,\alpha)$ such that $\frac{1}{N}\sum_{i=1}^{N}\boldsymbol{x_{i}} \in int(C)$. Then we can estimate $\mu,\kappa$, and $\alpha$ with method of moments using the first three sample moments. Estimating parameters using high order moments suffer from statistical efficiency problems. Fortunately, the hard clustering problem has an additional property that we can exploit. The parameters $\boldsymbol{\kappa}$ and $\boldsymbol{\alpha}$ are shared across mixture components, and each cluster is assumed to be disjoint. Therefore, the first two population moments for $K$ clusters allow us to write down $2JK$ moment conditions for $J(K+2)$ parameters. For $K=2$, we have exact identification, i.e., the number of moment conditions is the same as the number of parameters to estimate. When $K > 2$, we have more moment conditions than parameters, leading to an overidentified system using the first two population moments for each cluster. This is the ideal setting for Generalized Method of Moments (GMoM)~\citep{hansen1982large}. GMoM is a method of estimating the set of parameters of a model using moment conditions and arbitrary functions of random variables and parameters that are zero in expectation. Given a vector of moment conditions, GMoM minimizes the quadratic form defined using an arbitrary positive definite weight matrix~\citep{hansen1982large}. The choice of the weight matrix determines the efficiency of the estimator.

For the remainder of this section, we denote the vector of parameters in a heterogeneous mixture of steep EDMs as $\boldsymbol{\lambda} = \{\boldsymbol{\mu},\boldsymbol{\kappa},\boldsymbol{\alpha}\}$. We denote the feasible set for $\boldsymbol{\lambda}$ with $\boldsymbol{\Lambda}$. We construct the set $\boldsymbol{\Lambda}$ using $\left \{\Omega_{j} \right \}_{j=1}^{J}$ for the mean parameters $\boldsymbol{\mu}$, $\mathbb{R}^{J}_{+}$ for the dispersion parameters $\boldsymbol{\kappa}$, and $\left \{\mathcal{A}_{j} \right \}_{j=1}^{J}$ for the hyper-parameters $\boldsymbol{\alpha}$.

Using the parametrized variance function of steep EDMs (Eq.~\ref{eq:ada_scaled_variance}), we write the first two moments for the $j^{th}$ attribute and the $h^{th}$ cluster as
\begin{align*}
\boldsymbol{m}_{hj}(\boldsymbol{x};\boldsymbol{\lambda}) = \begin{bmatrix}
x_{j} - \mu_{hj}\\
x_{j}^{2} - \kappa_{j}\upsilon_{j}(\mu_{hj}\mid\alpha_{j})
\end{bmatrix}.
\end{align*}
Since $E[\boldsymbol{m}_{hj}(\boldsymbol{x};\boldsymbol{\lambda})] = \boldsymbol{0}$ for each attribute  $j = \{1, \dots, J\}$ and each cluster $h = \{1, \dots, K\}$, they can be used as moment conditions in GMoM. Having described the moment conditions, we need to specify a positive-definite weight matrix to formulate the GMoM objective. In Hansen's iterative GMoM method, the weight matrix is constructed by taking the inverse of the residual matrix~\citep{hansen1982large}. This construction yields asymptotically efficient estimates for the parameters. A variant of the iterative GMoM is the Continuously Updating Generalized Method of Moments (CUGMoM), which is preferred when analytical solutions in each iteration of the original method are not available~\citep{hansen1996finite}. Instead of iterating between the update for the parameter estimates and the weight matrix estimate, CUGMoM optimizes for the parameters directly using computational techniques.

The CUGMoM objective is given by
\begin{align}
\mathcal{L}_{GMoM}(\mathcal{C}(\mathcal{X}); \boldsymbol{\lambda}) = \sum\limits_{h=1}^{K}\sum\limits_{j=1}^{J}\boldsymbol{\overline{m}_{hj}}^{T}\boldsymbol{W}_{hj}
\boldsymbol{\overline{m}_{hj}}
\end{align}
where
\begin{align}
\boldsymbol{\overline{m}_{hj}}(\mathcal{C}(\mathcal{X}); \boldsymbol{\lambda}) &= \frac{1}{\left | \boldsymbol{\mathcal{X}_{h}} \right |}\sum\limits_{\boldsymbol{x_{i}}\in \boldsymbol{\mathcal{X}_{h}}}\boldsymbol{m_{hj}}(\boldsymbol{x_{i}};\boldsymbol{\lambda})\\
\boldsymbol{W}_{hj}(\mathcal{C}(\mathcal{X}); \boldsymbol{\lambda}) &= \left (\frac{1}{\left | \boldsymbol{\mathcal{X}_{h}} \right |}\sum\limits_{\boldsymbol{x_{i}}\in \boldsymbol{\mathcal{X}_{h}}}\boldsymbol{m_{hj}}(\boldsymbol{x_{i}}; \boldsymbol{\lambda})\;\boldsymbol{m_{hj}}(\boldsymbol{x_{i}}; \boldsymbol{\lambda})^{T} \right )^{-1}.
\end{align}
The CUGMoM parameter estimates are $\boldsymbol{\hat{\lambda}} =  {\operatorname{argmin}}_{\boldsymbol{\lambda} \in \boldsymbol{\Lambda}}\;\mathcal{L}_{GMoM}(\mathcal{C}(\mathcal{X}); \boldsymbol{\lambda})$. To solve this optimization problem, we used the limited-memory Broyden-Fletcher-Goldfarb-Shanno with boundaries (L-BFGS-B) algorithm~\citep{byrd1995limited}. The computational bottleneck for the GMoM routine is usually the inversion of the weight matrix. In the hard clustering problem, clusters are disjoint; hence, the residual matrix and the weight matrix are block diagonal. The inversion of the weight matrix then has complexity $O(KJ)$ rather than $O(K^3J)$.

GMoM gives us a way to estimate the parameters of the model given the partition of the data into clusters, $\mathcal{C}(\mathcal{X})$. This is somewhat equivalent to the M-step of AdaCluster. To mimic the E-step of AdaCluster, we use the weight matrix of each cluster to calculate the quadratic distance, and then we assign each sample to the cluster with minimum distance. We refer to this algorithm (Alg.~\ref{alg:ada_gmomhc}) as GMoM Hard Clustering (\textit{GMoM-HC}). Convergence can be assessed either through the objective or through a stable partition. A lighter version of this algorithm can also be derived by setting the mean estimates to the maximum likelihood estimates. In this case, the only moment conditions are the second population moments; then, the inverse weight matrix becomes a diagonal matrix with the mean squared error for the moment conditions as the diagonal entries. The analogous Bayesian algorithm can be derived by fixing $b^{\mu}_{h}$ many pseudo samples at location $\boldsymbol{a^{\mu}_{h}}$ to the $h^{th}$ cluster. Recall that the geometric interpretation of conjugate prior to the mean parameter is having $b$ many pseudo samples at a location $a$~\citep{agarwal2010geometric}.

\section{Results}
\label{sec:ada_results}
\subsection{Setup}
\label{subsec:ada_setup}
We used normalized mutual information (NMI) to quantify the results from a clustering algorithm with respect to ground truth. NMI is a metric that takes values between $0$ and $1$, where $0$ corresponds to random cluster assignments.

In addition to AdaCluster and GMoM-HC, we used k-means and the EM algorithm for a Gaussian mixture model (GMM) for comparison. We force the covariance matrix of the GMM to be diagonal and shared across mixture components. We note that AdaCluster reduces to a GMM when $\Phi_{\mathcal{A}}$ is a singleton set with only Gaussian members. We also note that in such a setting $\boldsymbol{\kappa} = \{\kappa\}_{j=1}^{J}$ corresponds to the diagonal entries of the GMM covariance matrix and is shared across clusters.

To set up AdaCluster and GMoM-HC, we categorized each attributes based on whether we found all of the values to be positive, non-negative, or on the real line, and whether the values were discrete or continuous. Based on the different categories of attribute values, we selected the parameterization of the EDM, $\Phi_{\mathcal{A}}$, as outlined in Table~\ref{tbl:ada_edm_family}.

To initialize all of the algorithms in this comparison, we used \textit{k-means++}~\citep{arthur2007k}. In AdaCluster and GMoM-HC, we used the k-means++ centroids to set the hyper-parameters $\boldsymbol{a^{\mu}_{h}}$ while fixing $\boldsymbol{b^{\mu}_{h}} = \boldsymbol{1}$. We set the hyper-parameters of $\boldsymbol{\kappa}$ as $a^{\kappa} = 1.0, b_{\kappa} = 10^{-9}$ to ensure numerical stability. Similarly, we add $10^{-9}$ to the diagonal entries of the covariance matrix in the GMM.

We ran each algorithm for a maximum of $1,000$ iterations and terminated early if the hard cluster assignments did not change in two consecutive iterations. We restarted each algorithm $1,000$ times and selected the results from these $1,000$ runs with the highest likelihood for the GMM and AdaCluster, and with the lowest inertia for k-means and GMoM-HC.

\begin{figure*}[t!]
  \begin{center}
      \begin{subfigure}[h]{0.32\textwidth}
        \includegraphics[width=\textwidth]{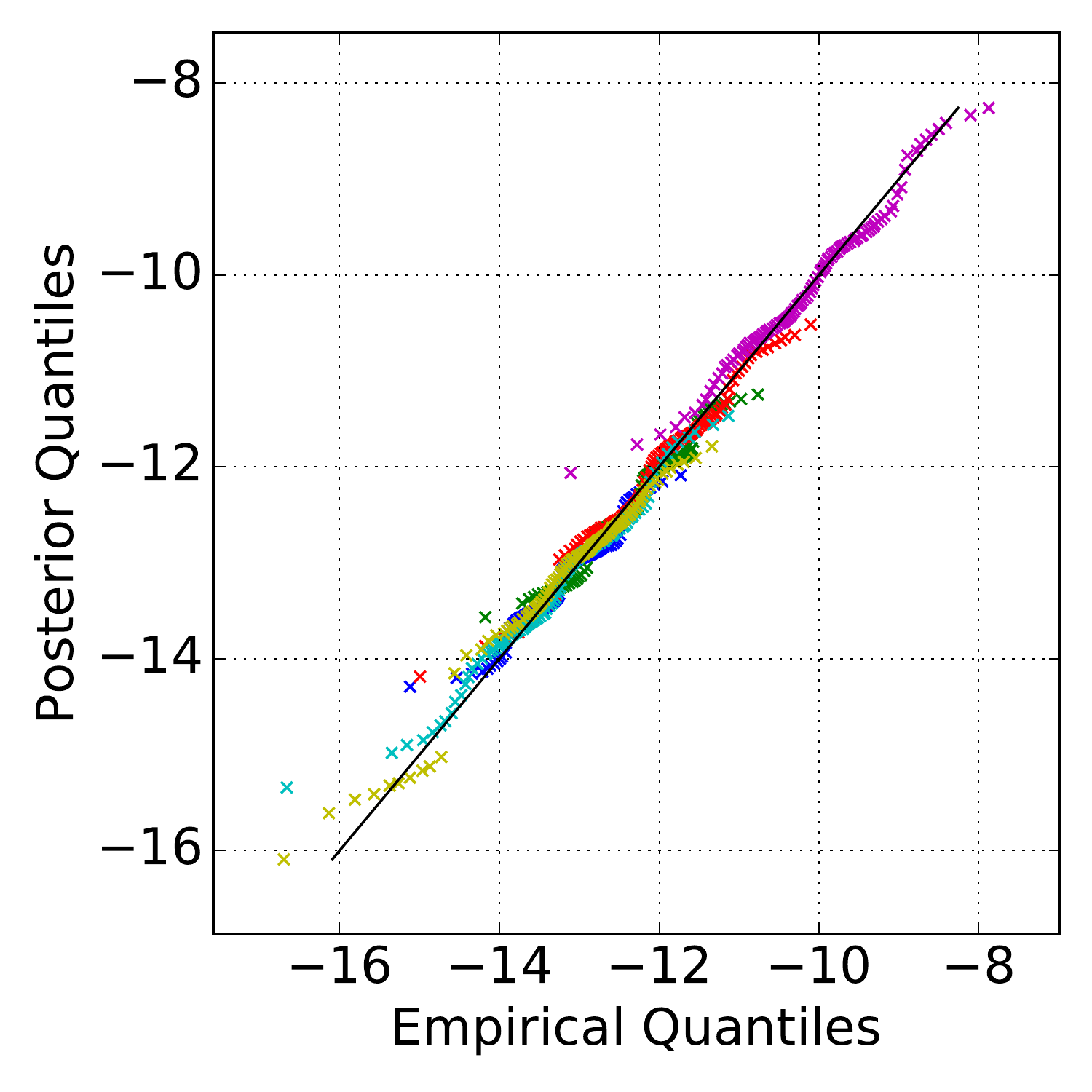}
        \caption{}
        \label{fig:ada_qq_gaussian}
      \end{subfigure}
      ~
      \begin{subfigure}[h]{0.32\textwidth}
        \includegraphics[width=\textwidth]{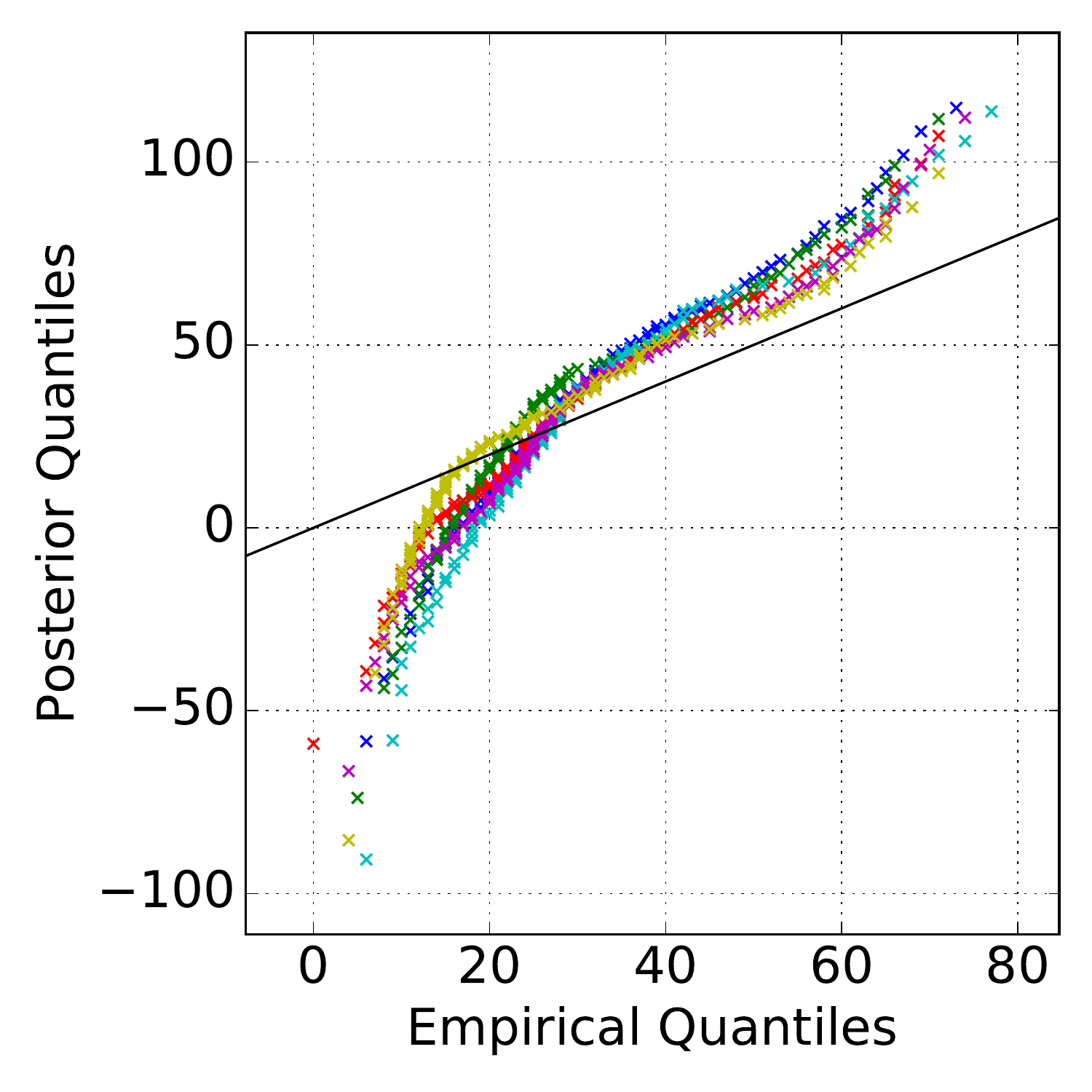}
        \caption{}
        \label{fig:ada_qq_poisson}
      \end{subfigure}
      ~
      \begin{subfigure}[h]{0.32\textwidth}
        \includegraphics[width=\textwidth]{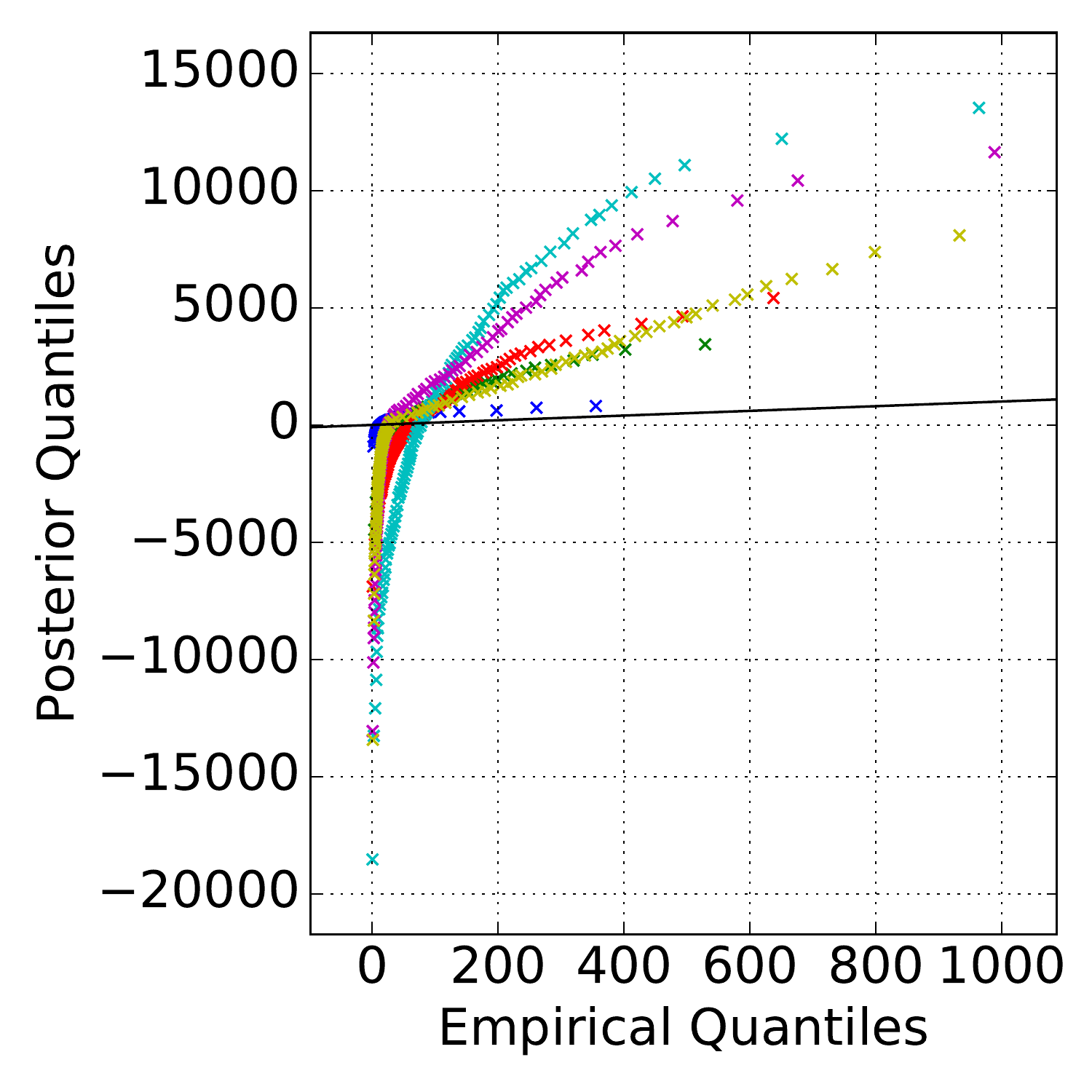}
        \caption{}
        \label{fig:ada_qq_gamma}
      \end{subfigure}

      \begin{subfigure}[h]{0.32\textwidth}
        \includegraphics[width=\textwidth]{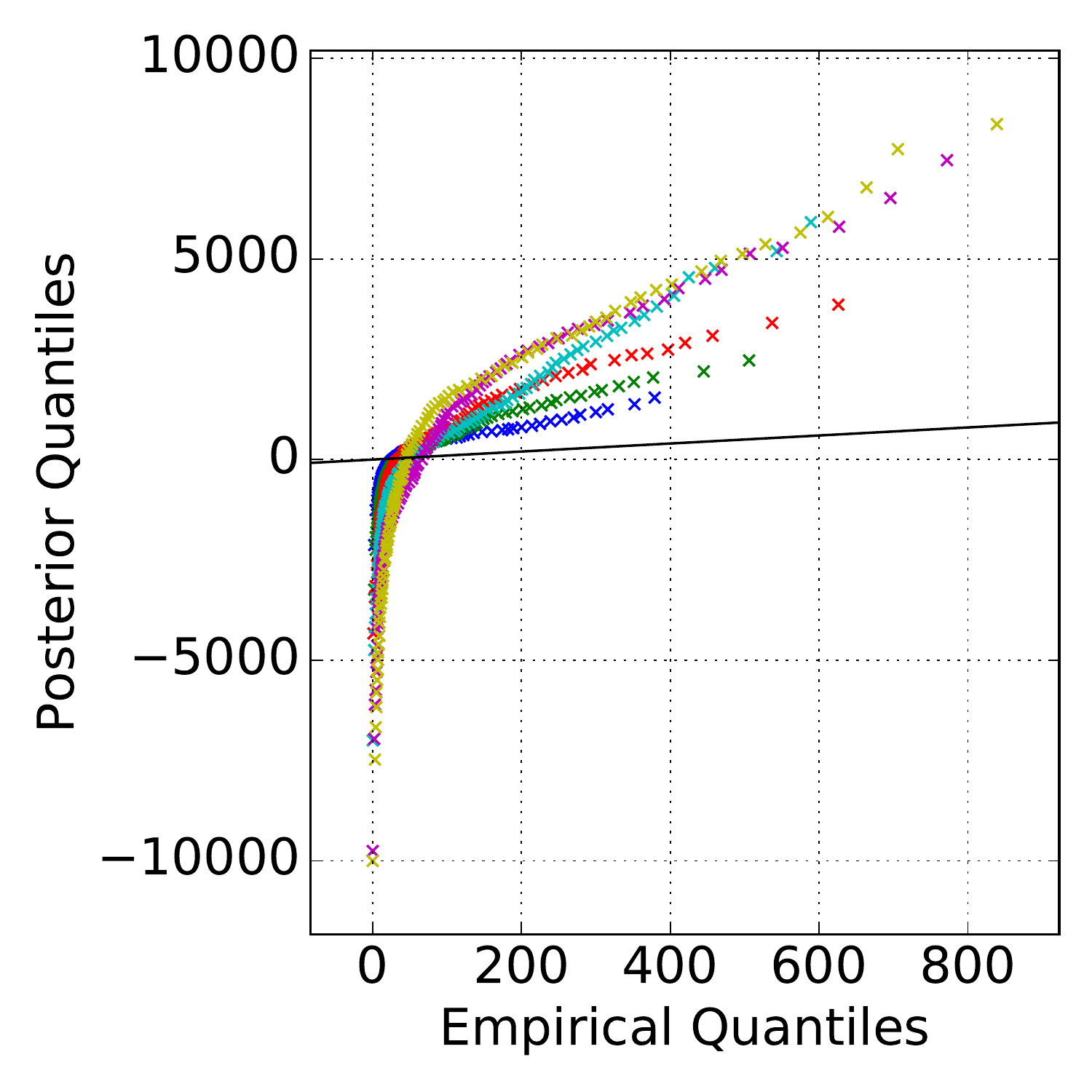}
        \caption{}
        \label{fig:ada_qq_negative_binomial}
      \end{subfigure}
      ~
      \begin{subfigure}[h]{0.32\textwidth}
        \includegraphics[width=\textwidth]{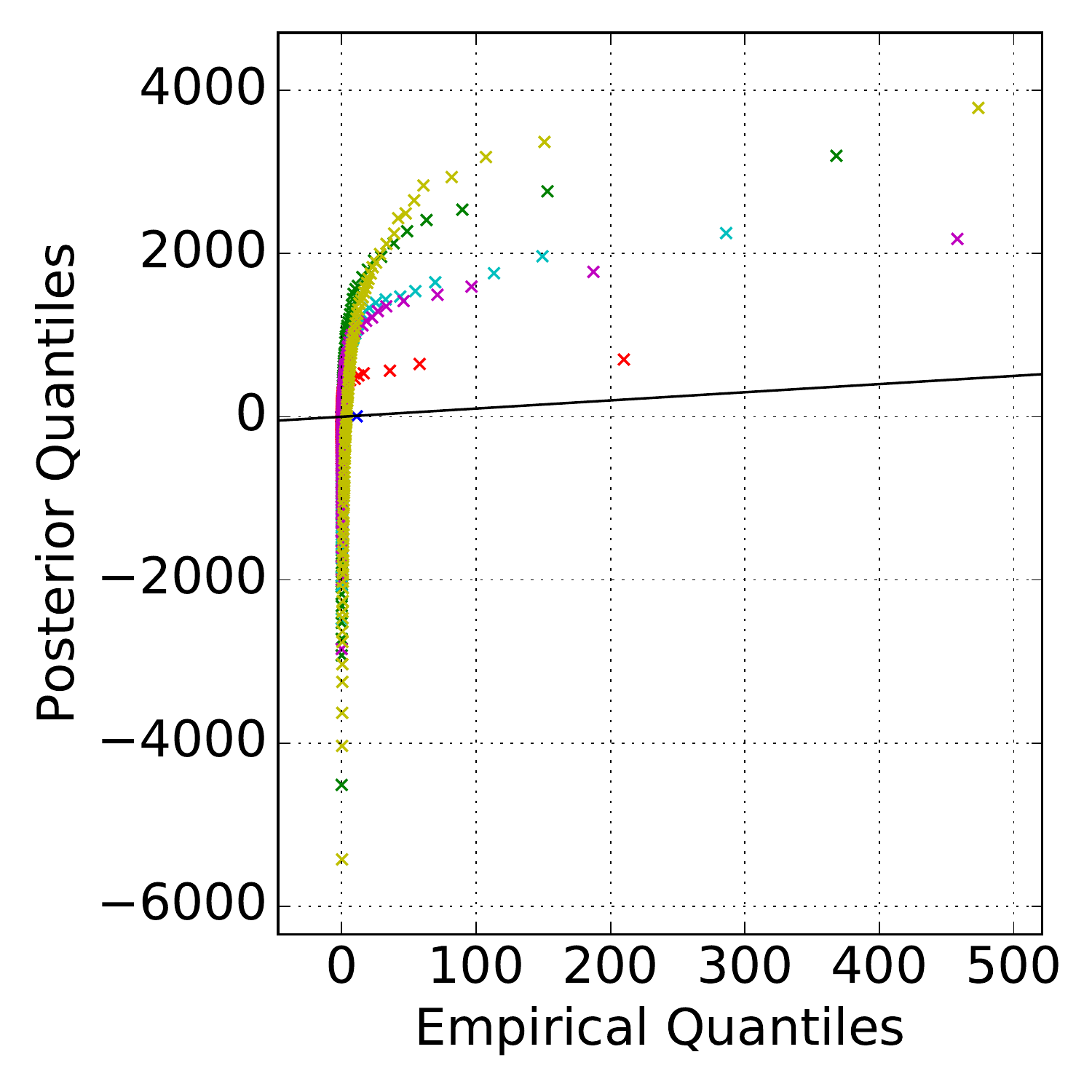}
        \caption{}
        \label{fig:ada_qq_inverse_gaussian}
      \end{subfigure}
      \caption{{\bf Simulated density estimation results.} Quantile-quantile plots of the samples generated from a homogeneous steep EDM mixture models with an underlying distribution a) \textit{Gaussian}, b) \textit{Poisson} c) \textit{gamma}, d) \textit{negative binomial}, and e) \textit{inverse-Gaussian} against samples generated from the Gaussian mixture model fitted to synthetic data. Each color corresponds to a synthetic data set with distinct parameters, and the diagonal black line marks the perfect fit.}
      \label{fig:ada_simulated_density}
  \end{center}
   \vspace{-12pt}
\end{figure*}

\subsection{Synthetic data experiments}
\label{subsec:ada_synthetic}
To illustrate the benefits of our approach to clustering on homogeneous data, we considered the problem of density estimation with mixture models. For each distribution in $\{$Gaussian, gamma, inverse-Gaussian, Poisson, negative binomial$\}$, we generated six single dimensional data sets of size $N=1000$ from a mixture model with $K=4$ mixture components. We drew the dispersion parameter $\kappa$ from an inverse-gamma with shape parameter $1.01$ and scale parameter $1.0$. We chose the centroids such that $p_{\phi}(\mu_{h'}\mid\mu_{h},\kappa,\alpha) < 0.01$ for every $h=1,\dots,K$ and $h'\neq h$. We then fit a Gaussian mixture model to the data and generated $1,000$ samples from this model.

The quantile quantile plots between the data and the posterior samples generated from the fitted GMM show that the GMM recovers the true density accurately when the underlying distribution is Gaussian (Fig.~\ref{fig:ada_simulated_density}). However, the GMM fails in the case of Gamma, inverse-Gaussian, Poisson and negative binomial. Recall that the variance function, $\upsilon(x)$, of the Gaussian is constant $(1)$, linear for Poisson $(x)$, quadratic for Gamma $(x^2)$ and negative binomial $(x^2+x)$, and finally cubic for inverse-Gaussian $(x^3)$. Hence, the GMM performance progressively degrades as the variance-mean relationship deviates from constant (Fig.~\ref{fig:ada_qq_gaussian}--\ref{fig:ada_qq_inverse_gaussian}).

\begin{figure*}[t!]
  \begin{center}
      \begin{subfigure}[h]{0.45\textwidth}
        \includegraphics[width=\textwidth]{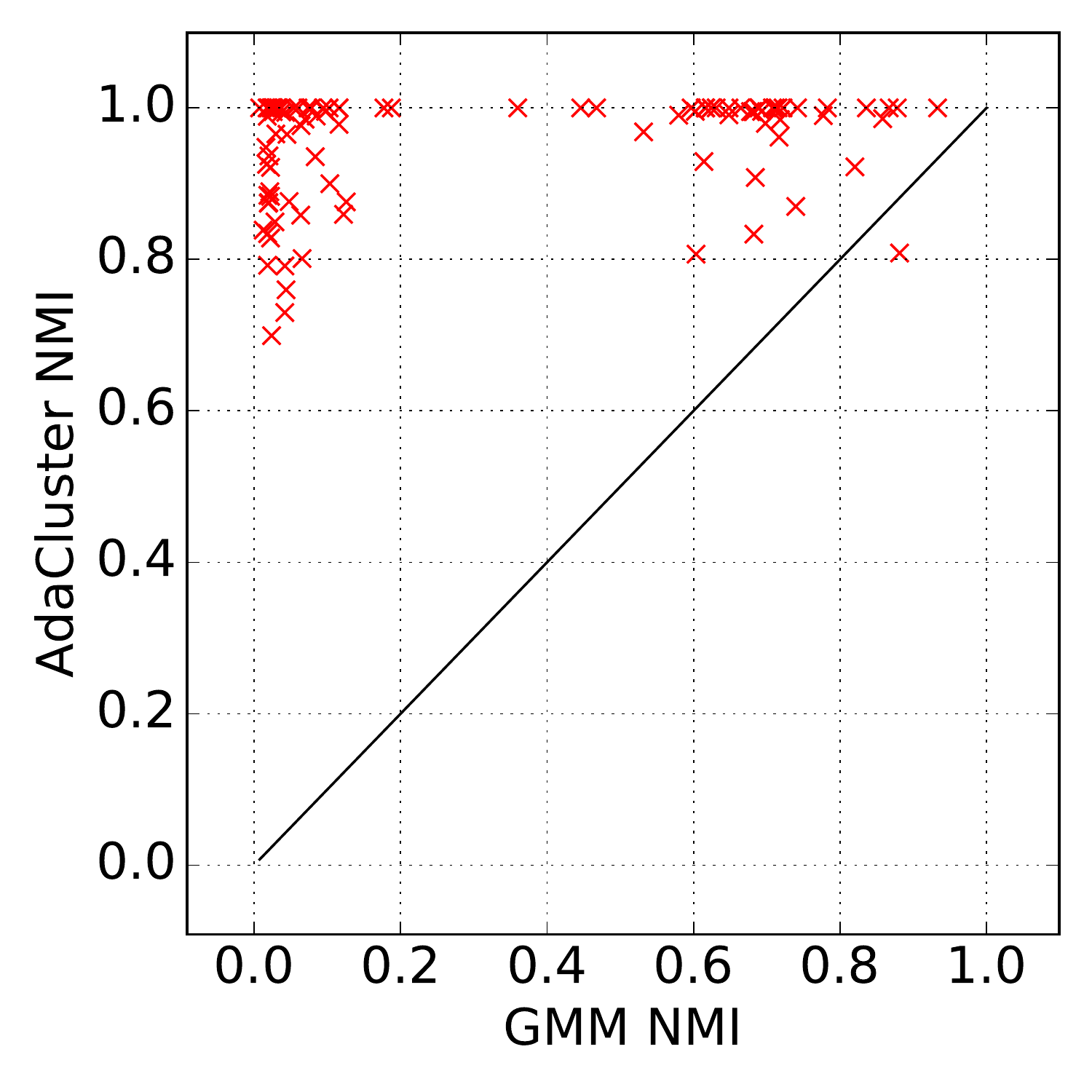}
        \caption{}
        \label{fig:ada_syn_nmi}
      \end{subfigure}
      ~
      \begin{subfigure}[h]{0.45\textwidth}
        \includegraphics[width=\textwidth]{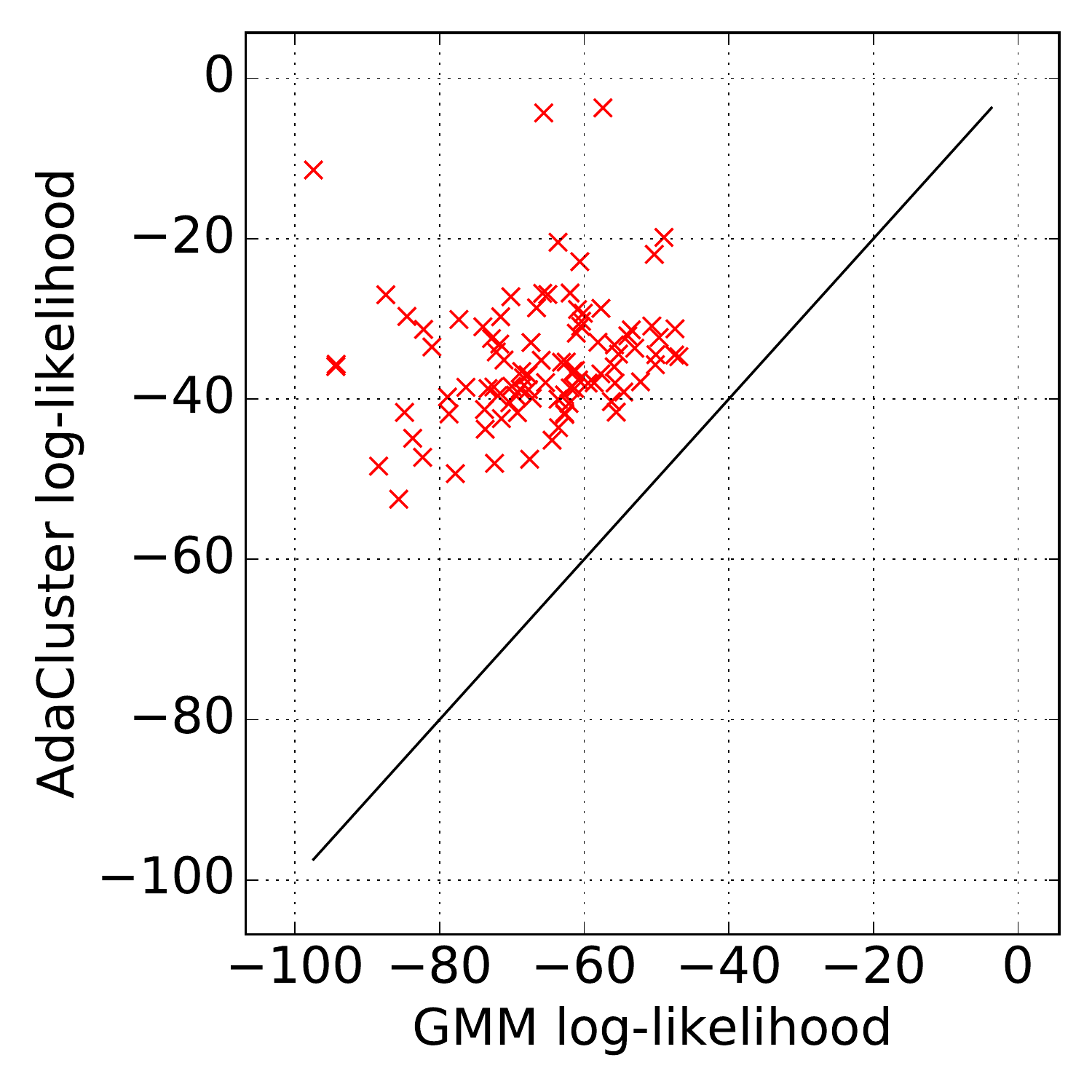}
        \caption{}
        \label{fig:ada_syn_loglik}
      \end{subfigure}
      \caption{{\bf Synthetic heterogeneous data results.} Comparing AdaCluster with the GMM on synthetic heterogeneous data in terms of a) normalized mutual information (NMI) and b) per sample log-likelihood. Each red point corresponds to a different data set, and the black diagonal line indicates equivalent performance.}
      \label{fig:ada_syn}
  \end{center}
   \vspace{-12pt}
\end{figure*}
To compare AdaCluster and GMM performance in clustering heterogenous data, we generated $100$ data sets of size $N=1000$ with $J=10$ dimensions and $K=4$ mixture components. We started by drawing cluster proportions from a Dirichlet distribution with equal concentration parameters. For each dimension $j=1,\dots,J$, we first drew a dispersion parameter $\kappa_{j}$ from an inverse-gamma with shape parameter $1.01$ and scale parameter $1.0$. We then randomly selected the underlying distribution from $\{$Gaussian, gamma, inverse-Gaussian, Poisson, negative Binomial$\}$ and set $\alpha_{j}$ and  $\phi_{j}(\cdot\mid\alpha_{j})$ accordingly. Finally, we selected the cluster centroids such that $p_{\phi_{j}}(\mu_{h'j}\mid\mu_{hj},\kappa_{j},\alpha_{j}) < 0.01$ for every $h=1,\dots,K$ and $h'\neq h$. We then ran AdaCluster and EM for the GMM as described in the previous section.

AdaCluster outperforms the GMM in every synthetic data set except for one in terms of NMI (Fig.~\ref{fig:ada_syn}). More interestingly, AdaCluster often achieves perfect NMI score whereas the GMM performance is below $0.2$ in roughly half of the data sets. In fact, GMM either struggles to cluster the data (top left quadrant of Fig.~\ref{fig:ada_syn_nmi}) or achieves a reasonable NMI value (top right quadrant of Fig.~\ref{fig:ada_syn_nmi}). 
In terms of the quasi-log-likelihood, AdaCluster outperforms GMM in every data set (Fig.~\ref{fig:ada_syn_loglik}), suggesting that AdaCluster not only yields a better clustering of data than GMM but also provides a better fit. We note that the quasi-log-likelihood corresponds to the exact log-likelihood in the case of GMM since saddle-point approximation is exact for the Gaussian distribution.

\begin{table*}[t!]
\caption{{\bf Comparison of NMI values for EM for the Gaussian mixture model (GMM), AdaCluster (AdaC.), k-means, and GMoM hard clustering (GMoM-HC) on nine UCI data sets.} $N$ is the number of samples; $J$ is the number of attributes; $K$ is the number of clusters; $S$ is the support.}
\begin{center}
\begin{sc}
\begin{tabular}{lrrrrrrrr}
\abovespace\belowspace
Data set     &     N &   J &   K &               S &    GMM &  AdaC. & k-means &  GMoM-HC \\
\hline
\abovespace\belowspace
iris        &   149 &   4 &   2 & $\mathbb{R}_{+}$ & \bf{1.000} & \bf{1.000} &  0.869 & \bf{1.000} \\
wine        &   177 &  13 &   3 & $\mathbb{R}_{+}/\mathbb{Z}_{+}$ & \bf{0.974} &  0.783 &  0.426 &  0.769 \\
steel       &  1940 &  20 &   7 & $\mathbb{R}_{+}/\mathbb{Z}_{+}$ &  0.178 &  \bf{0.265} &  0.115 &  0.204 \\
yeast       &  1483 &   6 &  10 & $\mathbb{R}_{+}$ &  0.289 &  \bf{0.292} &  0.260 &  0.167 \\
wholesale   &   440 &   6 &   2 & $\mathbb{Z}_{+}$ &  0.017 &  \bf{0.442} &  0.009 &  0.309 \\
stoneflakes &    78 &   8 &   4 & $\mathbb{R}_{0}/\mathbb{Z}_{0}$ &  0.416 &  \bf{0.474} &  0.436 &  0.246 \\
seeds       &   210 &   7 &   3 & $\mathbb{R}_{+}$ &  0.581 &  \bf{0.696} &  0.695 &  0.674 \\
leaf        &   340 &  14 &  30 & $\mathbb{R}_{+}$ &  \bf{0.742} &  0.699 &  0.653 &  0.172 \\
libras      &   360 &  90 &  15 & $\mathbb{R}_{+}$ &  0.548 &  \bf{0.596} &  0.595 &  0.589 \\
\hline
\end{tabular}
\label{tbl:ada_uci}
\end{sc}
\end{center}
\vskip -12pt
\end{table*}
\subsection{UCI data experiments}
\label{subsec:ada_uci}
We performed clustering with a GMM, AdaCluster, k-means, and GMoM-HC on nine data sets from the UCI repository~\citep{Lichman:2013}. Each data set includes different attribute characteristics (Table~\ref{tbl:ada_uci}). For instance, \textit{yeast} has only positive continuous attributes whereas \textit{stoneflakes} has both non-negative discrete and non-negative continuous attributes.

In six of the nine data sets, AdaCluster outperforms the GMM, and they both achieve a perfect score for \textit{iris}. In the \textit{wine} data set, a relatively small number of samples might explain the inferior performance of AdaCluster. Since the GMM improves over k-means, this suggests that the dispersion parameters are significantly different across attributes. In that case, AdaCluster needs to learn the varying dispersion parameters as well as the hyper-parameters $\boldsymbol{\alpha}$, and the small size of the data set might explain the under fitting. The \textit{steel} and \textit{stoneflakes} data are both heterogeneous data sets, and AdaCluster outperforms the GMM. Both AdaCluster and GMoM-HC achieves better scores than the GMM and k-means. In the \textit{yeast} data set, AdaCluster has a slight edge over the GMM, and k-means and the GMM have comparable performances. When we analyzed the fitted parameters by AdaCluster and the GMM, we found that the dispersion parameter estimates are similar for each attribute. This explains why the GMM and k-means achieved roughly the same scores. Furthermore, we observed that the variance function identified by AdaCluster is sub-linear, which explains why Gaussian can be a good proxy for such distributions. In the \textit{wholesale} data set, AdaCluster and GMoM-HC beat the GMM and k-means by a substantial margin. Since all the attributes are non-negative discrete, we used the family described in Section~\ref{subsec:ada_edm_nnd}. 
The fitted parameters from AdaCluster indicate that the attributes have negative binomial characteristics, i.e., quadratic polynomial variance function. Similar to \textit{yeast}, GMM and k-means have comparable NMI scores, suggesting that the dispersion is roughly the same across attributes. In \textit{seeds} and \textit{libras}, AdaCluster and k-means have similar performances, and the GMM is inferior to both. The common characteristics of these two data sets is a relatively small dispersion across attributes. We verified this hypothesis by examining the fitted parameters by AdaCluster. Small dispersion might explain why k-means outperformed GMM. Similarly, it can be argued that the Gaussian distribution approximates the underlying distribution when the dispersion is small. Finally, GMM beats AdaCluster in \textit{leaf} data set, which has a small number of samples per cluster, leading to noisy estimates of $\alpha$ and degrading the performance of the adaptive algorithms. This is especially true for GMoM-HC, where the performance relies on having good $\alpha$ estimates within each cluster.

\section{Discussion and Conclusion}
\label{sec:ada_discussion}
In this work, we addressed the problem of clustering heterogeneous data when the underlying distribution of each attribute is unknown and heterogeneous. We first showed that the density of a steep exponential dispersion model can be represented with a Bregman divergence. We then proposed families of steep EDMs for by parametrizing the divergence-generating functions. We showed that a certain sub-family of the Morris class yields distributions with a support on non-negative numbers, including the widely used Poisson and negative-binomial distributions. We proposed another sub-family of the Morris class that can be used to model continuous data on the real line. This family has the symmetric Gaussian distribution and asymmetric generalized hyperbolic secant distribution as members. We suggested that the same family can be used for data on the unit interval by transforming data with the logit function first. Thirdly, we looked at the Tweedie class and investigated its connection to Bregman divergences in light of the theorem introduced earlier. We showed that a certain sub-family of the Tweedie class---including Gaussian, gamma, Poisson, and inverse-Gaussian distributions---may be used for positive continuous data. We further showed that another Tweedie sub-class may be used for non-negative continuous data.

Using these parametrized family of steep EDMs, we derived an expectation-maximization algorithm, \textit{AdaCluster}, for soft clustering of heterogeneous data. AdaCluster has closed form updates for the mean and dispersion parameters of each mixture component as a result of saddle-point approximation. Throughout the derivations, we highlighted the contrasts with Bregman soft-clustering algorithm~\citep{banerjee2005clustering}. We used numerical optimization techniques to identify the topology of each attribute. We also presented the Bayesian equivalent of AdaCluster, where there are conjugate priors on the mean and dispersion parameters. In deriving the Bayesian update for the dispersion parameter, we established that the inverse-gamma distribution is the conjugate prior to a steep EDM in reproductive form under the saddle-point approximation.

We then turned our attention to the hard clustering of heterogeneous data. We demonstrated that the familiar asymptotic relationship between EM and k-means like algorithms, such as Bregman soft clustering and Bregman hard clustering, does not apply to the heterogeneous data. We proposed an alternative approach using generalized method of moments with moment conditions found through the parametrized variance functions. We presented an adaptive hard-clustering algorithm (GMoM-HC) that mimics the k-means algorithm.
Finally, we studied the behavior of a Gaussian mixture model in density estimation for non-Gaussian homogeneous data. We concluded that the Gaussian mixture model fails to capture the true density as the variance-mean relationship deviates from the independence assumption that the GMM makes. We then compared the GMM and AdaCluster for clustering heterogeneous data, and showed that AdaCluster has a significant edge over GMM as measured with normalized mutual information and log-likelihood. We also compared AdaCluster and GMM on nine data sets from the UCI repository with distinct topology and dispersion characteristics. We found that that AdaCluster clusters heterogeneous data consistently better than EM for the GMM.


\acks{We would like to acknowledge support for this project
for MEB from the Princeton Innovation J. Insley Blair Pyne Fund Award, and, for BEE, from NIH R00 HG006265, NIH R01 MH101822, NIH U01 HG007900, and a Sloan Faculty Fellowship.}


\newpage








\bibliography{adaptive_clustering_for_heterogeneous_data}

\end{document}